\documentclass{article}

\usepackage[T1]{fontenc}
\usepackage[utf8]{inputenc}
\usepackage{lmodern}

\usepackage{microtype}
\usepackage{graphicx}
\usepackage{subcaption}
\usepackage{booktabs} 

\usepackage{hyperref}

\usepackage[accepted]{icml2026}

\usepackage{amsmath}
\usepackage{amssymb}
\usepackage{mathtools}
\usepackage{amsthm}
\usepackage{xspace}
\usepackage[table]{xcolor} 
\usepackage{xspace}        
\usepackage{pifont}        
\usepackage{bm}            
\usepackage{dsfont}        
\usepackage{pgf}           
\usepackage{wrapfig}       
\usepackage{tcolorbox}     
\usepackage{rotating}      
\usepackage{placeins}      
\usepackage{nicefrac}      
\usepackage{multirow}      
\usepackage{enumitem}      
\usepackage{csquotes}      
\usepackage{arydshln}

\definecolor{darkblue}{rgb}{0.0,0.0,0.65}
\definecolor{lightblue}{RGB}{100, 160, 200}
\definecolor{darkred}{rgb}{0.68,0.05,0.0}
\definecolor{lightred}{RGB}{230, 120, 120}

\definecolor{lab_orange}{RGB}{227,114,34}
\definecolor{comp_blue}{RGB}{0, 101, 189}
\definecolor{surf_green}{RGB}{162,173,0}

\newcommand{\cmark}{\ding{51}}%
\newcommand{\xmark}{\ding{55}}%

\newcommand{\tablemin}{0}
\newcommand{\tablemax}{100}

\newcommand{\heatcell}[1]{%
  \pgfmathsetmacro{\percent}{(#1-\tablemin)/(\tablemax-\tablemin)*100}%
  \edef\tempcolor{\noexpand\cellcolor{lightblue!\percent}}%
  \tempcolor\(#1\)%
}

\usepackage[capitalize,noabbrev]{cleveref}

\crefname{equation}{Eq.}{Eqs.}
\Crefname{equation}{Eq.}{Eqs.}

\crefname{figure}{fig.}{figs.}
\Crefname{figure}{Fig.}{Figs.}

\crefname{table}{table}{tables}
\Crefname{table}{Table}{Tables}

\crefname{section}{§}{§§}
\Crefname{section}{§}{§§}

\crefname{subsection}{§}{§§}
\Crefname{subsection}{§}{§§}

\crefname{subsubsection}{§}{§§}
\Crefname{subsubsection}{§}{§§}

\crefname{appendix}{§}{§§}
\Crefname{appendix}{§}{§§}

\theoremstyle{plain}

\theoremstyle{definition}

\theoremstyle{remark}

\usepackage[textsize=tiny]{todonotes}

\icmltitlerunning{Lost in Serialization: Invariance and Generalization of LLM Graph Reasoners}

\newcommand{\GOne}{\texttt{G1}\xspace}
\newcommand{\Qwen}{\texttt{Qwen}\xspace}
\newcommand{\GPT}{\texttt{gpt-oss}\xspace}
\newcommand{\Erdos}{\texttt{Erdős}\xspace}

\newcommand{\bij}{\xrightarrow{\smash{\sim}}}

\newcommand{\opt}[1]{\text{\texttt{#1}}}

\newcommand{\R}{\mathbb{R}}

\newcommand{\graph}{G}
\newcommand{\nodes}{V}
\newcommand{\edges}{E}

\newcommand{\mA}{\bm{A}}
\newcommand{\mP}{\bm{P}}
\newcommand{\lab}{\ell}
\newcommand{\ser}{\mathcal{S}}
\newcommand{\task}{\mathcal{T}}

\tcbset{
  colback=black!5,
  colframe=black!40,
  fontupper=\scriptsize\ttfamily,
  before skip=8pt,
  after skip=8pt,
  before upper={\setlength{\parskip}{6pt}\setlength{\parindent}{0pt}}
}

\begin{document}

\twocolumn[
  \icmltitle{Lost in Serialization: \\
    Invariance and Generalization of LLM Graph Reasoners}



  \icmlsetsymbol{equal}{*}

  \begin{icmlauthorlist}
    \icmlauthor{Daniel Herbst}{equal,tum}
    \icmlauthor{Lea Karbevska}{equal,cam}
    \icmlauthor{Divyanshu Kumar}{equal,enk} \\
    \icmlauthor{Akanksha Ahuja}{equal,cam}
    \icmlauthor{Fatemeh Gholamzadeh Nasrabadi}{equal,ams} \\
    \icmlauthor{Fabrizio Frasca}{tcn}
  \end{icmlauthorlist}

  \icmlaffiliation{tum}{Technical University of Munich}
  \icmlaffiliation{cam}{University of Cambridge}
  \icmlaffiliation{enk}{Enkrypt AI}
  \icmlaffiliation{ams}{University of Amsterdam}
  \icmlaffiliation{tcn}{Technion}

  \icmlcorrespondingauthor{Fabrizio Frasca}{fabriziof@campus.technion.ac.il}

  \icmlkeywords{Machine Learning, ICML}

  \vskip 0.3in
]



\printAffiliationsAndNotice{}  

\begin{abstract}
    While promising, graph reasoners based on Large Language Models (LLMs) lack built-in invariance to symmetries in graph representations. Operating on sequential graph serializations, LLMs can produce different outputs under node reindexing, edge reordering, or formatting changes, raising robustness concerns. We systematically analyze these effects, studying how fine-tuning impacts encoding sensitivity and generalization on unseen tasks. We propose a principled decomposition of graph serializations into node labeling, edge encoding, and syntax, and evaluate LLM robustness to variations of each of these factors on a comprehensive benchmarking suite. We also contribute a novel set of spectral tasks to further assess generalization abilities of fine-tuned reasoners. Results show that larger (non-fine-tuned) models are more robust. Fine-tuning reduces sensitivity to node relabeling but may increase it to variations in structure and format, while it does not consistently improve performance on unseen tasks.
\end{abstract}

\section{Introduction}
Large Language Models (LLMs) are emerging as alternatives to Graph Neural Networks (GNNs) for graph reasoning~\citep{gilmer2017neural,xu2020what,dudzik2022gnn}.
While GNNs operate on adjacency and feature matrices, LLMs consume text-based serializations of nodes and edges, which enables reasoning in token space.
LLMs can leverage their reasoning skills acquired on large-scale text corpora and support flexible in-context formulations of graph problems. 
Early out-of-the-box LLM applications have achieved limited success~\citep{fatemi2023talk,tang2025grapharena,yuan2025gracore}, but subsequent fine-tuning strategies~\citep{sanford2024understanding,guo2025g1} have substantially improved performance, allowing moderately sized models to perform nontrivial structural reasoning.

This progress raises a fundamental question: \emph{what} is learned when graph reasoners are post-trained on serialized graphs?
While GNNs are natively in-/equivariant w.r.t.\ the central graph symmetries, LLMs operating on textual encodings need not produce consistent outputs for equivalent graphs representations, raising robustness concerns~\citep{fatemi2023talk,yuan2025gracore,ge2025graph}. 
Importantly, these variations are \emph{not} equivalent from the model's perspective but can alter locality, parsing, and the algorithmic path taken in token space.
As post-training shows promise in improving graph reasoning \citep{guo2025g1}, it becomes critical to assess whether this mitigates or amplifies sensitivity to such variations, and how it affects generalization to unseen tasks.

\begin{figure}[t]
    \centering
    \includegraphics[width=\linewidth]{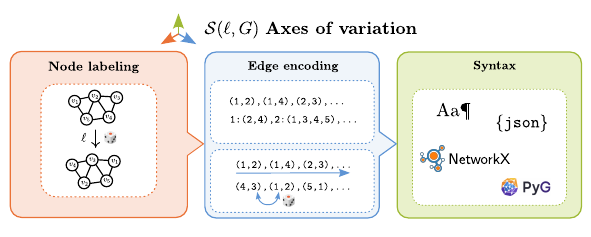}
    \caption{Systematic decomposition of a graph serialization into \textbf{\textcolor{lab_orange}{node labeling}}, \textbf{\textcolor{comp_blue}{computational structure}}, and \textbf{\textcolor{surf_green}{syntax}}.}
    \label{fig:decomposition_serialization}
    \vspace{-15pt}
\end{figure}

\textbf{Contributions.} In this work, we study robustness and generalization of LLMs for graph reasoning. To this end, we make the following individual contributions:
\begin{itemize}[itemsep=-3pt, topsep=0pt]
    \item We propose a principled decomposition of graph serializations into \textbf{{\color{lab_orange}node labeling}}, \textbf{{\color{comp_blue}computational structure}}, and \textbf{{\color{surf_green}syntax}} (see \cref{fig:decomposition_serialization}).
    \item Using the \Erdos benchmark~\citep{guo2025g1}, we conduct a large-scale evaluation of how graph serialization affects LLM graph reasoning. We compare post-trained graph reasoners (\GOne, trained with supervised fine-tuning and reinforcement learning) against non-fine-tuned open-weight LLMs (\Qwen~\citep{qwen}, \GPT~\citep{gptoss}), totaling \textbf{over 1.5M individual model inferences}.
    \item We introduce new spectral tasks to assess if graph reasoners generalize beyond purely combinatorial tasks.
\end{itemize}

We find that post-training reduces relabeling sensitivity, but increases sensitivity to computational structure and syntax, indicating specialization to the serialization present in the data. Importantly, our analysis suggests that in some cases, post-training can reinforce non-structural shortcuts that graph-native models would not observe. Thus, we argue that progress in LLM graph reasoning should not be judged solely from fixed-format accuracy, which motivates future work on invariance-aware post-training and benchmark design for LLM graph reasoners.

\section{Related Work}

\textbf{Graphs, isomorphisms and inductive biases.} 
Since pioneering works in the early 2000s~\citep{gori2005anew,scarselli2009computational}, Graph Neural Networks (GNNs) have attracted significant attention in the Deep Learning community in recent years~\citep{defferrard2016fast,kipf2017semi,veličković2018graph,gilmer2017neural}.
This area has focused on a fundamental question: ``\emph{How to best design neural networks that operate on graph-structured data?}''.
The overarching principle advocated by Geometric Deep Learning is that of constraining these architectures to be in-/equivariant to the main symmetry of graphs, i.e., node relabelings~\citep{bronstein2021geometric}. Permutation invariance, the central inductive bias of GNNs, ensures functions depend solely on the isomorphism class of the input graph, rather than any arbitrary labeling choice in their encodings.

Intriguingly, enforcing this inductive bias on Graph Neural Networks has profound consequences on their expressive power, as these architectures can be universal approximators only when parameterized with an intractable computational complexity~\citep{maron2019universality}. 
Importantly, Message-Passing Neural Networks~\citep{gilmer2017neural}, the most common and computationally efficient GNN variants, can even fail at straightforward graph tasks such as counting the number of substructures such as cycles~\citep{chen2020can}.

\textbf{LLMs on graphs and graph reasoning.}
Upon the advent of LLMs, researchers have started to investigate their applicability to more structured objects such as graphs. Not only have they started to explore hybrid methods integrating LLMs and GNNs~\citep{he2024gretriever,yang2024glfusion,liu2023oneforall}, but, also, the pure application of LLMs directly operating on graph textual encodings~\citep{zhao2023graphtext,fatemi2023talk}.
This approach is particularly attractive because, by appropriate prompting in natural language, a single pretrained model can assist in interactive graph explorations, mining and zero-shot property predictions.

Within this trend of applying LLMs on graph-structured data, an emerging line of work specifically focuses on \emph{graph reasoning}, which gathers tasks pertaining to answering queries or computing graph- or node-level parameters, such as shortest paths, connectivity, subgraph counting, and other structural inference problems.
These tasks are of interest because they are conceptual representatives of general reasoning problems with dependencies~\citep{besta2024graph,sanford2024understanding}, and because they constitute relevant benchmarks to study LLM capabilities in capturing structural patterns beyond the limits of standard GNNs~\citep{morris2019weisfeiler,xu2019how,zhang2024beyond,zhang2023rethinking}.

\textbf{Evaluating and improving LLMs for graph reasoning.}
A first evaluation of LLMs for graph reasoning has been presented in~\citep{wang2023nlgraph}, followed by a more systematic approach by~\citet{fatemi2023talk} and later evaluations on newly released LLMs in~\citep{tang2025grapharena,yuan2025gracore}. 
The investigated models showed preliminary but still inadequate capabilities.
Initial approaches include instruction tuning~\citep{luo2024graphinstruct,ye2024language,tang2025grapharena} preference optimization~\citep{chen2024graphwiz}, as well as supervised fine-tuning~\citep{sanford2024understanding,guo2025g1}.
These techniques can indeed boost the performance of LLMs as graph reasoners, but were shown to be outperformed by the more recent approach of~\citet{guo2025g1}, whereby supervised fine-tuning is followed by a second stage operating the Group Relative Policy Optimization (GRPO) reinforcement learning algorithm.
The authors release \GOne, a graph reasoner obtained by fine-tuning small-sized \Qwen LLMs~\citep{qwen} on a curated corpus of reasoning tasks called \Erdos.
\GOne represents a leading purpose-built effort on graph reasoning, and \Erdos stands as the most diverse post-training and benchmarking suite.

\textbf{LLM sensitivity to textual graph encodings.}
Contrary to GNNs, LLMs graph reasoners are not intrinsically in-/equivariant to graph symmetries or representations.
A first study on this aspect was conducted in~\citep{fatemi2023talk} over PaLM models~\citep{palm} and a small set of simple reasoning tasks.
\citet{ge2025graph} later performed further analyses on edge-list representations, focusing on the impact of edge orderings for various LLMs on five reasoning tasks.
The authors observed a pronounced preference for Breadth- (BFS) and Depth-First-Search (DFS) orderings compared to random ones, which consistently induced performance degradations.
BFS ordering was found to yield  better results on more local tasks, whereas DFS was found more competitive in tasks requiring a more global graph comprehension.
Similar observations were made in~\citet{yuan2025gracore}, where lexicographic ordering was found to generally outperform random ones on a small set of tasks.
In contrast to these analyses, this present work encompasses a much larger set of reasoning tasks, considers multiple sources of variations in textual graph encodings and, importantly, also extends to post-trained LLMs.
To the best of our knowledge, fine-tuned LLM graph reasoners are examined in their sensitivity only in~\citep{chu2025graphsos}, but the analyses are limited to edge reorderings on a small number of tasks.

\section{LLMs on Graphs and Graph Serializations}

\subsection{Processing Graphs with LLMs}

A graph is a tuple $\graph = (\nodes, \edges)$, consisting of a finite node set $\nodes$ of size $n$ and an edge set $\edges \subseteq V^2$, which can (optionally) be assigned a real-valued weight.
Graphs must be practically stored \emph{sequentially} for any kind of meaningful computation.
This requires fixing a node labeling (ordering) $\lab: \nodes \bij [n], [n] := \{1, \dots, n\}$.
Clearly, the relational structure encoded by $\graph$ does not depend on (the arbitrarily chosen) $\lab$.
This motivates the central inductive bias of GNNs, i.e., in-/equivariance to such node relabelings.
GNNs process graphs encoded algebraically as adjacency matrices $\mA$ and are inherently constrained to be permutation in-/equivariant:
\begin{align}
    \mathrm{GNN}_{\mathrm{graph}}(\mP \mA \mP^\top) \, &=\, \mathrm{GNN}_{\mathrm{graph}}(\mA), \\
    \mathrm{GNN}_{\mathrm{node}}(\mP \mA \mP^\top) \, &=\, \mP \,\mathrm{GNN}_{\mathrm{node}}(\mA),
\end{align}
for any permutation $\mP$, where $\mathrm{GNN}_{\mathrm{graph}}(\mA) \in \R^d$ outputs a global graph-level, or $\mathrm{GNN}_{\mathrm{node}}(\mA) \in \R^{n \times d}$ a node-level prediction.

Instead, LLMs operate on graphs encoded as \emph{sequences} of tokens $\ser(\graph, \lab) := (s_1, \dots, s_{Q_\text{graph}})$.
They take as input both this and a task description $\task := (t_1, \dots, t_{Q_\text{task}})$ to generate an answer
\begin{equation}
    \mathrm{LLM}(\task, \ser(\graph, \lab)) \::=\: (y_1, \dots, y_A).
\end{equation}
The answer tokens are iteratively sampled or greedily chosen from the autoregressive distribution defined by the parameters $\theta$ of the LLM's inner decode-only transformer:
\begin{equation}
    y_t \sim p_\theta(\cdot | \task, \ser(\graph, \lab), y_{<t}).
\end{equation}
The generation proceeds until an end-of-sequence token is produced or the maximum output length is reached.
Different from GNNs, a single LLM can tackle (virtually) any reasoning problem by prompt-conditioning via $\task$, a key advantage. However, LLMs are not permutation in-/equivariant by design.
Distinct labelings $\lab$ and serializations $\ser$ of the same graph yield distinct input strings for which LLMs may yield distinct outputs.

A sufficiently large network could, in principle, \emph{approximate} any (continuous) permutation invariant function, but limitations stem from computational and sample complexity and, practically, the characteristics of the employed graph dataset.
The network may, e.g., learn to exploit \emph{positional} or \emph{formatting regularities} in the available samples rather than capturing structural invariants.
Consider, for instance, learning to compute a graph-level (e.g., maximum degree) or node-level (e.g., the maximum-degree node) parameter.
If $(\lab, \ser)$ is confounded with the target, clues about the target can get directly leaked in the input representation.
As an degenerate example, if the distinguished node $v^\ast$ for a node-level task is always assigned the index $\lab(v^\ast) = n$, the target is trivially recoverable just from positional information, i.e., by reading out the parameters on the ``last'' node.
Biases towards particular positional and formatting regularities may affect pretrained LLMs and, as we will demonstrate through experiments, can be amplified by post-training.

Next, we introduce a principled decomposition of graph serializations to more systematically study and understand the impact of the above aspects.

\subsection{Decomposition of Serializations}\label{sec:decomposition_of_serializations}

Besides the labeling, we propose factoring serialization into: a \emph{data structure}, an \emph{internal edge-ordering rule}, and a \emph{surface encoding} (see~\cref{fig:decomposition_serialization}). 
We detail these components next.

\textbf{{\color{lab_orange}Node labeling}} determines how the nodes of a given graph are indexed.
As mentioned before, any labeling is arbitrary, and a main desideratum for a graph reasoner is to be invariant to these (for graph-level targets). 
In principle, node labeling invariance would follow automatically once a target algorithm is correctly implemented.
However, in practice, deviations may occur due to partial competence.
On the other hand, as capacity and data grow, variability across relabelings may already shrink, even before achieving exact algorithmic execution.
Either way, we crucially note that, absent potential positional spillage in $(\lab, \ser)$ and strong positional biases, node relabelings do not change the computational workload presented to the model, in terms of its algorithmic execution.
In principle, under these conditions, no systematic trends across $\lab$ are thus to be expected.

\textbf{{\color{comp_blue}Computational structure}} gathers the \textbf{data structure} and \textbf{internal edge ordering} rule.
The former determines how the relational structure is described, the latter how this is presented as a sequence.
Prominent choices for the data structure would be \opt{edge-list}, \opt{adjacency-list} or \opt{adjacency-matrix} (see~\cref{app:example_prompts}).
Each structure comes with its own symmetries, e.g., any edge ordering in an \opt{edge-list} equivalently represents the same graph, or both node and neighbor indices can be arbitrarily permuted in an \opt{adjacency-list}.
An ordering breaks these symmetries by presenting the data structure in an LLM-compatible sequential form.
Note that the chosen computational structure may significantly impact the model's algorithmic execution by determining its computational path:
e.g., with a \emph{sorted} adjacency list, neighborhoods are presented contiguously in localized blocks.
Many tasks such as BFS-like traversals or degree statistics are solvable by a single linear pass over them.
On the contrary, with a \emph{shuffled} edge list, a correct procedure must traverse the entire edge list for each neighborhood lookup or, alternatively, first build a canonical form to supporting linear scans, bringing about additional costs.
Hence, unlike node relabelings, one would expect that changes in the computational structural can induce \emph{persistent} performance and compute gaps.

\textbf{{\color{surf_green}Syntax}} captures variations in textual style, independently from the graph's structural representation.
An ideal reasoner abstracts away from such choices; real models may exhibit biases from pretraining, e.g., bias on code execution may lead to better performance when the input is presented as code.
Tokenization length or granularity may also induce side effects.
These are \emph{textual} rather than \emph{computational} biases but can still affect accuracy and efficiency.
Examples of surface encodings include edge list in plain text, in \opt{json} format, or as Python code in \opt{NetworkX} or \opt{PyG}.

LLM graph reasoners may be sensitive to variations in each of the above encoding components. In the following section, we empirically analyze and discuss this relevant aspect.

\section{Experimental Analyses}\label{sec:exp}

\paragraph{Models.}
In our experimental analysis, we evaluated a diverse set of LLMs with varying architectures, sizes, and training paradigms to thoroughly investigate graph reasoning capabilities.
Our model selection includes the following:
\begin{itemize}[itemsep=-3pt, topsep=0pt]
    \item \Qwen \textbf{(3B, 7B)} are open-weight models developed by Alibaba Cloud\footnote{\url{https://huggingface.co/collections/Qwen/qwen25-66e81a666513e518adb90d9e}}, which serve as our non-fine-tuned baseline. These models have been pre-trained on a large corpus of text data but have not been specifically optimized for graph reasoning tasks.
    \item \GOne \textbf{(3B, 7B)} are specialized graph reasoning models obtained by fine-tuning the \Qwen base models using a two-stage approach \citep{guo2025g1}. First, supervised fine-tuning was performed on the \Erdos dataset, followed by reinforcement learning using Group Relative Policy Optimization. These models represent the current state of the art in LLM-based graph reasoning.
    \item \GPT \textbf{(20B, 120B)} are open-weight models \citep{gptoss} with significantly larger parameter counts. These models provide a contemporary perspective on scaling effects in capabilities without dedicated post-training for graph reasoning, serving as a comparison point to the smaller specialized models.
\end{itemize}

This selection allows us to analyze the impact of model scale (3B to 120B parameters) and specialization on graph reasoning performance and robustness to representational variations. For a deterministic comparison in our sensitivity experiments, we set the inference temperature of all \Qwen, \GOne and \GPT models to $0$, ensuring identical outputs for identical inputs. Accuracy differences compared to evaluations at $0.06$ (as in \cite{guo2025g1}) are in \Cref{app:setup}.

\paragraph{Datasets.} 
We run comprehensive experiments on the \Erdos test set~\citep{guo2025g1}, encompassing $49$ topological tasks over $100$ graphs each, totaling $4{,}900$ samples\footnote{From the original suite, we only exclude \texttt{isomorphic\_mapping} due to a different input structure.}.
Across all ablations, shufflings, and models, this amounts to analyzing \textbf{over 1.5M individual model inferences}.
As in \Erdos, we group tasks by difficulty (Easy, Medium, Hard, Challenging).
We further introduce a novel benchmark of $12$ spectral tasks on $100$ graphs from the \Erdos test set (see \cref{sec:spectral}).

\subsection{Sensitivity to {\color{lab_orange} Node Labeling}}

\begin{figure}[h]
    \centering
    \includegraphics[width=0.9\linewidth]{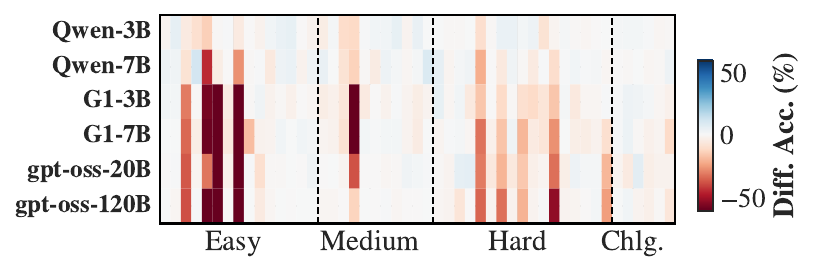}
    \captionsetup{margin=0.3cm}
    \caption{\textbf{Acc.\ diff.} w/ relabeled nodes vs.\ baseline \Erdos.}
    \label{fig:node_relabeling_acc}
\end{figure}
\begin{figure}[h]
    \centering
    \includegraphics[width=\linewidth]{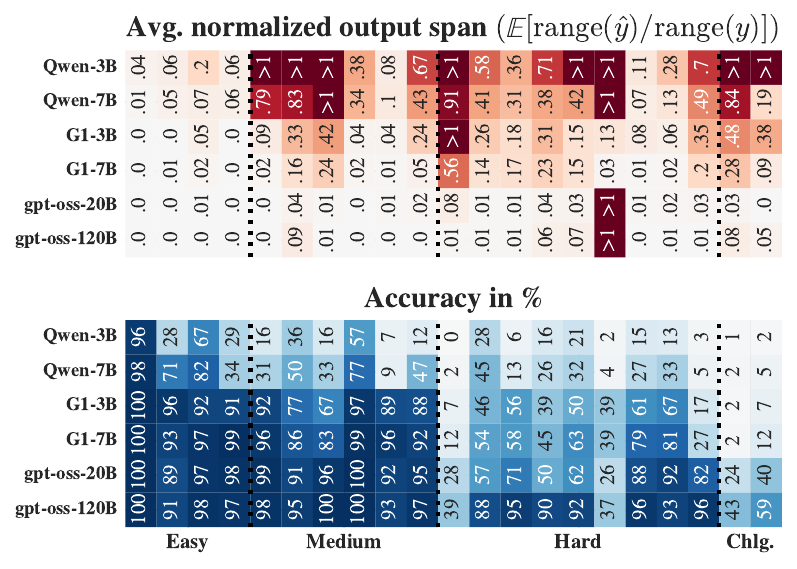}
    \captionsetup{margin=0.3cm}
    \caption{Average \textbf{per-example output spreads} normalized by per-task answer range (top), resp.\ accuracy (bottom).}
    \label{fig:invariance_node_relabeling}
\end{figure}

\begin{figure*}[t]
    \centering
    \begin{subfigure}[t]{0.323\textwidth}
        \includegraphics[width=\linewidth]{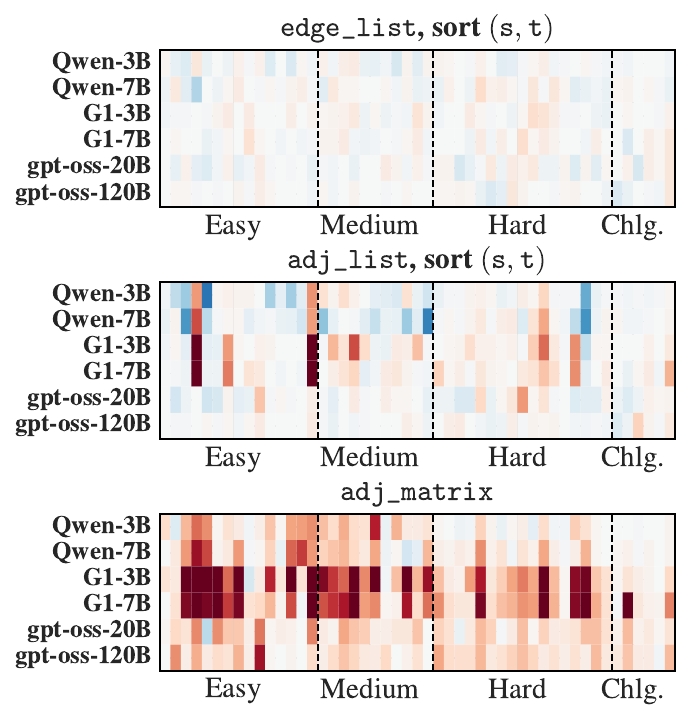}
        \captionsetup{margin=0.1cm}
    \caption{Different \textbf{sorted} comp.\ structures vs.\ baseline \Erdos.}
    \label{fig:edge_reordering_comp_structure}
    \end{subfigure}
    \begin{subfigure}[t]{0.29\textwidth}
        \includegraphics[width=\linewidth]{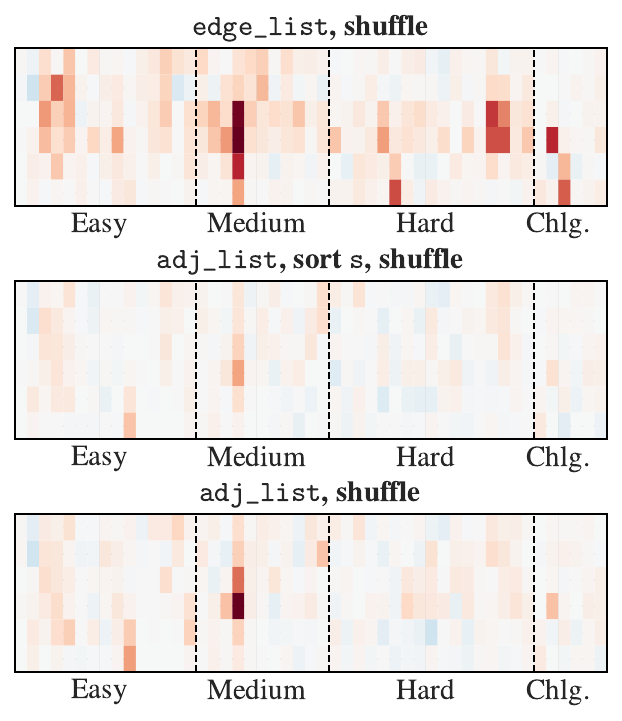}
    \captionsetup{margin=0.1cm}
    \caption{\textbf{Shuffled} comp.\ structures vs.\ baseline resp.\ sorted.}
    \label{fig:edge_reordering_shufflings}
    \end{subfigure}
    \begin{subfigure}[t]{0.36\textwidth}
        \includegraphics[width=\linewidth]{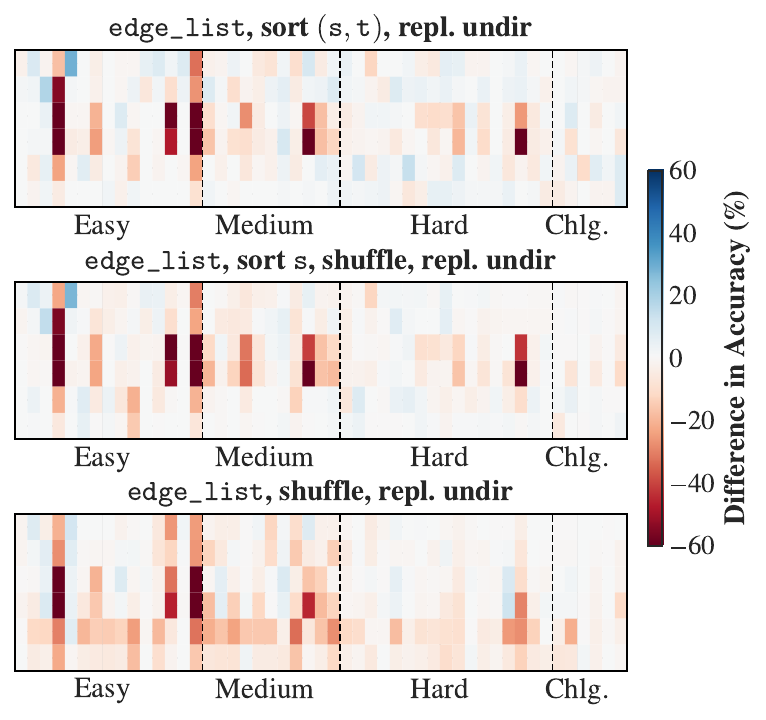}
    \captionsetup{margin=0.1cm}
    \caption{W/ \textbf{replicating undir.\ edges} vs.\ baseline w/o.}
    \label{fig:edge_reordering_repundir}
    \end{subfigure}
    \captionsetup{margin=2cm}
    \caption{Ablation of \emph{(a)} structure, \emph{(b)} reordering, and \emph{(c)} replicating undirected edges. Each colored cell corresponds to one out of the 49 \Erdos problems we analyze (columns), evaluated on one out of the 6 models considered (rows), and, thus, aggregates the results of 100 (resp.\ 1000, as 10 shuffles per individual prompt) individual model inferences for deterministic (resp.\ shuffled) comp.\ structures. For full heatmaps with individual task labels, see \cref{app:comp_struct_all_results}.}
    \label{fig:edge_reordering_heatmaps}
\end{figure*}

We quantify the effect of node permutations by evaluating LLMs across multiple random relabelings of the same graph.
Specifically, we draw $N = 10$ random node labelings and apply them on the \Erdos test graphs\footnote{After relabeling, we lexicographically sort edges to factor out the impact of shuffled edge-list~\citep{yuan2025gracore}.}, and then assess: \textbf{(a)} the \textbf{mean accuracy} over these relabelings; \textbf{(b)} the \textbf{variability of the models' outputs} on \emph{numerical} tasks.

As for \textbf{(a)}, \cref{fig:node_relabeling_acc} reports the accuracy difference between the original \Erdos dataset and sampled node relabelings across tasks.
We observe clear sensitivity to the relabeling, which is more pronounced for \GOne than for \Qwen.
This can often be explained by reliance on positional regularities: certain labelings make the target solution easier to \enquote{heuristically guess}.
For example, in \texttt{bipartite\_maximum\_matching}, the correct solution in \Erdos always induces a \emph{contiguous} bipartition, i.e., $(\{1,\dots,x\},\{x+1,\dots,n\})$.
Fine-tuning can internalize this, but after relabeling the ground-truth bipartition ceases to be contiguous and performance drops.

As for \textbf{(b)}, we capture output variability induced by node relabelings by calculating the normalized output span averaged over all test graphs in \emph{numerical} tasks ($0.0$ indicates no variability, as desired for invariance).
These values are reported in~\cref{fig:invariance_node_relabeling} (top).
We observe that the \Qwen models generally exhibit higher sensitivity, and that this typically correlates with task performance (bottom).
These results indicate that graph-oriented fine-tuning can indeed make LLM graph reasoners more \enquote{invariant} to node permutations, but this is mainly explained by improved reasoning performance rather than the general emergence of functional invariance.
Some exceptions are found, however, in the \enquote{Challenging} category, where fine-tuning lowers output variability \emph{even without} producing a meaningful accuracy increase.

\subsection{Sensitivity to {\color{comp_blue}Computational Structure}}

Here, we study robustness w.r.t.\ the data structure and its inner ordering. We consider three standard structures: \texttt{edge-list}, \texttt{adjacency-list}, and \texttt{adjacency-matrix}.
As pointed out in \cref{sec:decomposition_of_serializations}, each of these structures come with their \emph{own} symmetries.
Further, for undirected graphs, edges are sets of size 2, adding the symmetry that the endpoint order is irrelevant; we study this effect as well.
Next we discuss general trends, for full numerical results see \cref{fig:edge_reorderings_ablation_heatmap}, cf. \cref{app:comp_struct_all_results}.

We first evaluate the impact of the comp.\ structure itself, presented, when applicable, in an canonicalized form obtained by full lexicographic sort.
In \cref{fig:edge_reordering_comp_structure}, we plot accuracy differences for \emph{fully sorted} \texttt{edge-list}, \texttt{adjacency-list}, and \texttt{adjacency-matrix} encodings relative to the default \Erdos\footnote{The \Erdos dataset uses an edge-list representation where edges are ordered by a BFS from node~1 over a randomly sampled subgraph.}.
We observe sorted \opt{edge-lists} stay close to the baseline; \opt{adjacency-lists} introduce shifts on some tasks.
The largest drops occur on \texttt{edge\_number} and \texttt{density}, both edge-counting tasks that naturally favor edge lists.
The fine-tuned \GOne models are the most brittle to these changes in the data structure, \Qwen sometimes even improves performance, and \GPT remains fairly steady.
As for \opt{adjacency-matrix} encodings, they often incur sizable losses.
We hypothesize this is due to these representations being \enquote{misaligned} with counting tasks and inflating the description length to $\mathcal{O}(n^2)$, adding potentially distracting tokens.

We now probe robustness w.r.t.\ the \emph{inherent symmetries} of the \texttt{edge-list} and \texttt{adjacency-list} encodings.
In \cref{fig:edge_reordering_shufflings}, we report accuracy differences obtained by a (partly) \emph{shuffled} data structure as opposed to its canonicalized form.
The fine-tuned \GOne models are observed, again, to be the most brittle on average, with the largest drops occurring on \texttt{edge-list}.
In sorted \texttt{edge-list}, the presented edges are grouped by source node, so each node’s neighbors appear contiguously.
We hypothesize this \emph{locality}, i.e., neighbors appearing in \enquote{blocks}, facilitates retrieval to the model.
Random \opt{edge-list} shuffles break this locality: retrieving a node’s neighborhood requires a scan of the entire list, or a first internal reordering to allow linear scans, and both these operations can both introduce errors.
As for \texttt{adjacency-lists}, the aforementioned locality remains preserved under all of its symmetries, and consistently, we instead observe shuffles hit comparatively mildly.

Last, we examine replicating undirected edges in \texttt{edge-lists} enforcing \emph{full symmetric locality}: when edges are not replicated in a sorted listing, not all neighborhoods need to be continuous. 
Accuracy differences are reported in \cref{fig:edge_reordering_repundir}; again \GOne's are most brittle and edge-counting tasks are hit hardest.
This suggests, while replication resolves endpoint symmetries and aligns with edge extraction, it also inflates the sequence and can dilute signal.

\subsection{Sensitivity to \textcolor{surf_green}{Syntax}}

\begin{figure}[h]
    \centering
    \includegraphics[width=\linewidth]{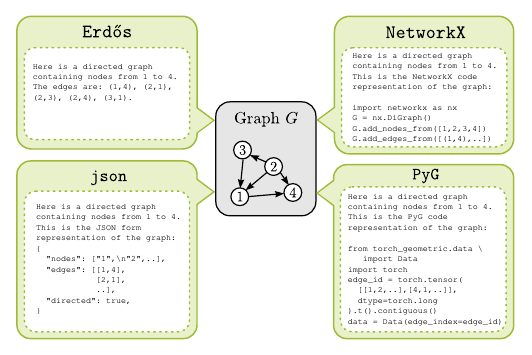}
    \caption{Examples of textual graph encodings.}
    \label{fig:encoding_examples}
\end{figure}
We now investigate robustness under varying \emph{syntax}: we fix the data structure to \texttt{edge-list} and re-encode the \Erdos test graphs by each of the following encodings: \opt{json}, \opt{NetworkX}, and \opt{PyG}.
\opt{json} is a common text-based serialization for storage and data exchange; \opt{NetworkX} and \opt{PyG} follow code-based conventions from graph processing libraries (cf.\ \cref{fig:encoding_examples}).
We choose these formats as widely represented in text and code corpora likely used for LLM pretraining, thereby potentially better activating the models' reasoning abilities.

\begin{figure}[t]
    \centering
    \includegraphics[width=\linewidth]{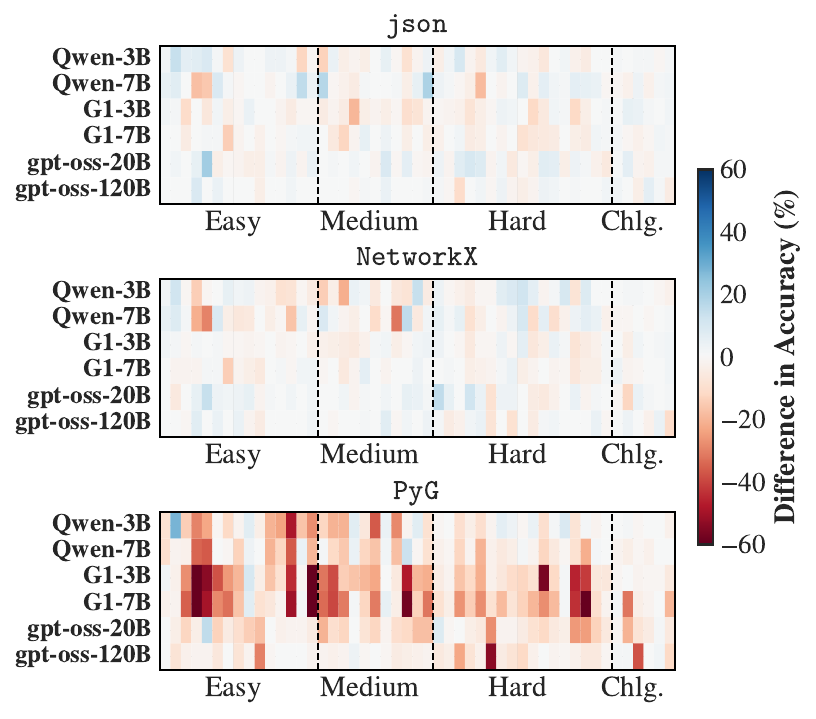}
    \caption{Different \textbf{syntax} vs.\ baseline \Erdos.}
    \label{fig:surface_encoding_acc}
\end{figure}

\begin{figure}[t]
    \centering
    \includegraphics[width=\linewidth]{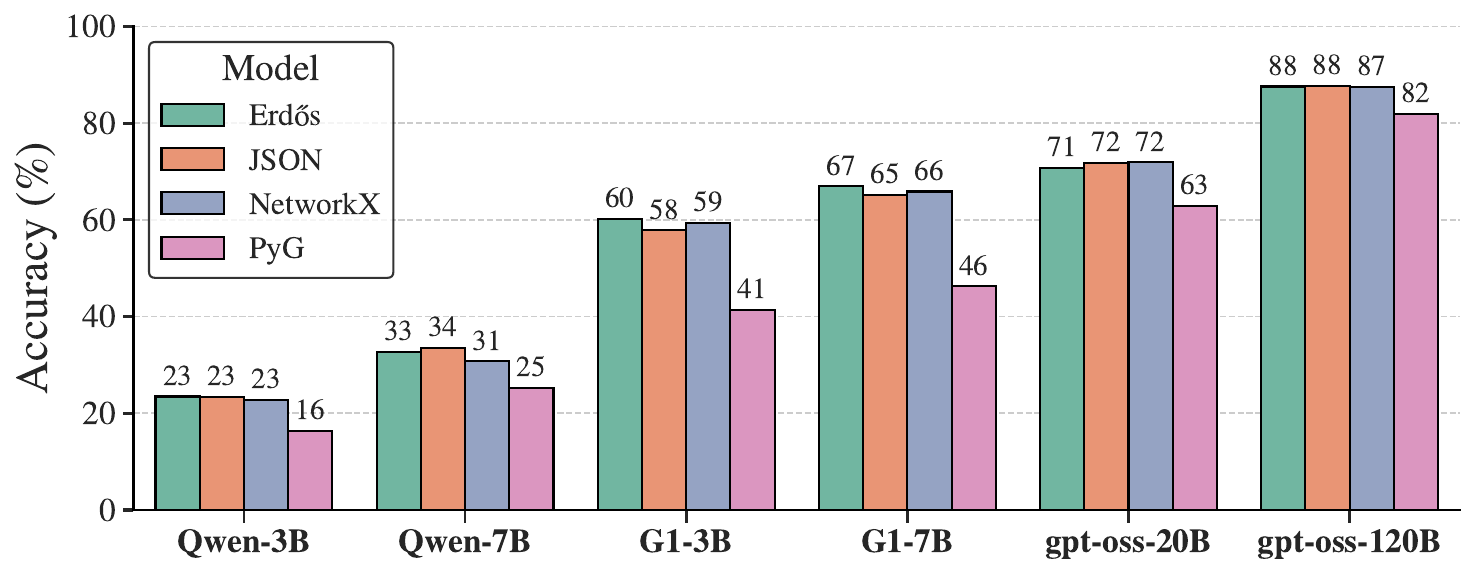}
    \caption{Average model accuracy by \textbf{syntax}.}
    \label{fig:encoding_models}
\end{figure}

We evaluate \GOne, \Qwen and \GPT on the newly encoded graph instances and study the performance impact of these formats, also compared to the \enquote{standard} \Erdos encoding.
Results are all reported aggregated in \cref{fig:encoding_models} and in full in \cref{fig:encode_heatmap}.
As expected, \Erdos remains, on average, the highest-performing encoding across tasks for both \GOne's (-7B: best in $\nicefrac{28}{49}$ tasks).
As for the non-fine-tuned models, \Qwen-7B shows, instead, a preference towards the \opt{json} format ($34\%$ accuracy, best in \nicefrac{24}{49} tasks).
Both \opt{json} and \opt{NetworkX} formats give the best performance on \GPT-20B, while the larger -120B variant exhibits robustness: barring \opt{PyG}, all encodings work equally well ($88\%$ accuracy all).
We interestingly remark the general underwhelming performance of \opt{PyG}, which we comment on in \cref{app:surf_encoding_additional_results}.

\begin{figure*}[t] 
    \centering
    \includegraphics[width=0.85\linewidth]{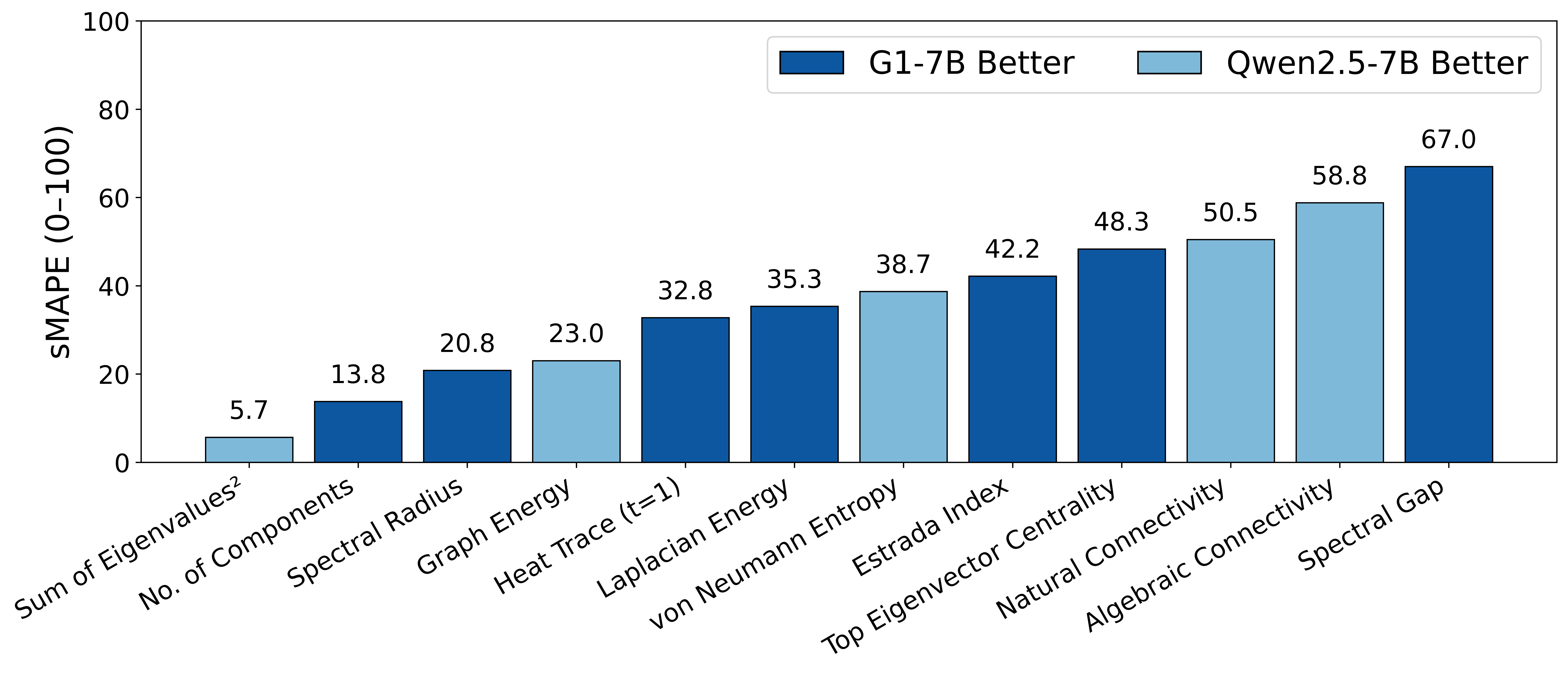}
    \caption{Performance is oscillating between the two models, there is no general consistent winner.}
    \label{fig:spectral_winners}
\end{figure*}

We also measure the average per-task variability in terms of standard deviation over the performance obtained by the four considered encodings.
Consistent with the above, \GPT models attain the lowest variability ($5.41\%$ for -20B, $3.45\%$ for -120B), followed by \Qwen's ($5.98\%$ for -3B, $5.86\%$ for -7B) and \GOne's ($9.69\%$ for -3B, $10.51\%$ for -7B).
However, we note that the largest variability of \GOne is mostly attributed to \opt{PyG}.

\section{Testing Generalization on Spectral Tasks}\label{sec:spectral}

\textbf{A Spectral Graph Reasoning benchmark.} Spectral quantities are fundamental tools to analyze robustness and diffusion in real-world networks~\citep{jamakovic2008robustness,van2023graph,matzner2024topology,ahuja2025topology}, yet  graph reasoning has almost exclusively focused on topological or combinatorial tasks~\citep{tang2025grapharena,yuan2025gracore,guo2025g1}.
Moreover, prior work has primarily emphasized quantitative evaluation; here, we complement it with a qualitative analysis of the actual generated reasoning.
For this, we construct a novel suite of 12 spectral tasks.

\begin{table}[h]
\centering
\caption{Spectral graph-theoretic tasks by difficulty.}
\label{tab:spectral-tasks-appendix}
\resizebox{0.85\linewidth}{!}{
\begin{tabular}{l p{0.72\linewidth}}
\toprule
\textbf{Difficulty} & \textbf{Tasks} \\
\midrule
\textbf{Easy} & Graph Energy, Number of Connected Components, Sum of Squared Eigenvalues. \\
\textbf{Medium }& Algebraic Connectivity, Estrada Index, Laplacian Energy,  Natural Connectivity, Spectral Gap, Spectral Radius. \\
\textbf{Hard} & Top Eigenvector Centrality, Heat Trace at $t=1$, von Neumann Entropy. \\
\bottomrule
\label{tab:spectral_task_classification_easy_medium_hard}
\end{tabular}}
\end{table}

We instantiate these tasks on a sample of $100$ graphs from the \Erdos test set and cluster them by difficulty (Easy, Medium, Hard), see \cref{tab:spectral_task_classification_easy_medium_hard}.
On these real-valued tasks, we consider the well interpretable Symmetric Mean Absolute Percentage Error (sMAPE), as well as Relative Mean  Absolute Error (RelMAE).
For both, lower is better.
sMAPE is bounded in $[0,100]$; as for RelMAE, robust to near-zero values, a score below $1.0$ indicates the model outperforms the mean baseline.
Finally, we complement the evaluation with automated and manual reasoning-style classification of answers (\emph{analytical}, \emph{heuristic}, \emph{incomplete}).

\begin{table}[h] 
    \centering
    \captionsetup{margin=0.7cm}
    \caption{Average performance on spectral tasks by difficulty (sMAPE$_{0\text{--}100}$, $\downarrow$).}
    \label{tab:spectral-tasks}
    \resizebox{0.7\linewidth}{!}{%
        \begin{tabular}{l c c c}
        \toprule
        \textbf{Model} & \textbf{Easy} & \textbf{Medium} & \textbf{Hard} \\
        \midrule
        Qwen2.5-3B & 35.01 & 53.89 & 56.60 \\
        Qwen2.5-7B & \textbf{19.73} & 50.05 & 51.31 \\
        G1-3B      & 30.33 & 55.74 & 59.14 \\
        G1-7B      & 20.80 & \textbf{49.58} & \textbf{39.95} \\
        \bottomrule
        \end{tabular}
    }
\end{table}

\textbf{Results} are reported aggregated in~\cref{tab:spectral-tasks}, in full in~\cref{app:spectral_results}.
\GOne-7B yields modest gains compared to \Qwen-7B on harder spectral problems (e.g., \opt{eigenvector centrality}), yet it outperforms \Qwen on only $7/12$ tasks, indicating no universal winner (\cref{fig:spectral_winners}). 
While fine-tuning with \GOne-7B improves sMAPE performance on harder problems (e.g., \opt{eigenvector\_centrality}, \opt{heat\_trace}, \cref{fig:qwen_g1_task_wise_smape}, cf.\ \cref{app:spectral_results}), on these tasks it remains not better than the mean baseline, as RelMAE is larger than 1.0 (cf.\ \cref{tab:spectral_results_on_all_models_all_metrics}, cf.\ \cref{app:spectral_results}).
Overall, we observe \GOne-7B outperforms the mean baseline in 50\% of the tasks, but the gains are still limited (RelMAE in $[0.6,0.9]$). 
Manual inspection on $10\%$ of the generated answers from \GOne-7B show an interesting pattern where it first approaches the task by analytical derivations (e.g., computation of degrees, normalized adj. matrices), but reverts to heuristics as complexity increases (e.g., known properties, assumptions of intermediate values, approximations based on structure).

\section{Conclusion}

We presented an evaluation of LLMs graph reasoners under symmetry transformations. Our framework decomposes serializations into \emph{(i)} \textbf{\textcolor{lab_orange}{node labeling}}, \emph{(ii)} \textbf{\textcolor{comp_blue}{computational structure}}, and \emph{(iii)} \textbf{\textcolor{surf_green}{syntax}}. We also test \emph{(iv)} generalization on novel spectral tasks. Our experiments reveal the following key takeaways:
\begin{itemize}[itemsep=-3pt, topsep=0pt]
    \item In our analysis, \textbf{post-training can reduce sensitivity to node labeling}, although mostly reflecting performance enhancement rather than true invariance.
    \item The \textbf{serialization dictates computation paths} taken by the model, \textbf{altering algorithmic difficulty} and impacting performance.
    \item LLMs are not format-invariant, with specialized reasoners \textbf{excelling mostly on formats seen in post-training}.
    \item On spectral tasks, neither base nor specialized models generalize reliably.
    \item Overall, larger models appear more robust to the above variations, but are much costlier (cf. \cref{app:runtime}), and fine-tuning of smaller models does not consistently improve generalization or invariance, as sensitivity tends to be exacerbated by focus on datasets idiosyncrasies.
\end{itemize}

Based on these findings, we argue that progress in LLM graph reasoning should not be judged from fixed-format accuracy alone, and both post-training and evaluation should account for robustness across equivalent graph representations. An interesting future direction could be to design invariance-aware post-training and benchmarking pipelines for LLM graph reasoners.

\section*{Limitations}

Our study focuses on LLM-based graph reasoners using textual serializations. They are comprehensive across node labeling, edge encoding, and syntax, and cover small and large scale models, as well an example of a (leading) fine-tuned reasoner. We could, however, cover a broader set of LLM models and other previously proposed tuning strategies. Additionally, we could perform similar analyses on real-world node- and graph-property prediction tasks.

\section*{Societal Impact}

LLM graph reasoners have potential applications in knowledge discovery, network analysis, and scientific computation. Misinterpretation of outputs due to sensitivity to serializations could propagate errors in critical domain. In this sense, our research can educate the broader community of researchers and practitioners on the limits and brittleness of this general methodological approach, fostering the development of more reliable graph reasoners. On the other hand, we note that shedding light on this sensitivity could cue malicious actors into developing targeted exploits for adversarial attacks.

\section*{Acknowledgments}

We would like to thank all people involved in the 2025 London Geometry and Machine Learning Summer School (LOGML), where this research project started. In particular, we would like to express our gratitude to the members of the organizing committee: Vincenzo Marco De Luca, Massimiliano Esposito, Simone Foti, Valentina Giunchiglia, Daniel Platt, Pragya Singh, Arne Wolf, and Zhengang Zhong. DH would like to thank Yifei Wang and Stefanie Jegelka for helpful discussions. FF is very grateful to Haggai Maron and Giorgos Bouritsas for general support and constructive early exchanges in project scoping. FF is also extremely thankful to the members of the ``Eva Project'', whose support he immensely appreciates. DH acknowledges funding by the Alexander von Humboldt Foundation, the Munich Center for Machine Learning (MCML), and the Munich Data Science Institute (MDSI) via the MDSI Doctoral Fellowship. FF conducted this work supported by an Aly Kaufman Post-Doctoral Fellowship.

\bibliography{icml2026}
\bibliographystyle{icml2026}

\newpage
\appendix
\onecolumn

\section{Compute Resources \& Serving Stack}\label{app:resources}
All models are served using \texttt{vLLM}, a high-throughput LLM serving system based on \emph{Paged Attention} for efficient KV-cache management\footnote{\url{https://github.com/vllm-project/vllm}}. Unless otherwise specified, the evaluation was run on NVIDIA H100  80GB accelators with identical decoding and batching settings: temperature $0$ (deterministic), fixed maximum output lengths and matched prompt budgets across models and tasks. For \GPT, we use the default \texttt{medium} reasoning effort. For \GPT-120, we additionally use a multi-GPU configuration (2$\times$H100 80GB) with tensor parallelism.

\section{Inference Runtime vs Accuracy Trade-offs}\label{app:runtime}
Across all graph reasoning tasks, \GPT models attain the highest accuracy and robustness, consistently outperforming \Qwen (3B, 7B) and \GOne (3B, 7B). The gains come with increase in the latency per request when served via \texttt{vLLM} under matched scheduling and decoding policies. In \cref{tab:latency_tradeoff} report wall-clock latency relative to the 3B-7B class on a single H100 GPU.

\begin{table}[h]
\centering
\small
\captionsetup{margin=1cm}
\caption{Inference performance comparison across model sizes and configurations. Results averaged over 100 prompts using single or dual H100 80GB GPUs.}
\label{tab:latency_tradeoff}
\begin{tabular}{l c l l}
\toprule
\textbf{Model} & \textbf{GPUs (H100 80GB)} & \textbf{Latency (s)} & \textbf{Throughput (tokens/s)} \\
\midrule
\Qwen-3B   & 1 & 0.531  & 1,976.13 \\
\Qwen-7B   & 1 & 0.919  & 1,377.60 \\
\GOne-3B   & 1 & 0.765  & 5,349.55 \\
\GOne-7B   & 1 & 1.207  & 3,392.40 \\
\GPT-20B   & 1 & 5.457  & 2,013.14 \\
\GPT-120B  & 1 & 31.740 & 361.14 \\
\GPT-120B  & 2 & 7.103  & 1,459.86 \\
\bottomrule
\end{tabular}
\end{table}
These factors isolate the impact of model size and computational reasoning on response time by keeping the serving infrastructure (\texttt{vLLM} scheduler, KV-cache paging) and decoding settings constant. In practice, the decision involves balancing accuracy against latency: \GPT delivers state-of-the-art accuracy but requires longer inference time. While multi-GPU deployment helps offset some of the slowdown, it introduces additional costs.

\section{Example Prompts}\label{app:example_prompts}

This section illustrates, at the prompt level, our encodings under node relabelings, the computational structures, and the surface encodings. As a running example throughout this section, consider the following \opt{shortest\_path} prompt. For all list-style encodings, the right side includes a heatmap of node indices: $1$ appears lightest and $19$ darkest.

\begin{tcolorbox}[
  title={Example prompt  \opt{shortest\_path}, \Erdos encoding},
  sidebyside,
  sidebyside align=center seam,
  righthand width=0.06\textwidth,
  sidebyside gap=6mm,
  boxrule=1.5pt,
]
{\color{gray}The task is to determine the shortest path between two nodes.

The input nodes are guaranteed to be connected.

Here is an undirected graph containing nodes from 1 to 19.} \textbf{The edges are: (1, 7), (1, 12), (1, 6), (1, 3), (1, 2), (7, 3), (7, 6), (7, 12), (12, 3), (6, 17), (6, 9), (3, 2), (4, 5), (4, 8), (4, 10), (4, 11), (5, 15), (5, 16), (5, 8), (5, 10), (5, 13), (5, 11), (5, 14), (8, 10), (10, 11), (10, 14), (11, 16), (11, 13), (16, 18), (13, 18), (17, 9), (17, 19), (9, 19).}

{\color{gray}Question: What is the shortest path between node 12 and node 19?

You need to format your answer as a list of nodes, e.g., [node-1, node-2, ..., node-n].}
\tcblower
\vspace{-8pt}
\includegraphics[width=\linewidth]{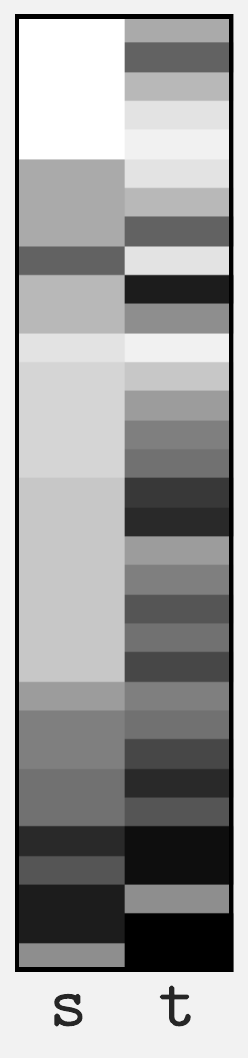}
\vspace{-20pt}
\end{tcolorbox}

\subsection{Node Relabeling ($\lab$)}

\begin{tcolorbox}[
  title={Random node relabeling},
  sidebyside,
  sidebyside align=center seam,
  righthand width=0.06\textwidth,
  sidebyside gap=6mm,
  boxrule=1.5pt,
]
\textcolor{gray}{\ldots}

\textcolor{gray}{Here is an undirected graph containing nodes from 1 to 19.}
\textcolor{lab_orange}{\textbf{The edges are: (1, 6), (1, 10), (1, 16), (2, 7), (2, 18), (3, 14), (3, 19), (4, 18), (5, 7), (5, 13), (5, 18), (6, 8), (6, 10), (6, 16), (7, 13), (7, 15), (7, 18), (8, 10), (9, 11), (9, 12), (9, 17), (10, 16), (10, 17), (11, 12), (11, 17), (13, 15), (13, 18), (14, 15), (14, 18), (15, 18), (15, 19), (16, 17), (18, 19).}}

\textcolor{gray}{\ldots}

\tcblower
\includegraphics[width=\linewidth]{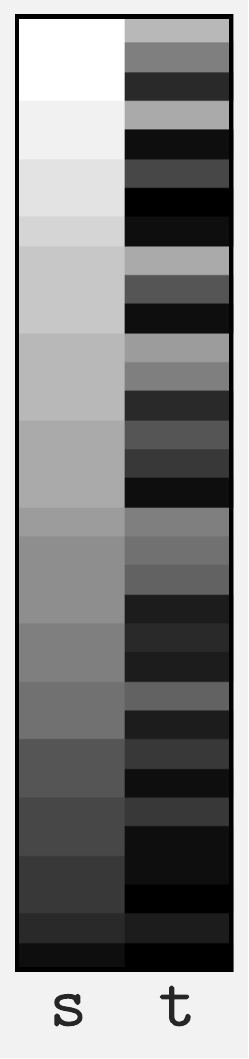}
\end{tcolorbox}

\subsection{Computational Structure}

\begin{tcolorbox}[
  title={$\texttt{edge\_list}$, sorting key =  $\texttt{(source,target)}$},
  sidebyside,
  sidebyside align=center seam,
  righthand width=0.06\textwidth,
  sidebyside gap=6mm,
  boxrule=1.5pt,
]
\textcolor{gray}{...}

\textcolor{gray}{Here is an undirected graph containing nodes from 1 to 19.} \textcolor{comp_blue}{\textbf{The edges are: (1, 2), (1, 3), (1, 6), (1, 7), (1, 12), (2, 3), (3, 7), (3, 12), (4, 5), (4, 8), (4, 10), (4, 11), (5, 8), (5, 10), (5, 11), (5, 13), (5, 14), (5, 15), (5, 16), (6, 7), (6, 9), (6, 17), (7, 12), (8, 10), (9, 17), (9, 19), (10, 11), (10, 14), (11, 13), (11, 16), (13, 18), (16, 18), (17, 19).}}

\textcolor{gray}{...}
\tcblower
\includegraphics[width=\linewidth]{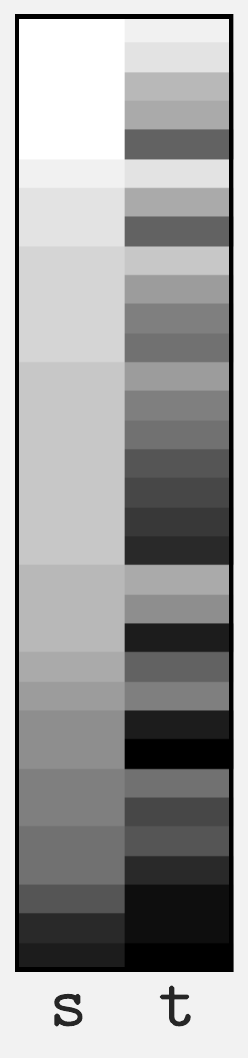}
\end{tcolorbox}

\begin{tcolorbox}[
  title={$\texttt{edge\_list}$, sorting key =  $\texttt{(source,target)}$, replicate undirected edges},
  sidebyside,
  sidebyside align=center seam,
  righthand width=0.06\textwidth,
  sidebyside gap=6mm,
  boxrule=1.5pt,
]
\textcolor{gray}{...}

\textcolor{gray}{Here is an undirected graph containing nodes from 1 to 19.} \textcolor{comp_blue}{\textbf{The edges are (each undirected edge is listed in both directions): (1, 2), (1, 3), (1, 6), (1, 7), (1, 12), (2, 1), (2, 3), (3, 1), (3, 2), (3, 7), (3, 12), (4, 5), (4, 8), (4, 10), (4, 11), (5, 4), (5, 8), (5, 10), (5, 11), (5, 13), (5, 14), (5, 15), (5, 16), (6, 1), (6, 7), (6, 9), (6, 17), (7, 1), (7, 3), (7, 6), (7, 12), (8, 4), (8, 5), (8, 10), (9, 6), (9, 17), (9, 19), (10, 4), (10, 5), (10, 8), (10, 11), (10, 14), (11, 4), (11, 5), (11, 10), (11, 13), (11, 16), (12, 1), (12, 3), (12, 7), (13, 5), (13, 11), (13, 18), (14, 5), (14, 10), (15, 5), (16, 5), (16, 11), (16, 18), (17, 6), (17, 9), (17, 19), (18, 13), (18, 16), (19, 9), (19, 17).}}

\textcolor{gray}{...}
\tcblower
\includegraphics[width=\linewidth]{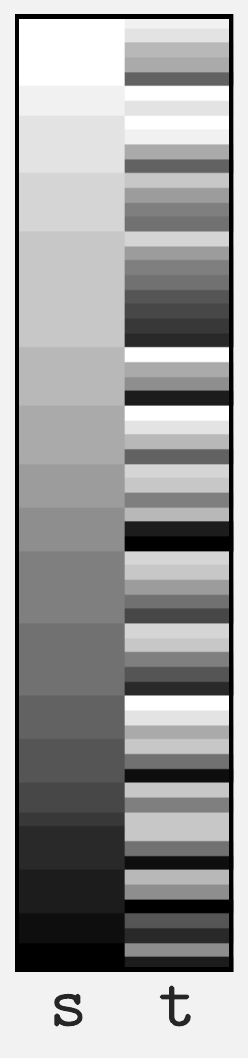}
\end{tcolorbox}

\begin{tcolorbox}[
  title={$\texttt{edge\_list}$, sorting key =  $\texttt{source}$, shuffle $\texttt{target}$},
  sidebyside,
  sidebyside align=center seam,
  righthand width=0.06\textwidth,
  sidebyside gap=6mm,
  boxrule=1.5pt,
]
\textcolor{gray}{...}

\textcolor{gray}{Here is an undirected graph containing nodes from 1 to 19.} \textcolor{comp_blue}{\textbf{The edges are: (1, 3), (1, 2), (1, 12), (1, 7), (1, 6), (2, 3), (3, 12), (3, 7), (4, 8), (4, 10), (4, 11), (4, 5), (5, 11), (5, 13), (5, 15), (5, 8), (5, 14), (5, 10), (5, 16), (6, 9), (6, 7), (6, 17), (7, 12), (8, 10), (9, 17), (9, 19), (10, 11), (10, 14), (11, 13), (11, 16), (13, 18), (16, 18), (17, 19).}}

\textcolor{gray}{...}
\tcblower
\includegraphics[width=\linewidth]{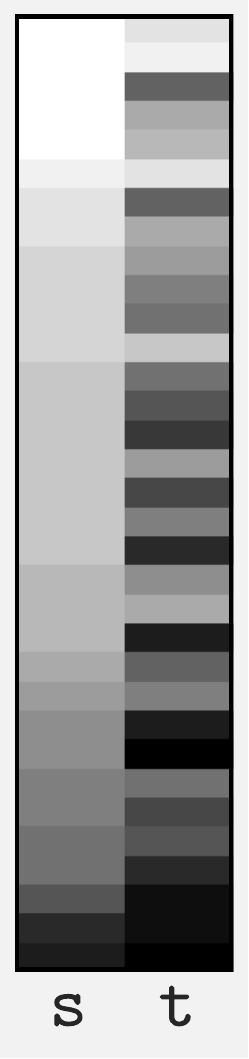}
\end{tcolorbox}

\begin{tcolorbox}[
  title={$\texttt{edge\_list}$, sorting key $\texttt{source}$, shuffle $\texttt{target}$, replicate undirected edges},
  sidebyside,
  sidebyside align=center seam,
  righthand width=0.06\textwidth,
  sidebyside gap=6mm,
  boxrule=1.5pt,
]
\textcolor{gray}{...}

\textcolor{gray}{Here is an undirected graph containing nodes from 1 to 19.} \textcolor{comp_blue}{\textbf{The edges are (each undirected edge is listed in both directions): (1, 3), (1, 2), (1, 12), (1, 7), (1, 6), (2, 3), (2, 1), (3, 1), (3, 7), (3, 12), (3, 2), (4, 10), (4, 5), (4, 11), (4, 8), (5, 14), (5, 4), (5, 11), (5, 16), (5, 8), (5, 10), (5, 15), (5, 13), (6, 7), (6, 1), (6, 17), (6, 9), (7, 1), (7, 12), (7, 6), (7, 3), (8, 10), (8, 5), (8, 4), (9, 6), (9, 19), (9, 17), (10, 5), (10, 4), (10, 11), (10, 8), (10, 14), (11, 10), (11, 4), (11, 5), (11, 16), (11, 13), (12, 3), (12, 1), (12, 7), (13, 5), (13, 11), (13, 18), (14, 5), (14, 10), (15, 5), (16, 18), (16, 5), (16, 11), (17, 9), (17, 6), (17, 19), (18, 13), (18, 16), (19, 17), (19, 9).}}

\textcolor{gray}{...}
\tcblower
\includegraphics[width=\linewidth]{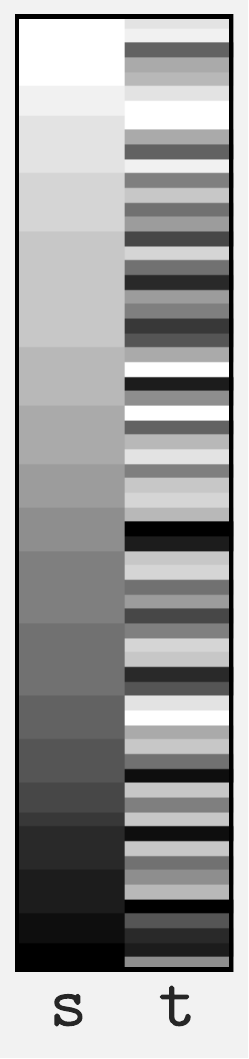}
\end{tcolorbox}

\begin{tcolorbox}[
  title={$\texttt{edge\_list}$, shuffle entire list},
  sidebyside,
  sidebyside align=center seam,
  righthand width=0.06\textwidth,
  sidebyside gap=6mm,
  boxrule=1.5pt,
]
\textcolor{gray}{...}

\textcolor{gray}{Here is an undirected graph containing nodes from 1 to 19.} \textcolor{comp_blue}{\textbf{The edges are: (4, 11), (1, 2), (9, 6), (11, 10), (14, 5), (11, 5), (11, 16), (15, 5), (4, 10), (5, 10), (5, 16), (14, 10), (17, 19), (17, 6), (8, 5), (5, 13), (9, 19), (16, 18), (8, 4), (1, 12), (9, 17), (11, 13), (4, 5), (7, 12), (7, 1), (1, 3), (6, 1), (6, 7), (3, 2), (12, 3), (13, 18), (3, 7), (8, 10).}}

\textcolor{gray}{...}
\tcblower
\includegraphics[width=\linewidth]{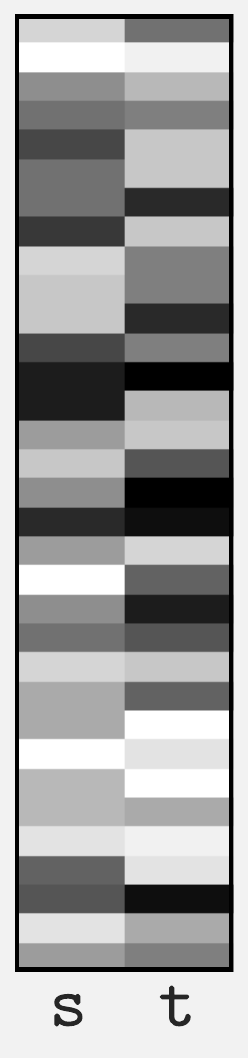}
\end{tcolorbox}

\begin{tcolorbox}[
  title={$\texttt{edge\_list}$, shuffle entire list, replicate undirected edges},
  sidebyside,
  sidebyside align=center seam,
  righthand width=0.06\textwidth,
  sidebyside gap=6mm,
  boxrule=1.5pt,
]
\textcolor{gray}{...}

\textcolor{gray}{Here is an undirected graph containing nodes from 1 to 19.} \textcolor{comp_blue}{\textbf{The edges are (each undirected edge is listed in both directions): (8, 5), (7, 1), (6, 17), (16, 11), (16, 18), (11, 5), (3, 7), (6, 17), (3, 7), (9, 19), (6, 1), (10, 5), (10, 4), (11, 16), (10, 14), (13, 18), (13, 18), (17, 9), (2, 3), (3, 1), (4, 10), (7, 6), (19, 17), (5, 10), (11, 4), (12, 7), (5, 4), (1, 7), (6, 1), (5, 15), (7, 12), (16, 5), (12, 3), (1, 12), (17, 19), (8, 10), (3, 2), (5, 8), (11, 5), (9, 17), (3, 12), (10, 14), (6, 9), (5, 4), (5, 13), (1, 2), (16, 5), (11, 10), (14, 5), (13, 11), (1, 3), (5, 15), (18, 16), (11, 10), (13, 5), (11, 4), (12, 1), (14, 5), (6, 7), (1, 2), (11, 13), (4, 8), (8, 10), (19, 9), (6, 9), (4, 8).}}

\textcolor{gray}{...}
\tcblower
\includegraphics[width=\linewidth]{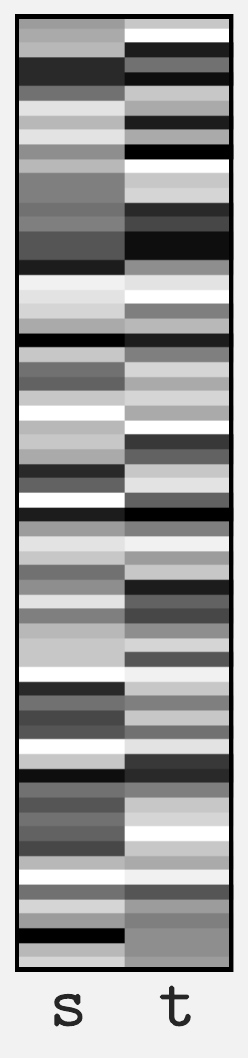}
\end{tcolorbox}

\begin{tcolorbox}[
  title={$\texttt{adj\_list}$, sorting key =  $\texttt{(source,target)}$},
  sidebyside,
  sidebyside align=center seam,
  righthand width=0.06\textwidth,
  sidebyside gap=6mm,
  boxrule=1.5pt,
]
\textcolor{gray}{...}

\textcolor{gray}{Here is an undirected graph containing nodes from 1 to 19.} \textcolor{comp_blue}{\textbf{The adjacency list is: 
\begin{itemize}[topsep=0pt,itemsep=0pt,parsep=0pt,leftmargin=*]
    \item node 1 is connected to (2, 3, 6, 7, 12),
    \item node 2 is connected to (1, 3),
    \item node 3 is connected to (1, 2, 7, 12),
    \item node 4 is connected to (5, 8, 10, 11),
    \item node 5 is connected to (4, 8, 10, 11, 13, 14, 15, 16), ...
\end{itemize}}}
\textcolor{gray}{...}
\tcblower
\includegraphics[width=\linewidth]{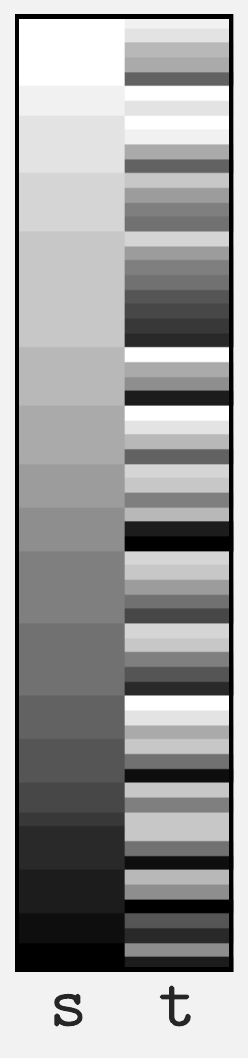}
\end{tcolorbox}

\begin{tcolorbox}[
  title={$\texttt{adj\_list}$, sorting key =  $\texttt{source}$, shuffle $\texttt{target}$},
  sidebyside,
  sidebyside align=center seam,
  righthand width=0.06\textwidth,
  sidebyside gap=6mm,
  boxrule=1.5pt,
]
\textcolor{gray}{...}

\textcolor{gray}{Here is an undirected graph containing nodes from 1 to 19.} \textcolor{comp_blue}{\textbf{The adjacency list is:
\begin{itemize}[topsep=0pt,itemsep=0pt,parsep=0pt,leftmargin=*]
\item node 1 is connected to (12, 6, 7, 3, 2),
\item node 2 is connected to (1, 3),
\item node 3 is connected to (7, 1, 2, 12),
\item node 4 is connected to (5, 11, 8, 10),
\item node 5 is connected to (8, 15, 13, 10, 14, 16, 11, 4), ...
\end{itemize}}}

\textcolor{gray}{...}
\tcblower
\includegraphics[width=\linewidth]{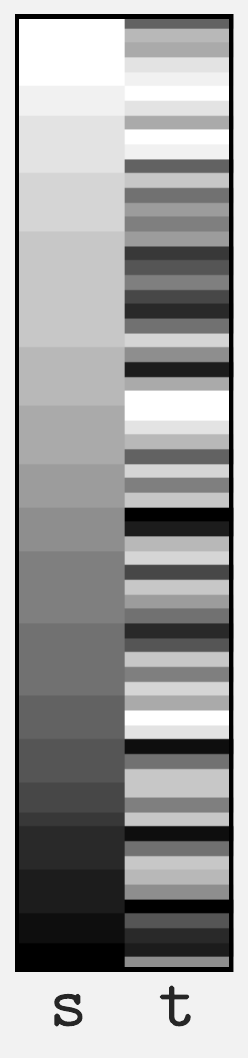}
\end{tcolorbox}

\begin{tcolorbox}[
  title={$\texttt{adj\_list}$, sorting key = $\texttt{target}$, shuffle $\texttt{source}$},
  sidebyside,
  sidebyside align=center seam,
  righthand width=0.06\textwidth,
  sidebyside gap=6mm,
  boxrule=1.5pt,
]
\textcolor{gray}{...}

\textcolor{gray}{Here is an undirected graph containing nodes from 1 to 19.} \textcolor{comp_blue}{\textbf{The adjacency list is:
\begin{itemize}[topsep=0pt,itemsep=0pt,parsep=0pt,leftmargin=*]
\item node 4 is connected to (5, 8, 10, 11),
\item node 13 is connected to (5, 11, 18),
\item node 18 is connected to (13, 16),
\item node 10 is connected to (4, 5, 8, 11, 14),
\item node 19 is connected to (9, 17), ...
\end{itemize}}}

\textcolor{gray}{...}
\tcblower
\includegraphics[width=\linewidth]{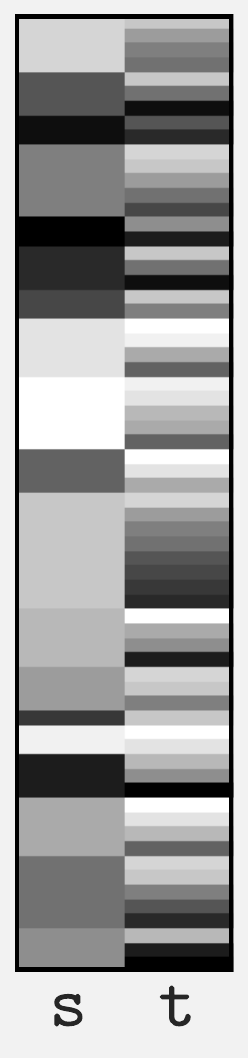}
\end{tcolorbox}

\begin{tcolorbox}[
  title={$\texttt{adj\_list}$, shuffle all},
  sidebyside,
  sidebyside align=center seam,
  righthand width=0.06\textwidth,
  sidebyside gap=6mm,
  boxrule=1.5pt,
]
\textcolor{gray}{...}

\textcolor{comp_blue}{\textbf{\textcolor{gray}{Here is an undirected graph containing nodes from 1 to 19.} \textcolor{comp_blue}{\textbf{The adjacency list is:
\begin{itemize}[topsep=0pt,itemsep=0pt,parsep=0pt,leftmargin=*]
\item node 9 is connected to (17, 19, 6),
\item node 16 is connected to (18, 11, 5),
\item node 18 is connected to (16, 13),
\item node 8 is connected to (4, 5, 10),
\item node 7 is connected to (6, 1, 3, 12), ...
\end{itemize}}}}}

\textcolor{gray}{...}
\tcblower
\includegraphics[width=\linewidth]{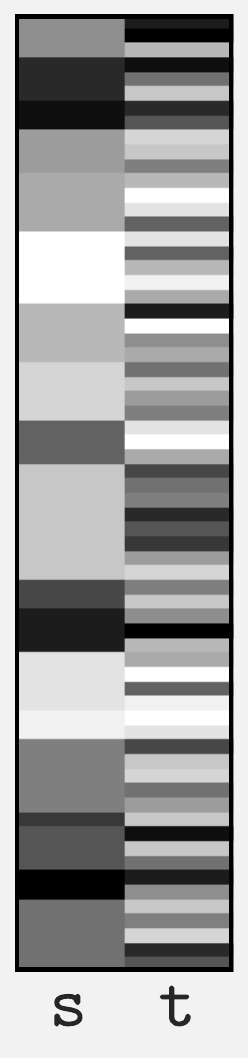}
\end{tcolorbox}

\begin{tcolorbox}[title={\texttt{adj\_matrix}}]
\textcolor{gray}{...}

\textcolor{comp_blue}{Here is an undirected graph containing nodes from 1 to 19.} \textcolor{comp_blue}{\textbf{This is the binary adjacency matrix representation of the graph where 1 denotes an edge between nodes:}}

\textcolor{comp_blue}{[[0, 1, 1, 0, 0, 1, 1, 0, 0, 0, 0, 1, 0, 0, 0, 0, 0, 0, 0],} \\
\vspace{1pt}
\textcolor{comp_blue}{[1, 0, 1, 0, 0, 0, 0, 0, 0, 0, 0, 0, 0, 0, 0, 0, 0, 0, 0],} \\
\vspace{1pt}
\textcolor{comp_blue}{[1, 1, 0, 0, 0, 0, 1, 0, 0, 0, 0, 1, 0, 0, 0, 0, 0, 0, 0],} \\
\vspace{1pt}
\textcolor{comp_blue}{[0, 0, 0, 0, 1, 0, 0, 1, 0, 1, 1, 0, 0, 0, 0, 0, 0, 0, 0],} \\
\vspace{1pt}
\textcolor{comp_blue}{[0, 0, 0, 1, 0, 0, 0, 1, 0, 1, 1, 0, 1, 1, 1, 1, 0, 0, 0],}\\
\textcolor{comp_blue}{...]}

\textcolor{gray}{...}
\end{tcolorbox}

\subsection{Surface Encoding}\label{sec:surf_examples}

\begin{tcolorbox}[title={\opt{json}}]
\textcolor{gray}{...}

{\color{surf_green}\textbf{Here is a undirected graph containing nodes from 1 to 19. This is the JSON form representation of the graph:}\\
\{\\
"nodes": [    "1", "2",    "3",    "4",    "5",    "6",    "7",    "8",    "9",    "10",    "11",    "12",    "13",    "14",    "15",    "16",    "17",    "18",    "19"  ], \\
"edges": [    [      1,      7    ],    [      1,      12    ],    [      1,      6    ],    [      1,      3    ],    [      1,      2    ],    [      7,      3    ],    [      7,      6    ],    [      7,      12    ],    [      12,      3    ],    [      6,      17    ],    [      6,      9    ],    [      3,      2    ],    [      4,      5    ],    [      4,      8    ],    [      4,      10    ],    [      4,      11    ],    [      5,      15    ],    [      5,      16    ],    [      5,      8    ],    [      5,      10    ],    [      5,      13    ],    [      5,      11    ],    [      5,      14    ],    [      8,      10    ],    [      10,      11    ],    [      10,      14    ],    [      11,      16    ],    [      11,      13    ],    [      16,      18    ],    [      13,      18    ],    [      17,      9    ],    [      17,      19    ],    [ 9,      19    ]  ], \\
"directed": false\\
\}}

\textcolor{gray}{...}
\end{tcolorbox}

\begin{tcolorbox}[title={\opt{NetworkX}}]
\textcolor{gray}{...}

{\color{surf_green} \textbf{Here is an undirected graph containing nodes from 1 to 19. This is the NetworkX code representation of the graph:}\\
import networkx as nx \\
G = nx.Graph() \\
G.add\_nodes\_from([1, 2, 3, 4, 5, 6, 7, 8, 9, 10, 11, 12, 13, 14, 15, 16, 17, 18, 19]) \\
G.add\_edges\_from([(1, 7), (1, 12), (1, 6), (1, 3), (1, 2), (7, 3), (7, 6), (7, 12), (12, 3), (6, 17), (6, 9), (3, 2), (4, 5), (4, 8), (4, 10), (4, 11), (5, 15), (5, 16), (5, 8), (5, 10), (5, 13), (5, 11), (5, 14), (8, 10), (10, 11), (10, 14), (11, 16), (11, 13), (16, 18), (13, 18), (17, 9), (17, 19), (9, 19)])}

\textcolor{gray}{...}
\end{tcolorbox}

\begin{tcolorbox}[title={\opt{PyG}}]
\textcolor{gray}{...}

{\color{surf_green} \textbf{Here is a undirected graph containing nodes from 1 to 19. This is the PyG code representation of the graph:}\\
from torch\_geometric.data import Data \\
import torch \\
edge\_index = torch.tensor([[1, 7, 1, 12, 1, 6, 1, 3, 1, 2, 7, 3, 7, 6, 7, 12, 12, 3, 6, 17, 6, 9, 3, 2, 4, 5, 4, 8, 4, 10, 4, 11, 5, 15, 5, 16, 5, 8, 5, 10, 5, 13, 5, 11, 5, 14, 8, 10, 10, 11, 10, 14, 11, 16, 11, 13, 16, 18, 13, 18, 17, 9, 17, 19, 9, 19], [7, 1, 12, 1, 6, 1, 3, 1, 2, 1, 3, 7, 6, 7, 12, 7, 3, 12, 17, 6, 9, 6, 2, 3, 5, 4, 8, 4, 10, 4, 11, 4, 15, 5, 16, 5, 8, 5, 10, 5, 13, 5, 11, 5, 14, 5, 10, 8, 11, 10, 14, 10, 16, 11, 13, 11, 18, 16, 18, 13, 9, 17, 19, 17, 19, 9]], dtype=torch.long).t().contiguous() \\
data = Data(edge\_index=edge\_index)}

\textcolor{gray}{...}
\end{tcolorbox}

\section{Temperature Setting}\label{app:setup}
To ensure a fair and deterministic comparison across our experiments, we set the inference temperature of both the \GOne and \Qwen models to $0$. A temperature of $0$ eliminates randomness in the models' outputs, guaranteeing that identical inputs always yield identical responses.

\begin{figure}[htbp]
    \centering
\includegraphics[width=1\linewidth]{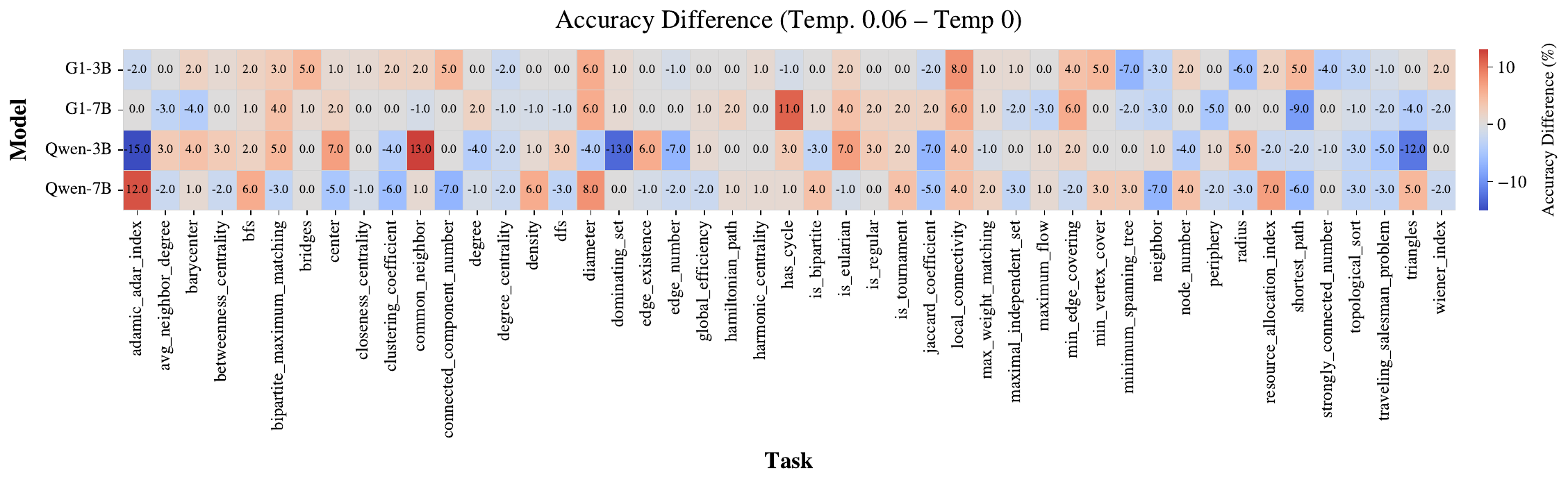}
\captionsetup{margin=1cm}
\caption{Accuracy difference between \GOne and \Qwen evaluated at temperature $0.06$ (as reported by \cite{guo2025g1}) and the deterministic version at temperature $0$ used in our analysis.}
\label{fig:temp_comparison}
\end{figure}

\Cref{fig:temp_comparison} illustrates the accuracy differences between evaluations performed at temperature $0.06$, as reported by \citet{guo2025g1}, and our deterministic evaluations at temperature $0$. To maintain consistency, we exclude the task \opt{isomorphic\_mapping}, following our main analysis, and the tasks \opt{weighted\_minimum\_spanning\_tree} and \opt{weighted\_shortest\_path}, as results for them are not reported in \citet{guo2025g1}. Overall, we observe only marginal differences in accuracy between the two temperature settings. Specifically, \GOne-3B achieves $62.53\,\%$ at temperature $0$ compared to $63.21\,\%$ at $0.06$, while \GOne-7B reaches $69.13\,\%$ and $69.36\,\%$, respectively. For \Qwen, the 3B model scores $24.34\,\%$ at temperature $0$ and $24.09\,\%$ at $0.06$, whereas the 7B models both achieve $33.96\,\%$. Although these overall differences are small, we note that some individual tasks exhibit larger variations, for example \opt{dominating\_set} and \opt{common\_neighbor} in \Qwen-3B, and \opt{connected\_component\_number} in both \Qwen-3B and \Qwen-7B, where discrepancies of up to $15\,\%$ occur.

\section{Additional Results}\label{app:additional_results}

\subsection{Edge Encoding}\label{app:comp_struct_all_results}

\begin{figure}
    \centering
    \includegraphics[width=\linewidth]{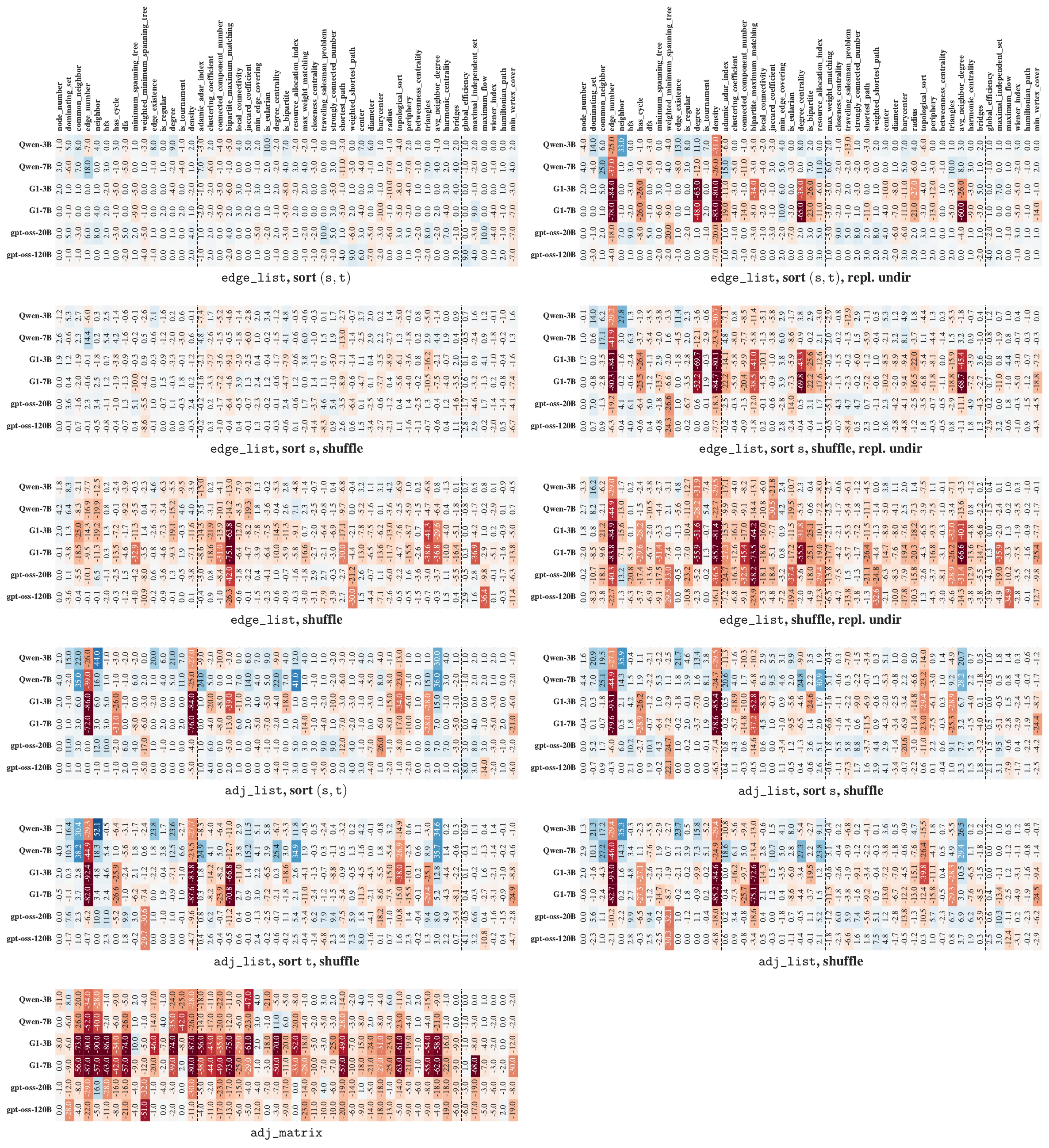}
    \captionsetup{margin=1cm}
    \caption{Heatmaps for the absolute difference in $\%$ between \Erdos baseline and corresponding datasets with different graph computational structures and orderings. For sorting, \texttt{s} = source and \texttt{t} = target, and the entry represents the sorting key. Shuffling is applied to all remaining ambiguities.}
    \label{fig:edge_reorderings_ablation_heatmap}
\end{figure}

We report per-task accuracies for the 3B models in \cref{tab:edge_reorderings_all_3B}, for the 7B models in \cref{tab:edge_reorderings_all_7B}, and for the \GPT models in \cref{tab:gpt_oss_edge_reordering}. Further, we visualize the per-task results in \cref{fig:edge_reorderings_ablation_heatmap}.

\begin{sidewaystable}
    \small
    \centering
    \captionsetup{margin=1cm}
    \caption{Edge reordering ablation results for the 3B models. We report the accuracy in \%. \Erdos baseline (w/ zero temperature) is shaded. For the computational structure, \texttt{edge} = edge list, \texttt{adjl} = adjacency list, \texttt{adjm} = adjacency matrix. For sorting, \texttt{s} = source and \texttt{t} = target, and the entry represents the sorting key. Shuffling is applied to all remaining ambiguities; in this case we report the mean accuracy $\pm$ std over 10 seeds.}
    \resizebox{\linewidth}{!}{
        {\renewcommand{\arraystretch}{1.2}
\begin{tabular}{ll >{\columncolor{gray!20}}rrrrrrrrrrrr : >{\columncolor{gray!20}}rrrrrrrrrrrr}

& & \multicolumn{12}{c:}{\large\textbf{Qwen-3B}} & \multicolumn{12}{:c}{\large\textbf{G1-3B}} \\[2ex]
\toprule
& \textbf{Comp. structure} & \texttt{edge} & \texttt{edge} & \texttt{edge} & \texttt{edge} & \texttt{edge} & \texttt{edge} & \texttt{edge} & \texttt{adjl} & \texttt{adjl} & \texttt{adjl} & \texttt{adjl} & \texttt{adjm} & \texttt{edge} & \texttt{edge} & \texttt{edge} & \texttt{edge} & \texttt{edge} & \texttt{edge} & \texttt{edge} & \texttt{adjl} & \texttt{adjl} & \texttt{adjl} & \texttt{adjl} & \texttt{adjm} \\
& \textbf{Sort (rest shuffled)} & \Erdos & \texttt{(s,t)} & \texttt{(s,t)} & \texttt{s} & \texttt{s} & \xmark & \xmark & \texttt{(s,t)} & \texttt{s} & \texttt{t} & \xmark & \texttt{NA} & \Erdos & \texttt{(s,t)} & \texttt{(s,t)} & \texttt{s} & \texttt{s} & \xmark & \xmark & \texttt{(s,t)} & \texttt{s} & \texttt{t} & \xmark & \texttt{NA} \\
& \textbf{Repl. undir} & \xmark & \xmark & \cmark & \xmark & \cmark & \xmark & \cmark & (\cmark) & (\cmark) & (\cmark) & (\cmark) & (\cmark) & \xmark & \xmark & \cmark & \xmark & \cmark & \xmark & \cmark & (\cmark) & (\cmark) & (\cmark) & (\cmark) & (\cmark) \\
\midrule
\multirow{15}{*}{\rotatebox{90}{\textbf{Easy}}} & node\_number & $98.0$ & $97.0$ & $94.0$ & $96.8 \pm 2.3$ & $97.9 \pm 2.0$ & $96.2 \pm 1.3$ & $94.7 \pm 1.7$ & $100.0$ & $99.6 \pm 0.5$ & $99.1 \pm 0.6$ & $99.3 \pm 0.5$ & $87.0$ & $98.0$ & $100.0$ & $100.0$ & $99.9 \pm 0.3$ & $99.7 \pm 0.5$ & $100.0 \pm 0.0$ & $99.8 \pm 0.4$ & $100.0$ & $100.0 \pm 0.0$ & $100.0 \pm 0.0$ & $100.0 \pm 0.0$ & $90.0$ \\
& dominating\_set & $36.0$ & $41.0$ & $50.0$ & $41.3 \pm 4.2$ & $50.0 \pm 6.1$ & $44.3 \pm 4.0$ & $52.2 \pm 5.1$ & $51.0$ & $56.9 \pm 5.8$ & $52.4 \pm 4.3$ & $57.3 \pm 5.1$ & $44.0$ & $98.0$ & $99.0$ & $99.0$ & $99.2 \pm 0.8$ & $98.0 \pm 1.1$ & $99.3 \pm 0.7$ & $98.8 \pm 1.1$ & $97.0$ & $98.3 \pm 0.9$ & $98.3 \pm 0.8$ & $96.4 \pm 1.3$ & $92.0$ \\
& common\_neighbor & $31.0$ & $39.0$ & $37.0$ & $33.7 \pm 2.3$ & $37.2 \pm 4.4$ & $28.9 \pm 2.5$ & $24.8 \pm 3.4$ & $53.0$ & $50.5 \pm 4.6$ & $61.4 \pm 3.0$ & $48.2 \pm 4.6$ & $11.0$ & $89.0$ & $90.0$ & $94.0$ & $87.1 \pm 1.9$ & $85.5 \pm 2.7$ & $64.0 \pm 2.0$ & $67.8 \pm 3.4$ & $95.0$ & $87.2 \pm 1.8$ & $91.8 \pm 3.0$ & $84.7 \pm 2.2$ & $16.0$ \\
& edge\_number & $38.0$ & $31.0$ & $13.0$ & $32.0 \pm 2.9$ & $8.8 \pm 2.4$ & $30.3 \pm 2.8$ & $9.0 \pm 2.4$ & $12.0$ & $10.9 \pm 1.8$ & $8.7 \pm 1.7$ & $8.6 \pm 3.1$ & $4.0$ & $97.0$ & $97.0$ & $13.0$ & $96.9 \pm 1.3$ & $12.9 \pm 0.3$ & $82.7 \pm 1.8$ & $12.1 \pm 1.2$ & $11.0$ & $3.9 \pm 2.7$ & $4.6 \pm 1.5$ & $4.0 \pm 1.3$ & $7.0$ \\
& neighbor & $35.0$ & $39.0$ & $68.0$ & $35.3 \pm 2.1$ & $62.8 \pm 3.9$ & $22.5 \pm 3.9$ & $33.3 \pm 4.1$ & $79.0$ & $70.9 \pm 3.1$ & $87.1 \pm 3.8$ & $70.3 \pm 4.8$ & $7.0$ & $94.0$ & $95.0$ & $94.0$ & $92.2 \pm 0.9$ & $92.4 \pm 2.5$ & $74.8 \pm 3.0$ & $78.4 \pm 2.5$ & $100.0$ & $96.1 \pm 1.3$ & $98.8 \pm 1.2$ & $95.6 \pm 2.0$ & $4.0$ \\
& bfs & $1.0$ & $1.0$ & $1.0$ & $3.5 \pm 1.8$ & $2.3 \pm 0.8$ & $1.2 \pm 0.6$ & $1.2 \pm 1.2$ & $0.0$ & $0.6 \pm 0.7$ & $0.5 \pm 0.7$ & $0.7 \pm 0.7$ & $0.0$ & $93.0$ & $91.0$ & $90.0$ & $93.7 \pm 1.2$ & $90.6 \pm 1.7$ & $94.3 \pm 1.8$ & $93.1 \pm 2.6$ & $98.0$ & $97.2 \pm 1.5$ & $97.6 \pm 1.2$ & $97.8 \pm 1.5$ & $7.0$ \\
& has\_cycle & $48.0$ & $51.0$ & $47.0$ & $46.6 \pm 4.3$ & $46.1 \pm 3.9$ & $45.6 \pm 3.5$ & $45.6 \pm 4.5$ & $45.0$ & $49.1 \pm 3.9$ & $41.6 \pm 2.4$ & $45.7 \pm 4.1$ & $39.0$ & $90.0$ & $85.0$ & $64.0$ & $86.2 \pm 1.5$ & $63.6 \pm 0.7$ & $82.8 \pm 2.7$ & $61.8 \pm 1.3$ & $64.0$ & $63.8 \pm 1.8$ & $64.1 \pm 1.5$ & $62.9 \pm 1.9$ & $56.0$ \\
& dfs & $6.0$ & $3.0$ & $3.0$ & $5.4 \pm 1.6$ & $2.5 \pm 1.4$ & $2.1 \pm 1.4$ & $2.2 \pm 1.8$ & $4.0$ & $3.9 \pm 2.0$ & $2.9 \pm 1.5$ & $2.5 \pm 1.3$ & $1.0$ & $100.0$ & $100.0$ & $100.0$ & $99.1 \pm 0.6$ & $98.9 \pm 1.4$ & $98.3 \pm 0.8$ & $98.0 \pm 1.1$ & $99.0$ & $98.8 \pm 0.9$ & $97.6 \pm 1.3$ & $97.4 \pm 1.7$ & $26.0$ \\
& minimum\_spanning\_tree & $8.0$ & $9.0$ & $8.0$ & $7.9 \pm 2.8$ & $4.2 \pm 1.8$ & $7.7 \pm 3.3$ & $5.3 \pm 2.3$ & $6.0$ & $5.8 \pm 2.1$ & $6.3 \pm 1.9$ & $5.1 \pm 2.0$ & $10.0$ & $88.0$ & $83.0$ & $91.0$ & $87.7 \pm 2.3$ & $90.9 \pm 2.5$ & $76.7 \pm 4.1$ & $84.7 \pm 2.7$ & $88.0$ & $89.0 \pm 2.3$ & $90.1 \pm 3.0$ & $85.9 \pm 2.7$ & $98.0$ \\
& weighted\_minimum\_spanning\_tree & $4.0$ & $1.0$ & $0.0$ & $1.4 \pm 1.1$ & $0.7 \pm 0.8$ & $1.7 \pm 1.1$ & $0.9 \pm 1.0$ & $4.0$ & $1.5 \pm 1.0$ & $1.6 \pm 1.1$ & $1.3 \pm 0.8$ & $0.0$ & $5.0$ & $5.0$ & $4.0$ & $5.9 \pm 2.2$ & $3.0 \pm 1.2$ & $6.4 \pm 1.6$ & $5.3 \pm 1.8$ & $2.0$ & $2.1 \pm 1.2$ & $3.8 \pm 1.8$ & $3.1 \pm 1.6$ & $0.0$ \\
& edge\_existence & $74.0$ & $82.0$ & $87.0$ & $81.1 \pm 2.9$ & $85.4 \pm 2.8$ & $78.6 \pm 3.9$ & $78.8 \pm 3.4$ & $94.0$ & $95.7 \pm 0.9$ & $97.8 \pm 1.5$ & $97.7 \pm 0.9$ & $57.0$ & $100.0$ & $100.0$ & $100.0$ & $99.7 \pm 0.7$ & $98.2 \pm 0.8$ & $97.4 \pm 1.4$ & $98.7 \pm 0.8$ & $98.0$ & $98.2 \pm 0.8$ & $97.8 \pm 0.9$ & $97.9 \pm 2.0$ & $54.0$ \\
& is\_regular & $92.0$ & $92.0$ & $100.0$ & $90.4 \pm 1.2$ & $94.3 \pm 1.9$ & $85.7 \pm 3.2$ & $79.3 \pm 3.1$ & $98.0$ & $96.6 \pm 1.8$ & $93.7 \pm 1.6$ & $92.5 \pm 2.5$ & $91.0$ & $100.0$ & $100.0$ & $100.0$ & $99.1 \pm 1.0$ & $97.2 \pm 1.1$ & $92.7 \pm 1.1$ & $89.6 \pm 3.0$ & $100.0$ & $99.9 \pm 0.3$ & $99.6 \pm 0.7$ & $99.8 \pm 0.6$ & $93.0$ \\
& degree & $62.0$ & $71.0$ & $73.0$ & $62.2 \pm 4.3$ & $58.4 \pm 5.3$ & $56.5 \pm 3.6$ & $30.1 \pm 3.6$ & $83.0$ & $75.4 \pm 2.9$ & $85.6 \pm 2.7$ & $77.8 \pm 3.6$ & $38.0$ & $95.0$ & $91.0$ & $32.0$ & $91.7 \pm 1.3$ & $25.3 \pm 2.9$ & $75.9 \pm 2.3$ & $43.4 \pm 4.8$ & $95.0$ & $88.0 \pm 1.9$ & $87.9 \pm 2.1$ & $85.1 \pm 3.3$ & $21.0$ \\
& is\_tournament & $73.0$ & $72.0$ & $80.0$ & $73.6 \pm 3.6$ & $72.4 \pm 2.5$ & $63.5 \pm 3.2$ & $65.6 \pm 3.3$ & $80.0$ & $76.8 \pm 3.5$ & $70.3 \pm 6.1$ & $67.8 \pm 5.2$ & $48.0$ & $99.0$ & $99.0$ & $99.0$ & $98.9 \pm 0.3$ & $98.7 \pm 0.5$ & $98.7 \pm 0.7$ & $98.3 \pm 0.8$ & $99.0$ & $99.0 \pm 0.0$ & $98.0 \pm 0.9$ & $98.8 \pm 0.4$ & $91.0$ \\
& density & $32.0$ & $34.0$ & $1.0$ & $31.9 \pm 1.3$ & $1.3 \pm 0.5$ & $28.1 \pm 2.3$ & $2.5 \pm 1.3$ & $5.0$ & $2.5 \pm 1.8$ & $4.3 \pm 1.9$ & $2.6 \pm 1.3$ & $4.0$ & $92.0$ & $90.0$ & $12.0$ & $90.8 \pm 0.8$ & $11.9 \pm 0.6$ & $80.4 \pm 2.4$ & $10.6 \pm 1.8$ & $8.0$ & $6.6 \pm 1.9$ & $8.2 \pm 2.0$ & $7.4 \pm 2.9$ & $5.0$ \\
\midrule
\multirow{11}{*}{\rotatebox{90}{\textbf{Medium}}} & adamic\_adar\_index & $21.0$ & $18.0$ & $15.0$ & $13.6 \pm 2.3$ & $13.9 \pm 2.3$ & $6.0 \pm 2.7$ & $3.9 \pm 1.4$ & $12.0$ & $9.7 \pm 2.2$ & $12.7 \pm 3.1$ & $10.2 \pm 2.8$ & $3.0$ & $96.0$ & $95.0$ & $85.0$ & $93.9 \pm 2.0$ & $81.9 \pm 1.9$ & $81.7 \pm 1.6$ & $80.7 \pm 1.1$ & $97.0$ & $96.5 \pm 1.3$ & $97.8 \pm 1.3$ & $96.0 \pm 1.2$ & $40.0$ \\
& clustering\_coefficient & $35.0$ & $37.0$ & $35.0$ & $36.7 \pm 2.7$ & $35.8 \pm 3.2$ & $35.2 \pm 3.9$ & $31.0 \pm 4.0$ & $33.0$ & $33.4 \pm 2.5$ & $31.0 \pm 2.7$ & $29.4 \pm 3.4$ & $24.0$ & $80.0$ & $78.0$ & $77.0$ & $72.3 \pm 1.8$ & $74.2 \pm 3.2$ & $58.1 \pm 3.4$ & $63.9 \pm 1.7$ & $60.0$ & $61.1 \pm 3.0$ & $65.8 \pm 3.8$ & $63.8 \pm 1.7$ & $37.0$ \\
& connected\_component\_number & $27.0$ & $23.0$ & $26.0$ & $21.8 \pm 5.6$ & $18.5 \pm 2.3$ & $22.9 \pm 3.8$ & $18.8 \pm 2.5$ & $17.0$ & $16.7 \pm 4.9$ & $20.6 \pm 3.1$ & $17.6 \pm 4.6$ & $5.0$ & $74.0$ & $71.0$ & $73.0$ & $70.4 \pm 2.8$ & $64.1 \pm 2.6$ & $60.1 \pm 2.0$ & $57.0 \pm 2.0$ & $66.0$ & $64.0 \pm 2.1$ & $65.8 \pm 2.1$ & $64.3 \pm 2.8$ & $39.0$ \\
& bipartite\_maximum\_matching & $14.0$ & $9.0$ & $9.0$ & $9.4 \pm 2.5$ & $2.6 \pm 1.3$ & $1.0 \pm 1.1$ & $0.9 \pm 1.0$ & $11.0$ & $3.8 \pm 1.9$ & $3.0 \pm 1.3$ & $1.0 \pm 0.9$ & $3.0$ & $79.0$ & $73.0$ & $45.0$ & $69.9 \pm 2.4$ & $38.0 \pm 2.6$ & $15.2 \pm 3.6$ & $14.8 \pm 3.3$ & $40.0$ & $26.2 \pm 3.7$ & $12.2 \pm 1.5$ & $6.4 \pm 2.6$ & $4.0$ \\
& local\_connectivity & $58.0$ & $62.0$ & $56.0$ & $56.6 \pm 2.2$ & $52.9 \pm 1.7$ & $50.1 \pm 2.8$ & $52.0 \pm 2.7$ & $58.0$ & $60.9 \pm 2.4$ & $60.9 \pm 2.1$ & $57.4 \pm 1.8$ & $49.0$ & $82.0$ & $85.0$ & $80.0$ & $79.1 \pm 3.0$ & $71.9 \pm 3.2$ & $70.0 \pm 3.5$ & $66.0 \pm 3.4$ & $71.0$ & $73.7 \pm 3.5$ & $71.0 \pm 3.3$ & $67.7 \pm 2.6$ & $53.0$ \\
& jaccard\_coefficient & $55.0$ & $60.0$ & $52.0$ & $52.2 \pm 3.9$ & $49.2 \pm 6.4$ & $45.9 \pm 5.6$ & $33.2 \pm 4.6$ & $61.0$ & $60.5 \pm 3.4$ & $66.5 \pm 3.2$ & $56.5 \pm 5.9$ & $8.0$ & $97.0$ & $96.0$ & $96.0$ & $94.5 \pm 1.8$ & $96.0 \pm 1.2$ & $88.8 \pm 1.6$ & $89.3 \pm 1.8$ & $100.0$ & $96.3 \pm 1.6$ & $98.4 \pm 1.1$ & $98.5 \pm 0.8$ & $36.0$ \\
& min\_edge\_covering & $0.0$ & $2.0$ & $4.0$ & $2.0 \pm 0.7$ & $2.9 \pm 1.4$ & $1.3 \pm 0.9$ & $2.4 \pm 2.6$ & $7.0$ & $3.1 \pm 1.2$ & $5.1 \pm 1.7$ & $4.1 \pm 1.5$ & $4.0$ & $47.0$ & $50.0$ & $53.0$ & $47.4 \pm 4.0$ & $50.8 \pm 2.6$ & $39.2 \pm 3.5$ & $47.9 \pm 3.0$ & $51.0$ & $47.9 \pm 2.3$ & $49.2 \pm 3.4$ & $43.5 \pm 2.9$ & $49.0$ \\
& is\_eularian & $74.0$ & $84.0$ & $74.0$ & $77.4 \pm 2.3$ & $72.3 \pm 3.7$ & $73.8 \pm 4.0$ & $63.3 \pm 4.0$ & $83.0$ & $83.9 \pm 2.0$ & $79.8 \pm 2.1$ & $82.0 \pm 2.1$ & $53.0$ & $96.0$ & $97.0$ & $96.0$ & $95.6 \pm 1.3$ & $90.1 \pm 2.3$ & $88.5 \pm 2.0$ & $84.5 \pm 3.3$ & $91.0$ & $90.4 \pm 0.8$ & $89.5 \pm 1.5$ & $90.0 \pm 1.5$ & $78.0$ \\
& degree\_centrality & $9.0$ & $7.0$ & $17.0$ & $7.8 \pm 2.6$ & $12.8 \pm 2.0$ & $3.7 \pm 1.8$ & $11.3 \pm 2.7$ & $0.0$ & $0.0 \pm 0.0$ & $2.3 \pm 0.7$ & $3.6 \pm 1.9$ & $4.0$ & $91.0$ & $93.0$ & $53.0$ & $89.3 \pm 1.8$ & $47.7 \pm 4.0$ & $76.5 \pm 3.6$ & $59.2 \pm 3.6$ & $87.0$ & $86.4 \pm 1.6$ & $91.3 \pm 1.3$ & $87.6 \pm 2.5$ & $21.0$ \\
& is\_bipartite & $42.0$ & $49.0$ & $44.0$ & $46.8 \pm 2.6$ & $44.9 \pm 4.3$ & $44.8 \pm 3.8$ & $41.6 \pm 3.3$ & $46.0$ & $45.5 \pm 4.0$ & $38.7 \pm 2.5$ & $39.3 \pm 3.0$ & $37.0$ & $79.0$ & $71.0$ & $53.0$ & $71.1 \pm 2.1$ & $53.4 \pm 1.1$ & $67.7 \pm 3.8$ & $53.9 \pm 1.3$ & $61.0$ & $54.4 \pm 2.0$ & $60.4 \pm 2.3$ & $59.5 \pm 4.1$ & $59.0$ \\
& resource\_allocation\_index & $12.0$ & $10.0$ & $12.0$ & $11.5 \pm 2.5$ & $9.0 \pm 1.7$ & $7.2 \pm 2.8$ & $4.0 \pm 1.8$ & $24.0$ & $13.9 \pm 2.0$ & $23.8 \pm 3.0$ & $21.1 \pm 3.6$ & $4.0$ & $90.0$ & $88.0$ & $84.0$ & $89.4 \pm 1.0$ & $77.4 \pm 1.8$ & $80.9 \pm 1.4$ & $79.9 \pm 2.6$ & $91.0$ & $91.7 \pm 1.8$ & $92.6 \pm 1.1$ & $91.2 \pm 1.0$ & $38.0$ \\
\midrule
\multirow{17}{*}{\rotatebox{90}{\textbf{Hard}}} & max\_weight\_matching & $4.0$ & $2.0$ & $6.0$ & $3.4 \pm 1.9$ & $1.1 \pm 1.0$ & $2.6 \pm 1.3$ & $1.3 \pm 1.1$ & $8.0$ & $3.6 \pm 1.4$ & $3.5 \pm 1.5$ & $3.6 \pm 1.7$ & $3.0$ & $23.0$ & $25.0$ & $20.0$ & $26.8 \pm 4.0$ & $22.8 \pm 3.4$ & $23.7 \pm 2.8$ & $20.9 \pm 2.8$ & $23.0$ & $21.4 \pm 3.2$ & $18.9 \pm 2.8$ & $22.4 \pm 2.7$ & $5.0$ \\
& closeness\_centrality & $1.0$ & $0.0$ & $0.0$ & $0.7 \pm 0.7$ & $0.2 \pm 0.6$ & $0.3 \pm 0.7$ & $0.4 \pm 0.5$ & $2.0$ & $1.3 \pm 1.2$ & $1.5 \pm 1.0$ & $1.2 \pm 1.2$ & $1.0$ & $8.0$ & $8.0$ & $6.0$ & $6.7 \pm 0.9$ & $6.5 \pm 1.8$ & $6.0 \pm 1.4$ & $5.9 \pm 1.7$ & $7.0$ & $6.9 \pm 1.6$ & $6.7 \pm 1.4$ & $5.4 \pm 1.6$ & $3.0$ \\
& traveling\_salesman\_problem & $29.0$ & $29.0$ & $16.0$ & $29.1 \pm 2.5$ & $16.1 \pm 2.7$ & $28.0 \pm 2.7$ & $22.9 \pm 2.6$ & $30.0$ & $22.0 \pm 5.2$ & $26.6 \pm 2.8$ & $22.2 \pm 2.7$ & $32.0$ & $44.0$ & $44.0$ & $46.0$ & $44.7 \pm 3.7$ & $43.8 \pm 2.4$ & $37.6 \pm 4.6$ & $38.7 \pm 3.6$ & $42.0$ & $41.8 \pm 2.3$ & $43.3 \pm 1.9$ & $42.1 \pm 2.9$ & $41.0$ \\
& strongly\_connected\_number & $6.0$ & $7.0$ & $7.0$ & $7.7 \pm 2.2$ & $8.9 \pm 3.0$ & $6.4 \pm 2.6$ & $4.8 \pm 2.9$ & $4.0$ & $4.5 \pm 1.6$ & $6.4 \pm 2.8$ & $6.2 \pm 1.4$ & $8.0$ & $62.0$ & $57.0$ & $59.0$ & $57.0 \pm 1.2$ & $56.0 \pm 2.5$ & $55.1 \pm 1.3$ & $54.8 \pm 1.8$ & $52.0$ & $53.0 \pm 1.2$ & $53.2 \pm 1.0$ & $52.9 \pm 1.4$ & $37.0$ \\
& shortest\_path & $21.0$ & $18.0$ & $18.0$ & $18.3 \pm 3.0$ & $19.9 \pm 2.6$ & $14.2 \pm 3.2$ & $16.5 \pm 3.2$ & $18.0$ & $17.6 \pm 3.0$ & $17.8 \pm 3.0$ & $16.9 \pm 4.0$ & $7.0$ & $57.0$ & $56.0$ & $57.0$ & $54.9 \pm 2.9$ & $55.8 \pm 4.1$ & $39.9 \pm 2.5$ & $45.5 \pm 4.3$ & $54.0$ & $50.1 \pm 2.9$ & $56.1 \pm 4.6$ & $53.6 \pm 2.8$ & $8.0$ \\
& weighted\_shortest\_path & $2.0$ & $2.0$ & $4.0$ & $1.3 \pm 0.8$ & $2.5 \pm 1.1$ & $1.6 \pm 0.7$ & $2.0 \pm 1.2$ & $1.0$ & $2.3 \pm 0.8$ & $2.2 \pm 1.1$ & $2.1 \pm 1.1$ & $0.0$ & $8.0$ & $9.0$ & $9.0$ & $6.6 \pm 1.2$ & $9.0 \pm 1.6$ & $5.9 \pm 1.2$ & $6.7 \pm 1.4$ & $5.0$ & $7.4 \pm 1.4$ & $7.5 \pm 1.5$ & $7.6 \pm 1.8$ & $1.0$ \\
& center & $1.0$ & $8.0$ & $4.0$ & $4.7 \pm 0.9$ & $6.3 \pm 2.1$ & $4.2 \pm 1.4$ & $4.8 \pm 2.0$ & $5.0$ & $6.1 \pm 2.1$ & $5.2 \pm 2.3$ & $4.6 \pm 1.7$ & $5.0$ & $24.0$ & $24.0$ & $14.0$ & $25.1 \pm 1.5$ & $15.5 \pm 1.8$ & $16.2 \pm 2.1$ & $15.1 \pm 3.1$ & $23.0$ & $22.0 \pm 2.8$ & $19.5 \pm 2.8$ & $16.8 \pm 2.2$ & $9.0$ \\
& diameter & $12.0$ & $18.0$ & $12.0$ & $14.0 \pm 2.6$ & $13.2 \pm 3.6$ & $13.1 \pm 3.7$ & $13.1 \pm 3.7$ & $14.0$ & $13.0 \pm 2.5$ & $11.9 \pm 3.7$ & $12.5 \pm 3.8$ & $11.0$ & $40.0$ & $43.0$ & $32.0$ & $40.4 \pm 3.2$ & $31.1 \pm 4.0$ & $32.5 \pm 3.9$ & $32.0 \pm 4.3$ & $37.0$ & $37.5 \pm 3.1$ & $35.7 \pm 2.8$ & $33.0 \pm 3.3$ & $16.0$ \\
& barycenter & $11.0$ & $12.0$ & $18.0$ & $11.2 \pm 1.3$ & $15.9 \pm 4.8$ & $14.1 \pm 3.0$ & $13.0 \pm 2.5$ & $7.0$ & $11.0 \pm 2.6$ & $10.2 \pm 2.8$ & $10.1 \pm 1.8$ & $7.0$ & $37.0$ & $36.0$ & $26.0$ & $33.5 \pm 1.5$ & $27.6 \pm 3.4$ & $30.8 \pm 2.5$ & $29.4 \pm 3.4$ & $38.0$ & $33.6 \pm 4.2$ & $31.5 \pm 5.8$ & $32.5 \pm 4.1$ & $6.0$ \\
& radius & $18.0$ & $17.0$ & $22.0$ & $19.4 \pm 5.1$ & $19.8 \pm 3.8$ & $22.2 \pm 3.3$ & $19.4 \pm 4.0$ & $17.0$ & $23.0 \pm 3.7$ & $21.2 \pm 2.5$ & $22.7 \pm 3.3$ & $24.0$ & $62.0$ & $52.0$ & $35.0$ & $53.1 \pm 2.9$ & $40.0 \pm 3.0$ & $49.0 \pm 4.1$ & $43.8 \pm 2.6$ & $47.0$ & $47.2 \pm 4.1$ & $47.0 \pm 2.4$ & $46.9 \pm 3.1$ & $39.0$ \\
& topological\_sort & $17.0$ & $13.0$ & $11.0$ & $12.0 \pm 2.4$ & $12.6 \pm 2.5$ & $10.1 \pm 2.1$ & $9.5 \pm 1.8$ & $4.0$ & $3.0 \pm 1.6$ & $2.1 \pm 0.9$ & $1.5 \pm 1.2$ & $6.0$ & $70.0$ & $62.0$ & $66.0$ & $63.9 \pm 3.7$ & $64.6 \pm 2.8$ & $62.4 \pm 2.8$ & $63.5 \pm 2.5$ & $36.0$ & $40.6 \pm 3.1$ & $32.0 \pm 5.4$ & $30.2 \pm 4.2$ & $9.0$ \\
& periphery & $2.0$ & $4.0$ & $2.0$ & $1.8 \pm 0.9$ & $3.3 \pm 1.6$ & $3.0 \pm 1.2$ & $3.1 \pm 1.4$ & $3.0$ & $2.9 \pm 1.7$ & $2.6 \pm 1.5$ & $3.8 \pm 1.8$ & $4.0$ & $22.0$ & $18.0$ & $10.0$ & $18.4 \pm 1.9$ & $13.9 \pm 2.1$ & $13.3 \pm 2.8$ & $12.7 \pm 1.7$ & $16.0$ & $15.9 \pm 1.7$ & $11.9 \pm 1.8$ & $10.9 \pm 2.3$ & $3.0$ \\
& betweenness\_centrality & $1.0$ & $2.0$ & $2.0$ & $1.8 \pm 1.1$ & $1.3 \pm 0.8$ & $1.2 \pm 1.0$ & $0.9 \pm 0.9$ & $2.0$ & $2.4 \pm 1.3$ & $2.1 \pm 1.7$ & $1.7 \pm 1.1$ & $3.0$ & $38.0$ & $39.0$ & $39.0$ & $38.8 \pm 0.6$ & $38.0 \pm 0.7$ & $38.7 \pm 0.5$ & $37.7 \pm 1.1$ & $39.0$ & $38.3 \pm 0.7$ & $38.2 \pm 0.6$ & $38.4 \pm 0.5$ & $37.0$ \\
& triangles & $16.0$ & $16.0$ & $17.0$ & $11.0 \pm 2.2$ & $10.7 \pm 2.9$ & $9.2 \pm 2.4$ & $8.9 \pm 2.9$ & $17.0$ & $11.1 \pm 3.3$ & $13.0 \pm 2.5$ & $10.5 \pm 2.5$ & $1.0$ & $67.0$ & $68.0$ & $64.0$ & $50.8 \pm 3.6$ & $51.1 \pm 3.7$ & $25.7 \pm 3.6$ & $35.0 \pm 2.4$ & $39.0$ & $38.1 \pm 3.8$ & $41.9 \pm 4.0$ & $38.7 \pm 3.9$ & $13.0$ \\
& avg\_neighbor\_degree & $14.0$ & $13.0$ & $17.0$ & $12.6 \pm 2.7$ & $12.2 \pm 3.9$ & $14.8 \pm 3.0$ & $10.7 \pm 1.6$ & $44.0$ & $34.7 \pm 4.9$ & $48.6 \pm 4.0$ & $40.5 \pm 2.6$ & $5.0$ & $68.0$ & $68.0$ & $42.0$ & $65.9 \pm 2.6$ & $22.6 \pm 2.5$ & $38.4 \pm 3.3$ & $27.9 \pm 3.4$ & $83.0$ & $74.7 \pm 3.1$ & $80.8 \pm 2.5$ & $78.5 \pm 3.4$ & $10.0$ \\
& harmonic\_centrality & $3.0$ & $4.0$ & $3.0$ & $3.0 \pm 0.9$ & $2.3 \pm 1.6$ & $1.7 \pm 1.3$ & $1.8 \pm 0.6$ & $7.0$ & $3.7 \pm 1.3$ & $3.2 \pm 1.8$ & $3.2 \pm 1.1$ & $2.0$ & $18.0$ & $21.0$ & $15.0$ & $17.3 \pm 1.3$ & $14.1 \pm 1.4$ & $14.0 \pm 1.9$ & $13.2 \pm 2.5$ & $17.0$ & $14.0 \pm 2.6$ & $13.6 \pm 3.0$ & $11.5 \pm 3.0$ & $2.0$ \\
& bridges & $0.0$ & $2.0$ & $2.0$ & $0.9 \pm 0.7$ & $0.4 \pm 0.5$ & $0.1 \pm 0.3$ & $0.2 \pm 0.4$ & $1.0$ & $0.5 \pm 0.7$ & $0.3 \pm 0.5$ & $0.2 \pm 0.4$ & $0.0$ & $11.0$ & $15.0$ & $16.0$ & $13.0 \pm 1.5$ & $14.6 \pm 0.7$ & $5.9 \pm 1.7$ & $4.4 \pm 1.3$ & $10.0$ & $11.1 \pm 2.0$ & $8.8 \pm 1.9$ & $9.5 \pm 1.4$ & $2.0$ \\
\midrule
\multirow{6}{*}{\rotatebox{90}{\textbf{Challenging}}} & global\_efficiency & $0.0$ & $1.0$ & $1.0$ & $0.7 \pm 0.5$ & $1.2 \pm 0.6$ & $0.7 \pm 0.7$ & $0.4 \pm 0.5$ & $1.0$ & $1.0 \pm 0.0$ & $0.9 \pm 0.3$ & $1.2 \pm 0.6$ & $0.0$ & $1.0$ & $1.0$ & $1.0$ & $1.1 \pm 0.3$ & $1.0 \pm 0.0$ & $1.0 \pm 0.5$ & $1.0 \pm 0.0$ & $1.0$ & $1.0 \pm 0.0$ & $1.1 \pm 0.3$ & $1.0 \pm 0.0$ & $0.0$ \\
& maximal\_independent\_set & $1.0$ & $3.0$ & $2.0$ & $2.6 \pm 1.1$ & $1.7 \pm 1.4$ & $1.5 \pm 1.0$ & $1.1 \pm 1.0$ & $3.0$ & $2.8 \pm 1.7$ & $2.1 \pm 1.1$ & $2.0 \pm 1.2$ & $1.0$ & $12.0$ & $11.0$ & $19.0$ & $12.9 \pm 2.5$ & $12.8 \pm 2.0$ & $7.8 \pm 2.7$ & $10.0 \pm 1.6$ & $12.0$ & $11.5 \pm 1.8$ & $11.8 \pm 2.1$ & $10.9 \pm 2.5$ & $4.0$ \\
& maximum\_flow & $1.0$ & $1.0$ & $2.0$ & $2.2 \pm 1.4$ & $2.0 \pm 1.3$ & $1.8 \pm 1.1$ & $2.0 \pm 1.4$ & $2.0$ & $2.4 \pm 0.8$ & $1.4 \pm 1.6$ & $1.5 \pm 1.1$ & $2.0$ & $7.0$ & $7.0$ & $7.0$ & $11.1 \pm 1.9$ & $11.4 \pm 3.0$ & $8.0 \pm 1.1$ & $8.3 \pm 0.8$ & $7.0$ & $8.5 \pm 1.3$ & $8.6 \pm 1.7$ & $8.2 \pm 1.3$ & $3.0$ \\
& wiener\_index & $1.0$ & $1.0$ & $4.0$ & $0.9 \pm 1.0$ & $1.4 \pm 1.2$ & $1.1 \pm 0.9$ & $1.3 \pm 1.3$ & $3.0$ & $1.3 \pm 0.7$ & $2.4 \pm 1.2$ & $2.2 \pm 1.9$ & $1.0$ & $6.0$ & $6.0$ & $1.0$ & $5.0 \pm 0.8$ & $3.0 \pm 0.8$ & $6.2 \pm 1.8$ & $4.1 \pm 1.0$ & $3.0$ & $3.6 \pm 0.8$ & $3.9 \pm 1.4$ & $2.9 \pm 1.0$ & $4.0$ \\
& hamiltonian\_path & $1.0$ & $1.0$ & $1.0$ & $0.0 \pm 0.0$ & $0.0 \pm 0.0$ & $0.1 \pm 0.3$ & $0.0 \pm 0.0$ & $2.0$ & $0.4 \pm 0.5$ & $0.9 \pm 0.6$ & $0.2 \pm 0.4$ & $1.0$ & $2.0$ & $3.0$ & $3.0$ & $1.6 \pm 1.2$ & $1.3 \pm 1.2$ & $1.1 \pm 0.9$ & $1.1 \pm 1.1$ & $2.0$ & $0.5 \pm 0.5$ & $1.0 \pm 1.2$ & $0.4 \pm 0.5$ & $2.0$ \\
& min\_vertex\_cover & $3.0$ & $5.0$ & $3.0$ & $4.6 \pm 1.3$ & $3.0 \pm 1.5$ & $2.5 \pm 1.6$ & $2.9 \pm 1.3$ & $2.0$ & $1.8 \pm 1.0$ & $2.0 \pm 0.9$ & $2.3 \pm 0.9$ & $1.0$ & $16.0$ & $13.0$ & $15.0$ & $14.4 \pm 1.8$ & $8.8 \pm 2.9$ & $11.0 \pm 1.7$ & $8.3 \pm 2.6$ & $10.0$ & $7.3 \pm 1.8$ & $7.2 \pm 1.6$ & $6.6 \pm 1.5$ & $4.0$ \\
\bottomrule
\end{tabular}
}

    }
    \label{tab:edge_reorderings_all_3B}
\end{sidewaystable}

\begin{sidewaystable}
    \small
    \centering
    \captionsetup{margin=1cm}
    \caption{Edge reordering ablation results for the 7B models. We report the accuracy in \%. \Erdos baseline (w/ zero temperature) is shaded. For the computational structure, \texttt{edge} = edge list, \texttt{adjl} = adjacency list, \texttt{adjm} = adjacency matrix. For sorting, \texttt{s} = source and \texttt{t} = target, and the entry represents the sorting key. Shuffling is applied to all remaining ambiguities; in this case we report the mean accuracy $\pm$ std over 10 seeds.}
    \resizebox{\linewidth}{!}{
        {\renewcommand{\arraystretch}{1.2}
\begin{tabular}{ll >{\columncolor{gray!20}}rrrrrrrrrrrr : >{\columncolor{gray!20}}rrrrrrrrrrrr}

& & \multicolumn{12}{c:}{\large\textbf{Qwen-7B}} & \multicolumn{12}{:c}{\large\textbf{G1-7B}} \\[2ex]
\toprule
& \textbf{Comp. structure} & \texttt{edge} & \texttt{edge} & \texttt{edge} & \texttt{edge} & \texttt{edge} & \texttt{edge} & \texttt{edge} & \texttt{adjl} & \texttt{adjl} & \texttt{adjl} & \texttt{adjl} & \texttt{adjm} & \texttt{edge} & \texttt{edge} & \texttt{edge} & \texttt{edge} & \texttt{edge} & \texttt{edge} & \texttt{edge} & \texttt{adjl} & \texttt{adjl} & \texttt{adjl} & \texttt{adjl} & \texttt{adjm} \\
& \textbf{Sort (rest shuffled)} & \Erdos & \texttt{(s,t)} & \texttt{(s,t)} & \texttt{s} & \texttt{s} & \xmark & \xmark & \texttt{(s,t)} & \texttt{s} & \texttt{t} & \xmark & \texttt{NA} & \Erdos & \texttt{(s,t)} & \texttt{(s,t)} & \texttt{s} & \texttt{s} & \xmark & \xmark & \texttt{(s,t)} & \texttt{s} & \texttt{t} & \xmark & \texttt{NA} \\
& \textbf{Repl. undir} & \xmark & \xmark & \cmark & \xmark & \cmark & \xmark & \cmark & (\cmark) & (\cmark) & (\cmark) & (\cmark) & (\cmark) & \xmark & \xmark & \cmark & \xmark & \cmark & \xmark & \cmark & (\cmark) & (\cmark) & (\cmark) & (\cmark) & (\cmark) \\
\midrule
\multirow{15}{*}{\rotatebox{90}{\textbf{Easy}}} & node\_number & $95.0$ & $98.0$ & $99.0$ & $97.6 \pm 1.6$ & $96.0 \pm 1.8$ & $99.2 \pm 1.1$ & $97.7 \pm 1.2$ & $99.0$ & $99.4 \pm 0.7$ & $99.0 \pm 1.2$ & $97.9 \pm 1.9$ & $95.0$ & $100.0$ & $100.0$ & $100.0$ & $100.0 \pm 0.0$ & $100.0 \pm 0.0$ & $99.9 \pm 0.3$ & $98.0 \pm 0.8$ & $100.0$ & $99.6 \pm 0.5$ & $99.5 \pm 0.7$ & $99.4 \pm 0.7$ & $100.0$ \\
& dominating\_set & $27.0$ & $21.0$ & $23.0$ & $26.4 \pm 2.7$ & $27.6 \pm 4.4$ & $33.4 \pm 4.6$ & $35.2 \pm 3.6$ & $25.0$ & $34.0 \pm 5.1$ & $37.5 \pm 4.1$ & $37.1 \pm 6.4$ & $21.0$ & $98.0$ & $97.0$ & $98.0$ & $98.4 \pm 1.1$ & $97.2 \pm 1.3$ & $94.2 \pm 2.0$ & $89.5 \pm 2.1$ & $99.0$ & $98.1 \pm 1.1$ & $96.9 \pm 1.7$ & $97.0 \pm 1.5$ & $89.0$ \\
& common\_neighbor & $51.0$ & $58.0$ & $76.0$ & $53.3 \pm 3.0$ & $68.1 \pm 3.0$ & $42.7 \pm 2.7$ & $48.9 \pm 3.8$ & $86.0$ & $76.1 \pm 3.8$ & $89.2 \pm 2.5$ & $78.2 \pm 2.3$ & $25.0$ & $94.0$ & $94.0$ & $95.0$ & $92.0 \pm 1.8$ & $91.2 \pm 2.0$ & $75.5 \pm 1.8$ & $84.4 \pm 3.6$ & $97.0$ & $93.3 \pm 2.4$ & $97.7 \pm 0.9$ & $91.6 \pm 2.4$ & $38.0$ \\
& edge\_number & $60.0$ & $78.0$ & $23.0$ & $74.4 \pm 3.4$ & $18.1 \pm 2.5$ & $43.1 \pm 4.6$ & $15.1 \pm 4.2$ & $21.0$ & $15.1 \pm 1.7$ & $15.1 \pm 3.2$ & $14.0 \pm 1.9$ & $8.0$ & $97.0$ & $97.0$ & $19.0$ & $96.4 \pm 1.1$ & $16.7 \pm 1.6$ & $87.5 \pm 2.0$ & $13.2 \pm 1.9$ & $25.0$ & $17.4 \pm 2.0$ & $15.0 \pm 2.1$ & $14.3 \pm 3.0$ & $10.0$ \\
& neighbor & $72.0$ & $72.0$ & $73.0$ & $72.2 \pm 2.0$ & $75.0 \pm 2.1$ & $52.1 \pm 4.1$ & $59.0 \pm 3.2$ & $88.0$ & $86.3 \pm 3.2$ & $90.3 \pm 2.3$ & $86.3 \pm 3.1$ & $32.0$ & $96.0$ & $100.0$ & $95.0$ & $98.5 \pm 1.0$ & $95.8 \pm 1.5$ & $84.7 \pm 2.5$ & $88.9 \pm 2.1$ & $99.0$ & $96.5 \pm 1.2$ & $99.4 \pm 0.7$ & $96.2 \pm 1.2$ & $39.0$ \\
& bfs & $6.0$ & $9.0$ & $9.0$ & $11.4 \pm 2.7$ & $12.3 \pm 2.9$ & $6.8 \pm 1.2$ & $6.1 \pm 1.7$ & $7.0$ & $7.5 \pm 1.7$ & $11.4 \pm 2.9$ & $9.4 \pm 2.5$ & $4.0$ & $97.0$ & $99.0$ & $95.0$ & $98.2 \pm 1.0$ & $96.4 \pm 1.8$ & $97.3 \pm 1.4$ & $91.1 \pm 2.4$ & $100.0$ & $98.8 \pm 1.2$ & $99.4 \pm 0.7$ & $99.7 \pm 0.5$ & $34.0$ \\
& has\_cycle & $54.0$ & $53.0$ & $50.0$ & $58.2 \pm 3.5$ & $50.3 \pm 2.5$ & $53.9 \pm 4.6$ & $52.5 \pm 1.6$ & $53.0$ & $55.9 \pm 3.7$ & $55.0 \pm 4.5$ & $55.1 \pm 3.9$ & $48.0$ & $87.0$ & $86.0$ & $61.0$ & $85.1 \pm 2.2$ & $61.5 \pm 1.6$ & $73.5 \pm 3.3$ & $57.4 \pm 1.9$ & $56.0$ & $58.1 \pm 3.0$ & $60.4 \pm 1.6$ & $59.7 \pm 2.7$ & $45.0$ \\
& dfs & $30.0$ & $25.0$ & $25.0$ & $28.4 \pm 2.4$ & $24.6 \pm 3.8$ & $20.8 \pm 5.2$ & $19.5 \pm 3.9$ & $29.0$ & $27.8 \pm 3.9$ & $24.4 \pm 4.4$ & $22.4 \pm 4.2$ & $4.0$ & $100.0$ & $100.0$ & $99.0$ & $98.7 \pm 1.6$ & $98.8 \pm 1.0$ & $95.4 \pm 2.2$ & $93.3 \pm 1.9$ & $100.0$ & $99.3 \pm 1.1$ & $99.0 \pm 1.1$ & $98.8 \pm 0.9$ & $43.0$ \\
& minimum\_spanning\_tree & $12.0$ & $12.0$ & $11.0$ & $11.8 \pm 3.2$ & $10.0 \pm 1.8$ & $6.5 \pm 1.7$ & $7.6 \pm 1.6$ & $18.0$ & $17.3 \pm 2.9$ & $13.8 \pm 2.5$ & $13.9 \pm 1.8$ & $13.0$ & $68.0$ & $59.0$ & $62.0$ & $58.0 \pm 1.9$ & $54.3 \pm 2.6$ & $35.1 \pm 3.8$ & $36.6 \pm 2.6$ & $60.0$ & $61.6 \pm 2.6$ & $61.8 \pm 4.0$ & $53.3 \pm 2.0$ & $59.0$ \\
& weighted\_minimum\_spanning\_tree & $1.0$ & $2.0$ & $5.0$ & $1.5 \pm 0.7$ & $2.6 \pm 1.1$ & $1.4 \pm 1.3$ & $3.1 \pm 1.2$ & $3.0$ & $2.1 \pm 1.0$ & $1.6 \pm 1.2$ & $1.7 \pm 1.2$ & $0.0$ & $16.0$ & $15.0$ & $7.0$ & $11.8 \pm 2.4$ & $9.4 \pm 2.5$ & $12.5 \pm 2.3$ & $8.6 \pm 1.7$ & $10.0$ & $8.8 \pm 2.3$ & $8.6 \pm 2.4$ & $7.3 \pm 1.8$ & $4.0$ \\
& edge\_existence & $97.0$ & $96.0$ & $89.0$ & $96.4 \pm 1.1$ & $92.5 \pm 2.4$ & $91.9 \pm 2.5$ & $91.4 \pm 2.0$ & $98.0$ & $98.8 \pm 0.6$ & $98.8 \pm 1.2$ & $99.1 \pm 1.1$ & $83.0$ & $100.0$ & $100.0$ & $100.0$ & $100.0 \pm 0.0$ & $100.0 \pm 0.0$ & $99.2 \pm 0.8$ & $99.0 \pm 1.1$ & $100.0$ & $99.8 \pm 0.4$ & $100.0 \pm 0.0$ & $99.8 \pm 0.4$ & $80.0$ \\
& is\_regular & $96.0$ & $97.0$ & $99.0$ & $94.8 \pm 1.3$ & $98.0 \pm 0.8$ & $92.6 \pm 1.9$ & $85.0 \pm 3.0$ & $100.0$ & $99.9 \pm 0.3$ & $99.8 \pm 0.4$ & $99.9 \pm 0.3$ & $100.0$ & $98.0$ & $100.0$ & $99.0$ & $99.5 \pm 0.7$ & $94.5 \pm 0.7$ & $93.4 \pm 1.4$ & $91.0 \pm 1.2$ & $100.0$ & $100.0 \pm 0.0$ & $99.9 \pm 0.3$ & $99.8 \pm 0.4$ & $96.0$ \\
& degree & $78.0$ & $76.0$ & $66.0$ & $76.0 \pm 2.7$ & $65.9 \pm 3.4$ & $62.8 \pm 3.6$ & $49.8 \pm 3.9$ & $83.0$ & $79.6 \pm 2.8$ & $90.5 \pm 2.0$ & $85.3 \pm 3.7$ & $43.0$ & $97.0$ & $97.0$ & $49.0$ & $96.7 \pm 1.1$ & $44.8 \pm 2.5$ & $87.7 \pm 2.0$ & $41.1 \pm 2.5$ & $97.0$ & $93.3 \pm 1.3$ & $96.1 \pm 1.7$ & $96.1 \pm 1.4$ & $58.0$ \\
& is\_tournament & $82.0$ & $78.0$ & $81.0$ & $79.0 \pm 1.9$ & $79.1 \pm 2.6$ & $86.6 \pm 2.6$ & $87.4 \pm 2.2$ & $93.0$ & $90.1 \pm 2.3$ & $88.6 \pm 2.9$ & $87.1 \pm 2.6$ & $40.0$ & $97.0$ & $99.0$ & $99.0$ & $98.8 \pm 0.4$ & $98.9 \pm 0.3$ & $98.9 \pm 1.2$ & $98.3 \pm 1.3$ & $99.0$ & $98.9 \pm 0.6$ & $98.0 \pm 1.1$ & $97.2 \pm 1.9$ & $99.0$ \\
& density & $32.0$ & $33.0$ & $6.0$ & $32.6 \pm 2.5$ & $8.8 \pm 1.8$ & $36.0 \pm 2.9$ & $9.3 \pm 2.4$ & $7.0$ & $7.3 \pm 0.9$ & $8.5 \pm 2.4$ & $7.1 \pm 2.8$ & $6.0$ & $98.0$ & $99.0$ & $15.0$ & $98.2 \pm 0.8$ & $13.3 \pm 1.3$ & $90.9 \pm 3.1$ & $12.3 \pm 1.3$ & $22.0$ & $19.4 \pm 2.4$ & $10.4 \pm 2.0$ & $12.8 \pm 1.5$ & $18.0$ \\
\midrule
\multirow{11}{*}{\rotatebox{90}{\textbf{Medium}}} & adamic\_adar\_index & $27.0$ & $34.0$ & $42.0$ & $31.8 \pm 4.7$ & $31.8 \pm 3.6$ & $30.5 \pm 3.6$ & $17.1 \pm 2.3$ & $51.0$ & $47.6 \pm 2.2$ & $51.9 \pm 1.7$ & $45.6 \pm 3.6$ & $13.0$ & $98.0$ & $96.0$ & $79.0$ & $96.4 \pm 1.2$ & $75.8 \pm 1.4$ & $89.4 \pm 2.4$ & $70.7 \pm 3.2$ & $98.0$ & $97.8 \pm 0.9$ & $98.3 \pm 0.7$ & $97.6 \pm 0.8$ & $60.0$ \\
& clustering\_coefficient & $50.0$ & $44.0$ & $45.0$ & $47.4 \pm 4.1$ & $42.0 \pm 2.4$ & $46.9 \pm 2.6$ & $41.0 \pm 2.4$ & $55.0$ & $54.0 \pm 4.3$ & $54.1 \pm 3.1$ & $56.1 \pm 4.5$ & $33.0$ & $88.0$ & $86.0$ & $84.0$ & $84.7 \pm 2.0$ & $82.1 \pm 2.3$ & $69.4 \pm 3.2$ & $75.4 \pm 4.1$ & $84.0$ & $82.3 \pm 2.4$ & $79.8 \pm 2.2$ & $80.7 \pm 1.6$ & $44.0$ \\
& connected\_component\_number & $42.0$ & $41.0$ & $31.0$ & $40.5 \pm 3.6$ & $34.3 \pm 2.4$ & $31.9 \pm 3.5$ & $32.9 \pm 4.7$ & $42.0$ & $43.7 \pm 5.3$ & $47.0 \pm 3.2$ & $47.0 \pm 3.3$ & $22.0$ & $92.0$ & $87.0$ & $78.0$ & $84.8 \pm 2.6$ & $71.6 \pm 3.8$ & $61.0 \pm 3.1$ & $46.6 \pm 3.7$ & $84.0$ & $77.2 \pm 3.3$ & $68.1 \pm 3.4$ & $66.3 \pm 3.7$ & $43.0$ \\
& bipartite\_maximum\_matching & $15.0$ & $14.0$ & $12.0$ & $14.5 \pm 2.0$ & $9.2 \pm 3.1$ & $0.8 \pm 0.8$ & $0.6 \pm 0.7$ & $16.0$ & $8.2 \pm 1.9$ & $2.3 \pm 0.9$ & $1.6 \pm 0.7$ & $3.0$ & $83.0$ & $85.0$ & $75.0$ & $78.4 \pm 2.0$ & $44.5 \pm 3.9$ & $7.9 \pm 3.2$ & $9.5 \pm 2.8$ & $70.0$ & $45.8 \pm 5.7$ & $12.2 \pm 2.7$ & $7.9 \pm 1.8$ & $10.0$ \\
& local\_connectivity & $70.0$ & $70.0$ & $66.0$ & $68.2 \pm 4.3$ & $68.7 \pm 2.2$ & $61.7 \pm 2.1$ & $59.2 \pm 2.1$ & $77.0$ & $76.3 \pm 3.9$ & $73.8 \pm 1.5$ & $72.8 \pm 3.0$ & $64.0$ & $90.0$ & $93.0$ & $88.0$ & $93.9 \pm 2.4$ & $85.5 \pm 1.4$ & $83.6 \pm 2.8$ & $71.4 \pm 2.8$ & $96.0$ & $94.5 \pm 2.2$ & $92.7 \pm 2.3$ & $92.1 \pm 1.7$ & $65.0$ \\
& jaccard\_coefficient & $82.0$ & $82.0$ & $85.0$ & $76.0 \pm 2.8$ & $78.2 \pm 3.6$ & $62.7 \pm 3.0$ & $51.5 \pm 4.0$ & $96.0$ & $91.5 \pm 2.8$ & $97.2 \pm 1.5$ & $92.7 \pm 2.8$ & $59.0$ & $98.0$ & $100.0$ & $97.0$ & $99.3 \pm 0.8$ & $97.1 \pm 0.9$ & $93.3 \pm 1.4$ & $92.9 \pm 2.7$ & $100.0$ & $99.0 \pm 0.8$ & $98.9 \pm 1.0$ & $99.2 \pm 0.8$ & $69.0$ \\
& min\_edge\_covering & $3.0$ & $2.0$ & $8.0$ & $4.0 \pm 1.2$ & $9.0 \pm 1.1$ & $4.8 \pm 1.5$ & $7.2 \pm 1.8$ & $7.0$ & $7.6 \pm 2.6$ & $7.1 \pm 3.3$ & $8.0 \pm 2.1$ & $6.0$ & $44.0$ & $44.0$ & $54.0$ & $46.4 \pm 3.0$ & $47.9 \pm 3.4$ & $40.1 \pm 3.5$ & $44.8 \pm 4.4$ & $39.0$ & $43.0 \pm 3.7$ & $47.0 \pm 4.1$ & $44.4 \pm 3.9$ & $43.0$ \\
& is\_eularian & $82.0$ & $84.0$ & $79.0$ & $81.8 \pm 1.4$ & $73.4 \pm 3.0$ & $78.4 \pm 3.3$ & $62.5 \pm 5.0$ & $87.0$ & $84.0 \pm 1.7$ & $85.9 \pm 1.1$ & $84.8 \pm 1.0$ & $81.0$ & $93.0$ & $94.0$ & $90.0$ & $94.2 \pm 1.7$ & $85.7 \pm 2.7$ & $86.6 \pm 2.0$ & $75.8 \pm 2.8$ & $86.0$ & $83.5 \pm 2.2$ & $85.8 \pm 2.7$ & $84.4 \pm 2.8$ & $90.0$ \\
& degree\_centrality & $10.0$ & $8.0$ & $10.0$ & $8.5 \pm 2.2$ & $9.1 \pm 2.6$ & $9.6 \pm 2.8$ & $4.7 \pm 1.9$ & $32.0$ & $34.8 \pm 3.8$ & $35.4 \pm 3.2$ & $37.3 \pm 3.4$ & $21.0$ & $98.0$ & $97.0$ & $33.0$ & $97.4 \pm 0.7$ & $28.2 \pm 4.7$ & $88.0 \pm 2.5$ & $42.5 \pm 5.0$ & $96.0$ & $91.5 \pm 2.2$ & $95.9 \pm 1.4$ & $93.6 \pm 2.5$ & $48.0$ \\
& is\_bipartite & $48.0$ & $52.0$ & $55.0$ & $52.1 \pm 3.3$ & $47.8 \pm 3.0$ & $50.6 \pm 3.1$ & $47.4 \pm 2.3$ & $55.0$ & $50.2 \pm 3.0$ & $51.0 \pm 4.0$ & $50.2 \pm 3.5$ & $54.0$ & $79.0$ & $74.0$ & $56.0$ & $74.3 \pm 1.6$ & $56.8 \pm 1.6$ & $73.1 \pm 1.4$ & $53.9 \pm 2.4$ & $80.0$ & $80.4 \pm 1.3$ & $77.2 \pm 2.3$ & $75.3 \pm 2.3$ & $68.0$ \\
& resource\_allocation\_index & $38.0$ & $41.0$ & $49.0$ & $37.4 \pm 1.6$ & $40.1 \pm 2.8$ & $45.1 \pm 4.4$ & $30.8 \pm 3.6$ & $79.0$ & $68.9 \pm 3.4$ & $72.9 \pm 3.3$ & $61.8 \pm 4.1$ & $18.0$ & $92.0$ & $94.0$ & $81.0$ & $93.2 \pm 0.8$ & $74.4 \pm 0.8$ & $86.2 \pm 1.1$ & $73.0 \pm 1.3$ & $94.0$ & $94.0 \pm 1.2$ & $94.1 \pm 1.0$ & $93.5 \pm 1.2$ & $59.0$ \\
\midrule
\multirow{17}{*}{\rotatebox{90}{\textbf{Hard}}} & max\_weight\_matching & $10.0$ & $14.0$ & $16.0$ & $16.0 \pm 3.3$ & $12.0 \pm 2.7$ & $12.3 \pm 2.5$ & $7.5 \pm 2.3$ & $11.0$ & $15.1 \pm 1.5$ & $11.9 \pm 2.0$ & $13.1 \pm 2.6$ & $6.0$ & $42.0$ & $42.0$ & $39.0$ & $42.0 \pm 2.1$ & $39.5 \pm 2.8$ & $25.4 \pm 2.8$ & $24.3 \pm 3.3$ & $28.0$ & $35.4 \pm 2.2$ & $31.0 \pm 1.4$ & $30.7 \pm 3.9$ & $14.0$ \\
& closeness\_centrality & $4.0$ & $1.0$ & $2.0$ & $3.0 \pm 0.9$ & $3.8 \pm 1.5$ & $1.5 \pm 1.0$ & $2.3 \pm 1.5$ & $3.0$ & $3.4 \pm 1.6$ & $3.7 \pm 2.2$ & $3.6 \pm 1.8$ & $2.0$ & $11.0$ & $11.0$ & $9.0$ & $12.4 \pm 2.3$ & $9.7 \pm 1.8$ & $8.3 \pm 1.5$ & $6.8 \pm 1.3$ & $10.0$ & $9.4 \pm 1.4$ & $8.6 \pm 1.7$ & $9.2 \pm 1.4$ & $1.0$ \\
& traveling\_salesman\_problem & $44.0$ & $44.0$ & $42.0$ & $45.5 \pm 2.0$ & $42.7 \pm 1.1$ & $43.2 \pm 4.5$ & $38.6 \pm 3.3$ & $45.0$ & $44.3 \pm 2.5$ & $44.3 \pm 2.2$ & $43.1 \pm 3.8$ & $39.0$ & $53.0$ & $53.0$ & $52.0$ & $54.1 \pm 2.8$ & $50.7 \pm 1.6$ & $50.9 \pm 2.3$ & $47.3 \pm 2.2$ & $49.0$ & $51.6 \pm 2.2$ & $51.8 \pm 3.0$ & $49.2 \pm 2.8$ & $46.0$ \\
& strongly\_connected\_number & $11.0$ & $14.0$ & $9.0$ & $12.9 \pm 2.8$ & $12.5 \pm 2.3$ & $9.5 \pm 2.1$ & $9.2 \pm 2.3$ & $9.0$ & $8.8 \pm 2.0$ & $9.9 \pm 3.3$ & $9.8 \pm 3.8$ & $8.0$ & $59.0$ & $62.0$ & $60.0$ & $58.0 \pm 2.3$ & $58.8 \pm 1.5$ & $56.0 \pm 1.8$ & $57.3 \pm 2.6$ & $60.0$ & $59.6 \pm 1.6$ & $57.5 \pm 2.3$ & $57.4 \pm 1.1$ & $52.0$ \\
& shortest\_path & $41.0$ & $30.0$ & $38.0$ & $28.0 \pm 3.2$ & $35.1 \pm 3.5$ & $31.3 \pm 4.3$ & $29.9 \pm 3.3$ & $35.0$ & $40.3 \pm 5.1$ & $41.5 \pm 3.2$ & $38.8 \pm 4.4$ & $13.0$ & $79.0$ & $74.0$ & $68.0$ & $70.1 \pm 1.3$ & $71.8 \pm 3.3$ & $49.0 \pm 2.9$ & $52.6 \pm 3.2$ & $72.0$ & $67.5 \pm 3.7$ & $73.6 \pm 2.2$ & $69.8 \pm 3.3$ & $22.0$ \\
& weighted\_shortest\_path & $4.0$ & $1.0$ & $4.0$ & $2.6 \pm 1.2$ & $3.6 \pm 1.1$ & $4.0 \pm 1.1$ & $2.6 \pm 1.3$ & $5.0$ & $5.3 \pm 1.5$ & $5.3 \pm 1.3$ & $3.7 \pm 1.2$ & $1.0$ & $15.0$ & $17.0$ & $16.0$ & $14.4 \pm 1.7$ & $14.4 \pm 2.1$ & $10.6 \pm 1.3$ & $10.6 \pm 1.5$ & $13.0$ & $15.9 \pm 2.5$ & $15.9 \pm 2.2$ & $15.5 \pm 2.5$ & $5.0$ \\
& center & $13.0$ & $9.0$ & $13.0$ & $10.5 \pm 1.5$ & $10.9 \pm 1.4$ & $7.8 \pm 2.6$ & $8.0 \pm 1.7$ & $9.0$ & $7.7 \pm 1.1$ & $11.4 \pm 1.9$ & $9.2 \pm 1.6$ & $5.0$ & $33.0$ & $29.0$ & $29.0$ & $28.3 \pm 1.6$ & $22.8 \pm 2.3$ & $20.0 \pm 2.7$ & $18.3 \pm 2.0$ & $32.0$ & $28.7 \pm 2.3$ & $21.7 \pm 2.3$ & $20.0 \pm 2.5$ & $15.0$ \\
& diameter & $23.0$ & $22.0$ & $24.0$ & $25.9 \pm 2.4$ & $26.2 \pm 3.7$ & $21.4 \pm 3.1$ & $18.9 \pm 2.6$ & $18.0$ & $22.2 \pm 2.3$ & $23.0 \pm 2.9$ & $21.2 \pm 3.9$ & $25.0$ & $43.0$ & $43.0$ & $36.0$ & $43.1 \pm 1.9$ & $41.0 \pm 3.7$ & $36.5 \pm 3.5$ & $35.4 \pm 4.9$ & $43.0$ & $39.6 \pm 2.4$ & $40.9 \pm 3.4$ & $36.3 \pm 1.9$ & $22.0$ \\
& barycenter & $21.0$ & $22.0$ & $26.0$ & $23.2 \pm 2.7$ & $29.1 \pm 4.4$ & $21.1 \pm 2.8$ & $23.5 \pm 3.5$ & $29.0$ & $23.2 \pm 4.8$ & $22.8 \pm 3.3$ & $20.4 \pm 3.3$ & $13.0$ & $51.0$ & $41.0$ & $40.0$ & $43.3 \pm 3.2$ & $41.6 \pm 2.2$ & $37.4 \pm 3.7$ & $31.6 \pm 3.3$ & $45.0$ & $44.1 \pm 3.4$ & $42.4 \pm 3.2$ & $38.8 \pm 3.5$ & $24.0$ \\
& radius & $37.0$ & $30.0$ & $31.0$ & $34.3 \pm 2.8$ & $33.3 \pm 3.4$ & $33.4 \pm 5.0$ & $29.7 \pm 2.8$ & $29.0$ & $30.4 \pm 4.1$ & $31.6 \pm 2.8$ & $29.4 \pm 1.7$ & $34.0$ & $68.0$ & $67.0$ & $47.0$ & $68.4 \pm 3.6$ & $51.5 \pm 2.8$ & $56.3 \pm 4.0$ & $47.7 \pm 1.6$ & $60.0$ & $56.6 \pm 3.7$ & $60.7 \pm 2.6$ & $55.0 \pm 4.2$ & $43.0$ \\
& topological\_sort & $28.0$ & $29.0$ & $29.0$ & $29.3 \pm 2.8$ & $29.1 \pm 1.7$ & $25.5 \pm 3.2$ & $24.1 \pm 3.7$ & $5.0$ & $4.8 \pm 0.9$ & $1.1 \pm 1.1$ & $1.6 \pm 1.6$ & $5.0$ & $79.0$ & $74.0$ & $76.0$ & $73.4 \pm 2.0$ & $72.2 \pm 1.4$ & $74.3 \pm 2.2$ & $74.7 \pm 4.5$ & $62.0$ & $56.0 \pm 1.8$ & $64.0 \pm 3.8$ & $66.6 \pm 3.3$ & $16.0$ \\
& periphery & $13.0$ & $6.0$ & $10.0$ & $11.2 \pm 1.8$ & $8.6 \pm 1.6$ & $7.1 \pm 1.5$ & $5.5 \pm 1.6$ & $12.0$ & $10.0 \pm 1.8$ & $10.5 \pm 2.5$ & $8.9 \pm 1.5$ & $9.0$ & $36.0$ & $34.0$ & $23.0$ & $32.0 \pm 2.4$ & $24.2 \pm 3.1$ & $20.5 \pm 3.1$ & $19.2 \pm 3.9$ & $26.0$ & $28.5 \pm 3.3$ & $20.5 \pm 2.8$ & $20.2 \pm 2.9$ & $15.0$ \\
& betweenness\_centrality & $3.0$ & $2.0$ & $2.0$ & $3.2 \pm 1.6$ & $1.6 \pm 1.1$ & $1.5 \pm 1.0$ & $2.0 \pm 1.1$ & $5.0$ & $4.0 \pm 2.1$ & $3.6 \pm 1.3$ & $3.5 \pm 1.5$ & $4.0$ & $39.0$ & $39.0$ & $39.0$ & $38.8 \pm 0.6$ & $37.9 \pm 0.7$ & $36.4 \pm 1.5$ & $35.9 \pm 1.4$ & $39.0$ & $38.7 \pm 0.7$ & $38.9 \pm 0.3$ & $38.5 \pm 1.0$ & $38.0$ \\
& triangles & $25.0$ & $32.0$ & $35.0$ & $27.9 \pm 3.2$ & $26.5 \pm 2.8$ & $20.3 \pm 2.3$ & $21.6 \pm 3.0$ & $40.0$ & $34.2 \pm 5.3$ & $33.9 \pm 3.2$ & $29.9 \pm 3.1$ & $21.0$ & $83.0$ & $83.0$ & $78.0$ & $72.5 \pm 2.0$ & $64.2 \pm 3.4$ & $44.4 \pm 4.2$ & $56.8 \pm 5.0$ & $55.0$ & $57.7 \pm 2.8$ & $53.6 \pm 3.0$ & $53.7 \pm 3.1$ & $28.0$ \\
& avg\_neighbor\_degree & $32.0$ & $36.0$ & $40.0$ & $36.4 \pm 2.8$ & $35.2 \pm 3.6$ & $25.6 \pm 3.6$ & $18.4 \pm 4.0$ & $68.0$ & $60.2 \pm 2.8$ & $67.7 \pm 3.4$ & $61.4 \pm 5.0$ & $11.0$ & $85.0$ & $87.0$ & $25.0$ & $77.5 \pm 3.1$ & $16.3 \pm 2.0$ & $48.2 \pm 3.4$ & $18.4 \pm 4.0$ & $92.0$ & $88.2 \pm 2.1$ & $90.2 \pm 1.5$ & $88.0 \pm 1.6$ & $23.0$ \\
& harmonic\_centrality & $4.0$ & $6.0$ & $7.0$ & $5.9 \pm 2.2$ & $4.9 \pm 2.1$ & $4.4 \pm 1.4$ & $4.8 \pm 1.3$ & $7.0$ & $6.2 \pm 1.0$ & $5.4 \pm 2.0$ & $5.1 \pm 1.7$ & $3.0$ & $30.0$ & $26.0$ & $21.0$ & $26.0 \pm 2.2$ & $22.8 \pm 2.5$ & $20.0 \pm 2.8$ & $16.4 \pm 1.8$ & $30.0$ & $28.5 \pm 2.5$ & $25.7 \pm 2.8$ & $22.4 \pm 3.5$ & $8.0$ \\
& bridges & $3.0$ & $7.0$ & $4.0$ & $3.4 \pm 1.8$ & $2.0 \pm 1.6$ & $1.2 \pm 0.9$ & $0.8 \pm 0.9$ & $1.0$ & $1.2 \pm 0.8$ & $2.4 \pm 2.0$ & $1.2 \pm 1.0$ & $3.0$ & $22.0$ & $22.0$ & $19.0$ & $18.5 \pm 2.6$ & $19.3 \pm 1.8$ & $5.6 \pm 1.6$ & $7.3 \pm 1.3$ & $17.0$ & $18.0 \pm 1.6$ & $13.9 \pm 2.0$ & $13.5 \pm 1.4$ & $13.0$ \\
\midrule
\multirow{6}{*}{\rotatebox{90}{\textbf{Challenging}}} & global\_efficiency & $2.0$ & $1.0$ & $1.0$ & $1.5 \pm 0.5$ & $1.2 \pm 0.8$ & $1.2 \pm 1.0$ & $0.5 \pm 0.5$ & $1.0$ & $1.5 \pm 0.8$ & $1.9 \pm 0.7$ & $2.0 \pm 0.8$ & $1.0$ & $1.0$ & $1.0$ & $2.0$ & $1.6 \pm 0.7$ & $1.7 \pm 0.7$ & $1.3 \pm 0.9$ & $2.3 \pm 0.8$ & $1.0$ & $1.5 \pm 1.0$ & $1.4 \pm 0.8$ & $1.4 \pm 0.7$ & $2.0$ \\
& maximal\_independent\_set & $5.0$ & $3.0$ & $5.0$ & $2.6 \pm 1.1$ & $3.1 \pm 1.7$ & $3.3 \pm 1.3$ & $2.6 \pm 1.6$ & $7.0$ & $5.5 \pm 1.1$ & $6.3 \pm 1.8$ & $5.0 \pm 1.4$ & $2.0$ & $81.0$ & $90.0$ & $81.0$ & $77.8 \pm 1.6$ & $70.0 \pm 3.4$ & $44.1 \pm 2.6$ & $45.1 \pm 3.8$ & $85.0$ & $75.9 \pm 4.0$ & $80.7 \pm 3.9$ & $67.6 \pm 3.4$ & $13.0$ \\
& maximum\_flow & $5.0$ & $5.0$ & $6.0$ & $5.9 \pm 1.4$ & $5.8 \pm 1.1$ & $4.8 \pm 1.8$ & $4.8 \pm 2.1$ & $7.0$ & $5.8 \pm 1.3$ & $3.7 \pm 1.8$ & $3.8 \pm 1.5$ & $6.0$ & $12.0$ & $12.0$ & $12.0$ & $10.7 \pm 1.6$ & $11.0 \pm 1.6$ & $9.1 \pm 2.0$ & $8.7 \pm 1.3$ & $11.0$ & $9.2 \pm 1.9$ & $9.3 \pm 1.8$ & $9.5 \pm 1.6$ & $5.0$ \\
& wiener\_index & $6.0$ & $5.0$ & $4.0$ & $4.2 \pm 1.2$ & $5.3 \pm 1.6$ & $3.1 \pm 1.8$ & $2.5 \pm 0.8$ & $7.0$ & $5.6 \pm 2.0$ & $4.6 \pm 1.2$ & $4.9 \pm 0.9$ & $4.0$ & $15.0$ & $11.0$ & $10.0$ & $15.2 \pm 1.7$ & $10.0 \pm 2.3$ & $9.9 \pm 2.3$ & $7.5 \pm 2.5$ & $10.0$ & $12.1 \pm 2.3$ & $11.9 \pm 2.2$ & $10.1 \pm 2.4$ & $12.0$ \\
& hamiltonian\_path & $1.0$ & $2.0$ & $0.0$ & $0.8 \pm 0.6$ & $0.7 \pm 0.5$ & $1.1 \pm 1.0$ & $0.8 \pm 0.8$ & $1.0$ & $0.3 \pm 0.5$ & $0.8 \pm 0.8$ & $0.2 \pm 0.4$ & $1.0$ & $3.0$ & $2.0$ & $2.0$ & $2.2 \pm 0.6$ & $1.8 \pm 0.6$ & $1.4 \pm 0.5$ & $1.4 \pm 0.5$ & $5.0$ & $1.1 \pm 0.6$ & $2.1 \pm 1.0$ & $1.1 \pm 0.7$ & $1.0$ \\
& min\_vertex\_cover & $6.0$ & $10.0$ & $8.0$ & $7.3 \pm 1.2$ & $5.7 \pm 1.8$ & $6.7 \pm 1.9$ & $4.3 \pm 1.8$ & $4.0$ & $4.3 \pm 1.6$ & $5.4 \pm 1.5$ & $4.6 \pm 1.8$ & $4.0$ & $42.0$ & $35.0$ & $28.0$ & $34.6 \pm 2.5$ & $23.2 \pm 2.7$ & $28.2 \pm 2.9$ & $16.6 \pm 3.6$ & $21.0$ & $17.6 \pm 2.0$ & $17.1 \pm 3.3$ & $17.5 \pm 1.8$ & $12.0$ \\
\bottomrule
\end{tabular}
}
    }
    \label{tab:edge_reorderings_all_7B}
\end{sidewaystable}

\begin{sidewaystable}
    \small
    \centering
    \captionsetup{margin=1cm}
    \caption{Edge reordering ablation results for the \GPT models. We report the accuracy in \%. \Erdos baseline (w/ zero temperature) is shaded. For the computational structure, \texttt{edge} = edge list, \texttt{adjl} = adj.\ list, \texttt{adjm} = adj.\ matrix. For sorting, \texttt{s} = source and \texttt{t} = target, and the entry represents the sorting key. Shuffling is applied to all remaining ambiguities; in this case we report the mean accuracy $\pm$ std over 10 seeds (20B) or one seed (120B).}
    \resizebox{\linewidth}{!}{
        {\renewcommand{\arraystretch}{1.2}
\begin{tabular}{ll >{\columncolor{gray!20}}rrrrrrrrrrrr : >{\columncolor{gray!20}}rrrrrrrrrrrr}

& & \multicolumn{12}{c:}{\large\textbf{gpt-oss-20B}} & \multicolumn{12}{:c}{\large\textbf{gpt-oss-120B}} \\[2ex]
\toprule
& \textbf{Comp. structure} & \texttt{edge} & \texttt{edge} & \texttt{edge} & \texttt{edge} & \texttt{edge} & \texttt{edge} & \texttt{edge} & \texttt{adjl} & \texttt{adjl} & \texttt{adjl} & \texttt{adjl} & \texttt{adjm} & \texttt{edge} & \texttt{edge} & \texttt{edge} & \texttt{edge} & \texttt{edge} & \texttt{edge} & \texttt{edge} & \texttt{adjl} & \texttt{adjl} & \texttt{adjl} & \texttt{adjl} & \texttt{adjm} \\
& \textbf{Sort (rest shuffled)} & \Erdos & \texttt{(s,t)} & \texttt{(s,t)} & \texttt{s} & \texttt{s} & \xmark & \xmark & \texttt{(s,t)} & \texttt{s} & \texttt{t} & \xmark & \texttt{NA} & \Erdos & \texttt{(s,t)} & \texttt{(s,t)} & \texttt{s} & \texttt{s} & \xmark & \xmark & \texttt{(s,t)} & \texttt{s} & \texttt{t} & \xmark & \texttt{NA} \\
& \textbf{Repl. undir} & \xmark & \xmark & \cmark & \xmark & \cmark & \xmark & \cmark & (\cmark) & (\cmark) & (\cmark) & (\cmark) & (\cmark) & \xmark & \xmark & \cmark & \xmark & \cmark & \xmark & \cmark & (\cmark) & (\cmark) & (\cmark) & (\cmark) & (\cmark) \\

\midrule
\multirow{15}{*}{\rotatebox{90}{\textbf{Easy}}} & node\_number & $100.0$ & $100.0$ & $100.0$ & $100.0 \pm 0.0$ & $100.0 \pm 0.0$ & $100.0 \pm 0.0$ & $99.8 \pm 0.4$ & $100.0$ & $100.0 \pm 0.0$ & $100.0 \pm 0.0$ & $100.0 \pm 0.0$ & $99.0$ & $100.0$ & $100.0$ & $100.0$ & $100.0 \pm 0.0$ & $100.0 \pm 0.0$ & $100.0 \pm 0.0$ & $100.0 \pm 0.0$ & $100.0$ & $100.0 \pm 0.0$ & $100.0 \pm 0.0$ & $100.0 \pm 0.0$ & $100.0$ \\
& dominating\_set & $54.0$ & $59.0$ & $53.0$ & $57.6 \pm 4.8$ & $61.0 \pm 2.9$ & $55.1 \pm 3.8$ & $43.3 \pm 3.5$ & $65.0$ & $59.2 \pm 3.6$ & $61.6 \pm 3.6$ & $59.6 \pm 3.7$ & $42.0$ & $93.0$ & $92.0$ & $90.0$ & $92.9 \pm 2.3$ & $93.7 \pm 2.3$ & $89.4 \pm 3.7$ & $84.7 \pm 5.1$ & $94.0$ & $92.3 \pm 3.3$ & $91.3 \pm 1.8$ & $90.7 \pm 3.9$ & $65.0$ \\
& common\_neighbor & $97.0$ & $94.0$ & $99.0$ & $95.4 \pm 1.4$ & $95.7 \pm 2.1$ & $91.5 \pm 2.7$ & $78.9 \pm 2.4$ & $100.0$ & $98.7 \pm 1.2$ & $99.3 \pm 1.1$ & $98.1 \pm 0.7$ & $89.0$ & $99.0$ & $98.0$ & $100.0$ & $99.7 \pm 0.7$ & $99.9 \pm 0.3$ & $98.6 \pm 1.3$ & $95.2 \pm 2.7$ & $100.0$ & $99.9 \pm 0.3$ & $100.0 \pm 0.0$ & $100.0 \pm 0.0$ & $95.0$ \\
& edge\_number & $91.0$ & $97.0$ & $73.0$ & $93.3 \pm 0.9$ & $71.8 \pm 4.3$ & $80.9 \pm 2.0$ & $50.7 \pm 3.8$ & $91.0$ & $85.0 \pm 2.9$ & $84.8 \pm 3.2$ & $80.8 \pm 3.5$ & $62.0$ & $90.0$ & $91.0$ & $86.0$ & $89.9 \pm 0.3$ & $83.7 \pm 2.1$ & $89.9 \pm 0.7$ & $67.3 \pm 3.6$ & $90.0$ & $89.7 \pm 0.5$ & $89.3 \pm 1.2$ & $87.9 \pm 1.8$ & $68.0$ \\
& neighbor & $62.0$ & $70.0$ & $69.0$ & $65.4 \pm 2.7$ & $66.1 \pm 4.2$ & $68.5 \pm 4.4$ & $75.2 \pm 4.4$ & $74.0$ & $68.0 \pm 1.9$ & $72.0 \pm 3.1$ & $59.8 \pm 3.7$ & $78.0$ & $100.0$ & $100.0$ & $100.0$ & $99.5 \pm 0.7$ & $99.6 \pm 0.5$ & $99.9 \pm 0.3$ & $98.7 \pm 1.1$ & $99.0$ & $100.0 \pm 0.0$ & $100.0 \pm 0.0$ & $100.0 \pm 0.0$ & $95.0$ \\
& bfs & $86.0$ & $88.0$ & $95.0$ & $84.9 \pm 2.4$ & $92.1 \pm 2.2$ & $84.9 \pm 2.0$ & $65.2 \pm 4.9$ & $96.0$ & $96.2 \pm 1.0$ & $97.0 \pm 1.6$ & $95.9 \pm 2.0$ & $58.0$ & $94.0$ & $95.0$ & $95.0$ & $93.2 \pm 2.1$ & $98.0 \pm 1.5$ & $92.0 \pm 2.4$ & $87.7 \pm 3.1$ & $95.0$ & $96.1 \pm 2.1$ & $96.3 \pm 1.3$ & $96.8 \pm 2.3$ & $83.0$ \\
& has\_cycle & $96.0$ & $93.0$ & $88.0$ & $95.0 \pm 2.4$ & $89.6 \pm 3.1$ & $93.5 \pm 1.4$ & $78.6 \pm 3.7$ & $94.0$ & $93.3 \pm 1.6$ & $90.8 \pm 2.3$ & $89.5 \pm 3.8$ & $80.0$ & $100.0$ & $100.0$ & $100.0$ & $99.6 \pm 0.5$ & $99.4 \pm 0.8$ & $99.7 \pm 0.5$ & $94.3 \pm 1.6$ & $99.0$ & $100.0 \pm 0.0$ & $100.0 \pm 0.0$ & $99.7 \pm 0.5$ & $92.0$ \\
& dfs & $79.0$ & $84.0$ & $84.0$ & $80.3 \pm 2.2$ & $84.0 \pm 2.6$ & $80.5 \pm 3.5$ & $65.4 \pm 2.8$ & $85.0$ & $89.1 \pm 3.8$ & $88.9 \pm 2.8$ & $88.4 \pm 2.4$ & $63.0$ & $93.0$ & $91.0$ & $93.0$ & $92.3 \pm 2.5$ & $93.4 \pm 2.6$ & $91.8 \pm 3.4$ & $86.1 \pm 3.6$ & $95.0$ & $94.9 \pm 2.2$ & $94.8 \pm 2.0$ & $95.5 \pm 1.2$ & $72.0$ \\
& minimum\_spanning\_tree & $83.0$ & $85.0$ & $83.0$ & $88.1 \pm 3.9$ & $81.2 \pm 2.9$ & $78.8 \pm 3.3$ & $65.5 \pm 2.9$ & $87.0$ & $87.3 \pm 3.2$ & $86.0 \pm 2.4$ & $82.8 \pm 2.9$ & $79.0$ & $99.0$ & $100.0$ & $97.0$ & $99.4 \pm 1.1$ & $98.2 \pm 1.2$ & $95.0 \pm 2.4$ & $89.9 \pm 2.1$ & $98.0$ & $98.8 \pm 1.2$ & $98.8 \pm 0.8$ & $96.5 \pm 1.9$ & $95.0$ \\
& weighted\_minimum\_spanning\_tree & $66.0$ & $61.0$ & $46.0$ & $60.5 \pm 3.3$ & $39.4 \pm 3.1$ & $58.0 \pm 2.9$ & $33.0 \pm 1.4$ & $49.0$ & $41.3 \pm 2.8$ & $35.4 \pm 4.0$ & $33.9 \pm 2.4$ & $34.0$ & $90.0$ & $86.0$ & $88.0$ & $81.4 \pm 2.5$ & $65.7 \pm 3.6$ & $79.1 \pm 3.2$ & $60.5 \pm 2.6$ & $85.0$ & $67.9 \pm 7.8$ & $60.3 \pm 7.3$ & $59.7 \pm 7.7$ & $39.0$ \\
& edge\_existence & $99.0$ & $100.0$ & $100.0$ & $100.0 \pm 0.0$ & $100.0 \pm 0.0$ & $99.6 \pm 0.7$ & $98.5 \pm 1.0$ & $100.0$ & $100.0 \pm 0.0$ & $100.0 \pm 0.0$ & $100.0 \pm 0.0$ & $98.0$ & $100.0$ & $99.0$ & $99.0$ & $99.9 \pm 0.3$ & $100.0 \pm 0.0$ & $99.8 \pm 0.4$ & $100.0 \pm 0.0$ & $100.0$ & $100.0 \pm 0.0$ & $100.0 \pm 0.0$ & $100.0 \pm 0.0$ & $99.0$ \\
& is\_regular & $100.0$ & $100.0$ & $98.0$ & $99.3 \pm 1.1$ & $93.3 \pm 1.3$ & $96.4 \pm 1.8$ & $76.8 \pm 3.1$ & $99.0$ & $99.8 \pm 0.4$ & $99.0 \pm 0.8$ & $98.8 \pm 0.9$ & $96.0$ & $100.0$ & $100.0$ & $100.0$ & $100.0 \pm 0.0$ & $99.8 \pm 0.4$ & $99.1 \pm 0.6$ & $89.2 \pm 1.9$ & $100.0$ & $100.0 \pm 0.0$ & $100.0 \pm 0.0$ & $100.0 \pm 0.0$ & $100.0$ \\
& degree & $97.0$ & $98.0$ & $98.0$ & $98.1 \pm 0.3$ & $97.6 \pm 1.0$ & $98.3 \pm 0.8$ & $91.8 \pm 2.5$ & $98.0$ & $98.0 \pm 0.0$ & $98.0 \pm 0.0$ & $98.1 \pm 0.3$ & $94.0$ & $98.0$ & $98.0$ & $98.0$ & $98.0 \pm 0.0$ & $98.0 \pm 0.0$ & $97.9 \pm 0.3$ & $95.7 \pm 0.8$ & $98.0$ & $98.0 \pm 0.0$ & $98.0 \pm 0.0$ & $98.0 \pm 0.0$ & $96.0$ \\
& is\_tournament & $100.0$ & $100.0$ & $99.0$ & $99.7 \pm 0.5$ & $99.9 \pm 0.3$ & $99.5 \pm 0.5$ & $99.8 \pm 0.4$ & $100.0$ & $99.9 \pm 0.3$ & $99.9 \pm 0.3$ & $99.6 \pm 0.5$ & $100.0$ & $100.0$ & $100.0$ & $100.0$ & $100.0 \pm 0.0$ & $100.0 \pm 0.0$ & $100.0 \pm 0.0$ & $100.0 \pm 0.0$ & $100.0$ & $100.0 \pm 0.0$ & $100.0 \pm 0.0$ & $100.0 \pm 0.0$ & $100.0$ \\
& density & $95.0$ & $99.0$ & $75.0$ & $97.4 \pm 1.6$ & $76.7 \pm 4.4$ & $91.2 \pm 2.7$ & $58.5 \pm 5.2$ & $91.0$ & $87.6 \pm 4.5$ & $85.4 \pm 3.8$ & $77.0 \pm 2.7$ & $65.0$ & $97.0$ & $99.0$ & $90.0$ & $96.7 \pm 1.4$ & $89.3 \pm 2.9$ & $97.1 \pm 1.7$ & $80.9 \pm 4.8$ & $92.0$ & $90.5 \pm 3.6$ & $92.3 \pm 3.0$ & $90.2 \pm 3.5$ & $86.0$ \\
    \midrule
\multirow{11}{*}{\rotatebox{90}{\textbf{Medium}}} & adamic\_adar\_index & $99.0$ & $100.0$ & $100.0$ & $98.8 \pm 0.8$ & $96.5 \pm 1.5$ & $96.0 \pm 1.9$ & $74.3 \pm 4.0$ & $100.0$ & $99.8 \pm 0.4$ & $99.8 \pm 0.4$ & $99.2 \pm 0.6$ & $94.0$ & $99.0$ & $98.0$ & $100.0$ & $98.8 \pm 0.6$ & $98.8 \pm 0.9$ & $98.6 \pm 1.3$ & $91.8 \pm 2.2$ & $100.0$ & $99.4 \pm 1.1$ & $99.4 \pm 0.7$ & $99.6 \pm 0.7$ & $95.0$ \\
& clustering\_coefficient & $91.0$ & $93.0$ & $90.0$ & $91.2 \pm 1.8$ & $90.7 \pm 2.2$ & $91.8 \pm 2.3$ & $74.7 \pm 3.4$ & $97.0$ & $95.3 \pm 1.8$ & $97.2 \pm 0.9$ & $94.4 \pm 1.1$ & $89.0$ & $96.0$ & $98.0$ & $98.0$ & $96.3 \pm 1.4$ & $95.6 \pm 1.4$ & $96.9 \pm 1.2$ & $89.2 \pm 2.0$ & $100.0$ & $97.1 \pm 1.5$ & $98.6 \pm 1.3$ & $96.9 \pm 1.5$ & $85.0$ \\
& connected\_component\_number & $98.0$ & $98.0$ & $98.0$ & $96.3 \pm 1.3$ & $96.7 \pm 1.6$ & $89.6 \pm 2.5$ & $65.5 \pm 3.8$ & $98.0$ & $98.6 \pm 0.8$ & $97.3 \pm 1.9$ & $97.2 \pm 1.5$ & $75.0$ & $100.0$ & $100.0$ & $99.0$ & $99.0 \pm 0.8$ & $99.1 \pm 0.3$ & $99.0 \pm 0.9$ & $90.9 \pm 1.5$ & $100.0$ & $99.7 \pm 0.5$ & $99.6 \pm 0.5$ & $99.2 \pm 1.0$ & $83.0$ \\
& bipartite\_maximum\_matching & $84.0$ & $88.0$ & $81.0$ & $77.6 \pm 2.6$ & $72.0 \pm 2.3$ & $42.0 \pm 3.0$ & $25.8 \pm 4.0$ & $79.0$ & $69.4 \pm 4.4$ & $72.8 \pm 4.1$ & $65.4 \pm 3.1$ & $67.0$ & $96.0$ & $94.0$ & $94.0$ & $95.6 \pm 1.5$ & $94.2 \pm 2.3$ & $69.7 \pm 6.7$ & $72.1 \pm 5.7$ & $98.0$ & $95.5 \pm 2.2$ & $95.5 \pm 1.9$ & $92.6 \pm 3.3$ & $83.0$ \\
& local\_connectivity & $99.0$ & $99.0$ & $98.0$ & $98.5 \pm 1.3$ & $98.9 \pm 1.0$ & $97.2 \pm 1.8$ & $80.9 \pm 2.8$ & $100.0$ & $99.6 \pm 0.7$ & $99.4 \pm 0.7$ & $99.4 \pm 0.8$ & $84.0$ & $99.0$ & $99.0$ & $99.0$ & $99.4 \pm 0.5$ & $99.2 \pm 0.4$ & $98.4 \pm 0.7$ & $93.7 \pm 2.0$ & $100.0$ & $100.0 \pm 0.0$ & $99.6 \pm 0.5$ & $99.5 \pm 0.5$ & $93.0$ \\
& jaccard\_coefficient & $100.0$ & $100.0$ & $97.0$ & $99.3 \pm 0.7$ & $99.4 \pm 0.7$ & $97.8 \pm 1.4$ & $81.6 \pm 3.5$ & $100.0$ & $100.0 \pm 0.0$ & $100.0 \pm 0.0$ & $100.0 \pm 0.0$ & $96.0$ & $100.0$ & $100.0$ & $100.0$ & $99.9 \pm 0.3$ & $100.0 \pm 0.0$ & $99.9 \pm 0.3$ & $95.2 \pm 2.2$ & $100.0$ & $100.0 \pm 0.0$ & $99.9 \pm 0.3$ & $99.7 \pm 0.5$ & $95.0$ \\
& min\_edge\_covering & $30.0$ & $25.0$ & $28.0$ & $27.7 \pm 3.6$ & $27.2 \pm 1.7$ & $30.4 \pm 4.1$ & $26.8 \pm 4.1$ & $26.0$ & $26.6 \pm 2.2$ & $28.7 \pm 2.9$ & $28.2 \pm 2.9$ & $26.0$ & $93.0$ & $94.0$ & $95.0$ & $94.3 \pm 3.3$ & $95.8 \pm 1.6$ & $91.5 \pm 2.4$ & $85.8 \pm 3.3$ & $93.0$ & $94.1 \pm 2.7$ & $95.4 \pm 1.8$ & $93.1 \pm 2.7$ & $81.0$ \\
& is\_eularian & $97.0$ & $95.0$ & $90.0$ & $96.8 \pm 1.7$ & $83.0 \pm 3.4$ & $92.9 \pm 2.1$ & $59.6 \pm 3.1$ & $96.0$ & $98.2 \pm 1.1$ & $94.5 \pm 2.0$ & $97.7 \pm 2.1$ & $90.0$ & $100.0$ & $100.0$ & $100.0$ & $99.9 \pm 0.3$ & $97.6 \pm 1.2$ & $97.7 \pm 0.9$ & $80.6 \pm 4.4$ & $100.0$ & $99.5 \pm 0.7$ & $99.8 \pm 0.4$ & $98.9 \pm 0.9$ & $100.0$ \\
& degree\_centrality & $90.0$ & $92.0$ & $92.0$ & $89.9 \pm 1.3$ & $90.5 \pm 1.8$ & $91.0 \pm 0.9$ & $84.4 \pm 3.1$ & $89.0$ & $88.7 \pm 0.5$ & $89.3 \pm 0.8$ & $89.5 \pm 0.8$ & $83.0$ & $93.0$ & $92.0$ & $93.0$ & $92.7 \pm 1.2$ & $92.6 \pm 0.8$ & $92.4 \pm 1.1$ & $91.0 \pm 1.3$ & $89.0$ & $93.4 \pm 1.4$ & $92.4 \pm 1.0$ & $92.6 \pm 1.6$ & $90.0$ \\
& is\_bipartite & $90.0$ & $87.0$ & $93.0$ & $92.4 \pm 1.2$ & $93.1 \pm 2.4$ & $89.3 \pm 5.3$ & $72.0 \pm 2.3$ & $90.0$ & $93.6 \pm 2.0$ & $91.1 \pm 3.2$ & $91.1 \pm 2.2$ & $73.0$ & $100.0$ & $100.0$ & $100.0$ & $99.4 \pm 0.7$ & $99.6 \pm 0.5$ & $99.1 \pm 1.0$ & $95.1 \pm 1.8$ & $99.0$ & $99.4 \pm 0.5$ & $99.8 \pm 0.4$ & $99.9 \pm 0.3$ & $91.0$ \\
& resource\_allocation\_index & $94.0$ & $95.0$ & $99.0$ & $93.4 \pm 1.0$ & $95.7 \pm 1.6$ & $92.4 \pm 1.5$ & $64.6 \pm 3.1$ & $99.0$ & $99.1 \pm 0.7$ & $99.4 \pm 0.5$ & $99.2 \pm 0.8$ & $94.0$ & $97.0$ & $97.0$ & $100.0$ & $97.1 \pm 0.6$ & $98.1 \pm 0.6$ & $96.7 \pm 1.8$ & $84.7 \pm 2.1$ & $100.0$ & $99.6 \pm 0.7$ & $99.5 \pm 0.7$ & $99.1 \pm 0.7$ & $97.0$ \\
    \midrule
\multirow{17}{*}{\rotatebox{90}{\textbf{Hard}}} & max\_weight\_matching & $60.0$ & $58.0$ & $57.0$ & $61.7 \pm 4.4$ & $54.9 \pm 3.7$ & $58.2 \pm 4.0$ & $46.2 \pm 2.3$ & $63.0$ & $61.8 \pm 2.8$ & $56.6 \pm 3.7$ & $58.8 \pm 3.1$ & $46.0$ & $97.0$ & $96.0$ & $98.0$ & $95.0 \pm 2.7$ & $96.5 \pm 1.9$ & $94.0 \pm 2.6$ & $90.5 \pm 2.5$ & $97.0$ & $96.4 \pm 2.2$ & $96.6 \pm 2.3$ & $95.2 \pm 2.3$ & $74.0$ \\
& closeness\_centrality & $30.0$ & $28.0$ & $39.0$ & $26.3 \pm 3.5$ & $34.3 \pm 2.0$ & $32.9 \pm 2.5$ & $22.5 \pm 1.5$ & $33.0$ & $35.5 \pm 1.6$ & $36.2 \pm 1.5$ & $36.0 \pm 1.9$ & $21.0$ & $41.0$ & $40.0$ & $41.0$ & $36.6 \pm 1.4$ & $40.3 \pm 0.9$ & $37.9 \pm 2.7$ & $36.3 \pm 2.9$ & $37.0$ & $39.9 \pm 1.2$ & $39.6 \pm 1.6$ & $38.7 \pm 1.8$ & $30.0$ \\
& traveling\_salesman\_problem & $51.0$ & $61.0$ & $56.0$ & $55.6 \pm 2.2$ & $55.7 \pm 1.8$ & $53.7 \pm 2.4$ & $52.5 \pm 1.8$ & $60.0$ & $56.8 \pm 2.3$ & $58.9 \pm 3.3$ & $56.9 \pm 2.6$ & $55.0$ & $96.0$ & $94.0$ & $96.0$ & $87.7 \pm 1.3$ & $87.6 \pm 2.5$ & $88.1 \pm 2.8$ & $82.2 \pm 2.9$ & $91.0$ & $88.7 \pm 2.0$ & $89.2 \pm 3.0$ & $89.4 \pm 3.2$ & $86.0$ \\
& strongly\_connected\_number & $65.0$ & $70.0$ & $73.0$ & $70.4 \pm 2.3$ & $69.7 \pm 1.9$ & $64.7 \pm 1.9$ & $63.7 \pm 4.4$ & $74.0$ & $73.8 \pm 2.8$ & $74.4 \pm 2.8$ & $72.4 \pm 3.0$ & $59.0$ & $95.0$ & $94.0$ & $97.0$ & $95.9 \pm 2.3$ & $95.5 \pm 2.0$ & $91.1 \pm 3.1$ & $91.2 \pm 2.5$ & $95.0$ & $96.7 \pm 1.9$ & $97.3 \pm 1.3$ & $96.6 \pm 1.6$ & $85.0$ \\
& shortest\_path & $77.0$ & $78.0$ & $79.0$ & $73.5 \pm 2.9$ & $77.7 \pm 2.9$ & $74.3 \pm 3.1$ & $55.9 \pm 5.9$ & $65.0$ & $73.3 \pm 3.7$ & $69.5 \pm 5.6$ & $71.4 \pm 2.8$ & $58.0$ & $95.0$ & $99.0$ & $98.0$ & $97.6 \pm 1.2$ & $97.3 \pm 1.6$ & $97.7 \pm 1.6$ & $93.7 \pm 2.3$ & $97.0$ & $96.6 \pm 2.2$ & $96.8 \pm 2.0$ & $96.8 \pm 1.8$ & $75.0$ \\
& weighted\_shortest\_path & $42.0$ & $36.0$ & $50.0$ & $35.6 \pm 2.5$ & $40.9 \pm 1.4$ & $20.8 \pm 2.0$ & $17.2 \pm 1.5$ & $46.0$ & $46.4 \pm 3.2$ & $47.9 \pm 5.7$ & $47.1 \pm 4.0$ & $28.0$ & $77.0$ & $86.0$ & $79.0$ & $77.6 \pm 2.7$ & $78.0 \pm 2.9$ & $47.0 \pm 6.4$ & $44.4 \pm 3.5$ & $81.0$ & $80.9 \pm 3.2$ & $84.3 \pm 2.7$ & $84.5 \pm 3.5$ & $71.0$ \\
& center & $47.0$ & $50.0$ & $51.0$ & $47.1 \pm 3.2$ & $49.3 \pm 4.8$ & $49.6 \pm 3.8$ & $40.7 \pm 2.7$ & $46.0$ & $49.9 \pm 1.9$ & $48.8 \pm 4.9$ & $48.2 \pm 2.7$ & $48.0$ & $76.0$ & $82.0$ & $78.0$ & $77.5 \pm 2.1$ & $79.6 \pm 3.7$ & $77.5 \pm 4.5$ & $73.9 \pm 2.2$ & $79.0$ & $80.1 \pm 2.7$ & $84.0 \pm 3.0$ & $80.8 \pm 3.7$ & $67.0$ \\
& diameter & $57.0$ & $52.0$ & $51.0$ & $54.5 \pm 3.2$ & $52.8 \pm 1.7$ & $56.3 \pm 2.0$ & $48.7 \pm 2.3$ & $50.0$ & $55.6 \pm 2.6$ & $52.9 \pm 3.4$ & $54.2 \pm 2.4$ & $52.0$ & $93.0$ & $86.0$ & $91.0$ & $89.6 \pm 2.4$ & $90.2 \pm 1.9$ & $88.6 \pm 3.2$ & $83.0 \pm 2.5$ & $89.0$ & $89.6 \pm 2.4$ & $91.4 \pm 1.6$ & $91.3 \pm 2.8$ & $79.0$ \\
& barycenter & $47.0$ & $44.0$ & $41.0$ & $46.4 \pm 3.0$ & $42.9 \pm 2.6$ & $48.1 \pm 2.5$ & $39.1 \pm 1.7$ & $21.0$ & $26.4 \pm 2.0$ & $28.8 \pm 2.7$ & $33.2 \pm 3.3$ & $29.0$ & $90.0$ & $88.0$ & $93.0$ & $87.3 \pm 2.5$ & $85.2 \pm 3.6$ & $82.9 \pm 3.1$ & $72.2 \pm 3.8$ & $91.0$ & $89.3 \pm 2.7$ & $90.1 \pm 2.9$ & $89.5 \pm 3.4$ & $72.0$ \\
& radius & $65.0$ & $59.0$ & $67.0$ & $63.8 \pm 3.3$ & $62.5 \pm 2.5$ & $59.0 \pm 3.1$ & $49.2 \pm 3.1$ & $64.0$ & $62.0 \pm 3.6$ & $64.0 \pm 1.7$ & $63.7 \pm 2.8$ & $64.0$ & $94.0$ & $94.0$ & $97.0$ & $91.9 \pm 1.9$ & $92.8 \pm 1.5$ & $91.4 \pm 2.1$ & $83.7 \pm 3.5$ & $92.0$ & $91.8 \pm 1.0$ & $94.7 \pm 1.8$ & $92.8 \pm 3.4$ & $81.0$ \\
& topological\_sort & $78.0$ & $81.0$ & $81.0$ & $78.4 \pm 3.4$ & $74.5 \pm 3.7$ & $75.8 \pm 4.5$ & $75.4 \pm 2.5$ & $70.0$ & $67.0 \pm 2.9$ & $67.2 \pm 5.2$ & $67.5 \pm 4.0$ & $64.0$ & $97.0$ & $97.0$ & $98.0$ & $98.1 \pm 0.7$ & $98.8 \pm 1.5$ & $96.5 \pm 1.4$ & $98.3 \pm 1.1$ & $96.0$ & $97.6 \pm 1.6$ & $98.6 \pm 1.0$ & $97.3 \pm 1.5$ & $92.0$ \\
& periphery & $46.0$ & $49.0$ & $47.0$ & $47.4 \pm 3.5$ & $45.3 \pm 3.8$ & $47.6 \pm 4.2$ & $42.3 \pm 2.9$ & $47.0$ & $48.2 \pm 3.7$ & $47.4 \pm 3.5$ & $51.7 \pm 3.1$ & $41.0$ & $72.0$ & $73.0$ & $74.0$ & $70.4 \pm 2.1$ & $71.7 \pm 3.5$ & $71.5 \pm 3.0$ & $67.5 \pm 2.2$ & $70.0$ & $72.1 \pm 3.8$ & $74.3 \pm 4.0$ & $73.0 \pm 3.7$ & $63.0$ \\
& betweenness\_centrality & $30.0$ & $28.0$ & $30.0$ & $27.5 \pm 1.1$ & $29.5 \pm 2.2$ & $26.4 \pm 1.9$ & $21.9 \pm 1.4$ & $30.0$ & $30.6 \pm 2.1$ & $29.6 \pm 2.2$ & $27.7 \pm 1.9$ & $24.0$ & $39.0$ & $38.0$ & $39.0$ & $37.9 \pm 1.4$ & $38.5 \pm 1.0$ & $37.5 \pm 1.4$ & $36.0 \pm 1.9$ & $39.0$ & $38.6 \pm 0.7$ & $38.8 \pm 0.4$ & $38.3 \pm 0.7$ & $35.0$ \\
& triangles & $85.0$ & $93.0$ & $86.0$ & $86.3 \pm 1.7$ & $82.1 \pm 3.3$ & $82.0 \pm 2.9$ & $56.1 \pm 3.7$ & $93.0$ & $94.1 \pm 1.3$ & $94.4 \pm 2.5$ & $91.7 \pm 2.9$ & $77.0$ & $98.0$ & $96.0$ & $100.0$ & $97.3 \pm 0.9$ & $98.8 \pm 1.3$ & $97.3 \pm 1.5$ & $91.4 \pm 1.6$ & $100.0$ & $99.0 \pm 0.7$ & $99.3 \pm 0.9$ & $98.8 \pm 1.0$ & $89.0$ \\
& avg\_neighbor\_degree & $91.0$ & $89.0$ & $87.0$ & $90.6 \pm 2.0$ & $79.9 \pm 2.7$ & $87.3 \pm 2.5$ & $59.6 \pm 2.8$ & $98.0$ & $98.7 \pm 0.9$ & $99.0 \pm 0.7$ & $97.9 \pm 1.1$ & $73.0$ & $93.0$ & $93.0$ & $93.0$ & $92.4 \pm 1.2$ & $91.9 \pm 0.9$ & $92.7 \pm 1.3$ & $78.7 \pm 2.9$ & $95.0$ & $96.3 \pm 0.9$ & $96.0 \pm 1.2$ & $96.7 \pm 0.9$ & $89.0$ \\
& harmonic\_centrality & $81.0$ & $88.0$ & $82.0$ & $82.2 \pm 2.7$ & $85.9 \pm 3.6$ & $82.1 \pm 2.9$ & $68.1 \pm 2.8$ & $88.0$ & $86.5 \pm 3.5$ & $85.9 \pm 2.5$ & $87.2 \pm 3.7$ & $63.0$ & $95.0$ & $95.0$ & $96.0$ & $95.4 \pm 1.6$ & $96.8 \pm 1.5$ & $96.5 \pm 1.7$ & $91.2 \pm 2.7$ & $96.0$ & $96.5 \pm 2.0$ & $97.2 \pm 2.3$ & $96.9 \pm 1.7$ & $76.0$ \\
& bridges & $29.0$ & $28.0$ & $30.0$ & $24.4 \pm 2.4$ & $24.7 \pm 3.2$ & $23.5 \pm 1.8$ & $23.5 \pm 2.0$ & $26.0$ & $25.8 \pm 1.5$ & $25.6 \pm 3.1$ & $23.1 \pm 1.7$ & $20.0$ & $43.0$ & $46.0$ & $41.0$ & $44.1 \pm 1.4$ & $44.3 \pm 1.5$ & $43.4 \pm 1.0$ & $39.2 \pm 2.3$ & $45.0$ & $44.8 \pm 1.7$ & $43.7 \pm 1.4$ & $43.3 \pm 2.0$ & $37.0$ \\
\midrule
\multirow{6}{*}{\rotatebox{90}{\textbf{Challenging}}} & global\_efficiency & $25.0$ & $25.0$ & $23.0$ & $23.3 \pm 2.0$ & $24.8 \pm 3.3$ & $24.5 \pm 3.1$ & $20.7 \pm 2.7$ & $28.0$ & $26.5 \pm 4.2$ & $24.5 \pm 3.8$ & $25.6 \pm 2.5$ & $22.0$ & $43.0$ & $52.0$ & $47.0$ & $45.8 \pm 4.0$ & $40.1 \pm 4.0$ & $45.9 \pm 2.3$ & $39.1 \pm 1.5$ & $51.0$ & $45.1 \pm 2.3$ & $47.1 \pm 3.1$ & $45.5 \pm 3.0$ & $37.0$ \\
& maximal\_independent\_set & $71.0$ & $68.0$ & $72.0$ & $66.4 \pm 3.6$ & $71.0 \pm 4.8$ & $73.8 \pm 1.9$ & $52.0 \pm 4.6$ & $79.0$ & $80.5 \pm 3.2$ & $77.6 \pm 5.0$ & $81.3 \pm 3.2$ & $52.0$ & $96.0$ & $100.0$ & $98.0$ & $98.9 \pm 0.9$ & $99.2 \pm 0.6$ & $97.6 \pm 1.9$ & $91.1 \pm 2.8$ & $99.0$ & $99.1 \pm 0.9$ & $99.2 \pm 0.8$ & $99.0 \pm 0.8$ & $79.0$ \\
& maximum\_flow & $26.0$ & $36.0$ & $25.0$ & $27.7 \pm 4.2$ & $27.8 \pm 3.3$ & $16.2 \pm 0.9$ & $15.8 \pm 2.8$ & $29.0$ & $25.4 \pm 4.6$ & $26.4 \pm 3.6$ & $24.9 \pm 3.1$ & $23.0$ & $81.0$ & $81.0$ & $81.0$ & $80.8 \pm 2.9$ & $80.4 \pm 2.1$ & $44.6 \pm 3.1$ & $46.1 \pm 4.9$ & $67.0$ & $73.1 \pm 2.6$ & $70.2 \pm 2.0$ & $68.6 \pm 2.8$ & $76.0$ \\
& wiener\_index & $44.0$ & $40.0$ & $47.0$ & $42.6 \pm 2.2$ & $43.7 \pm 2.7$ & $43.9 \pm 2.2$ & $40.5 \pm 2.2$ & $41.0$ & $44.4 \pm 2.9$ & $42.4 \pm 4.1$ & $44.2 \pm 2.3$ & $38.0$ & $62.0$ & $63.0$ & $61.0$ & $60.0 \pm 2.2$ & $61.0 \pm 2.7$ & $62.1 \pm 2.5$ & $59.2 \pm 2.0$ & $60.0$ & $60.3 \pm 2.6$ & $61.8 \pm 2.9$ & $58.9 \pm 2.6$ & $57.0$ \\
& hamiltonian\_path & $4.0$ & $3.0$ & $5.0$ & $2.6 \pm 1.4$ & $2.5 \pm 0.8$ & $2.3 \pm 1.1$ & $1.8 \pm 0.8$ & $3.0$ & $1.8 \pm 0.8$ & $2.5 \pm 0.7$ & $1.7 \pm 0.7$ & $3.0$ & $2.0$ & $4.0$ & $2.0$ & $2.5 \pm 1.0$ & $2.9 \pm 1.5$ & $1.7 \pm 0.5$ & $1.9 \pm 0.9$ & $3.0$ & $3.1 \pm 1.3$ & $2.4 \pm 1.0$ & $1.8 \pm 1.4$ & $4.0$ \\
& min\_vertex\_cover & $39.0$ & $32.0$ & $40.0$ & $34.9 \pm 3.2$ & $34.5 \pm 3.7$ & $32.7 \pm 2.9$ & $29.2 \pm 1.8$ & $37.0$ & $34.8 \pm 4.0$ & $36.2 \pm 3.3$ & $32.8 \pm 4.6$ & $31.0$ & $90.0$ & $83.0$ & $91.0$ & $83.3 \pm 2.1$ & $85.7 \pm 3.8$ & $78.6 \pm 4.3$ & $77.3 \pm 4.3$ & $84.0$ & $87.5 \pm 1.4$ & $85.3 \pm 2.3$ & $87.1 \pm 2.7$ & $71.0$ \\
\bottomrule
\end{tabular}
}

    }
    \label{tab:gpt_oss_edge_reordering}
\end{sidewaystable}

\FloatBarrier 

\subsection{Syntax}\label{app:surf_encoding_additional_results}

We hypothesize that the general performance drop incurred by \opt{PyG} may be due to its characteristic order in which edge information is presented. Although fundamentally an edge-list, the \opt{PyG} encoding stores edges as an \opt{edge\_index} tensor of shape $[2, m]$, where $m$ is the number of edges and the two rows of such tensor correspond to, resp., source and target nodes. Importantly, as evidenced in the example in~\cref{sec:surf_examples}, the two rows are presented sequentially one after the other. Thus, in order to look up for neighbors of a certain node with such a surface encoding, the reasoner arguably has to skip, each time, over the entire first row, and look for corresponding targets in the second. This is a non-trivial search which requires the model to identify the index on the first row and find the corresponding one in the second, thus involving operations such as counting and offsetting. We hypothesize this complexity accounts for the worse performance reported w.r.t.\ the other surface encodings, whereby neighboring information is much more contiguous in the resulting serialization.

\begin{figure}[htbp]
    \centering   
    \includegraphics[width=\textwidth]{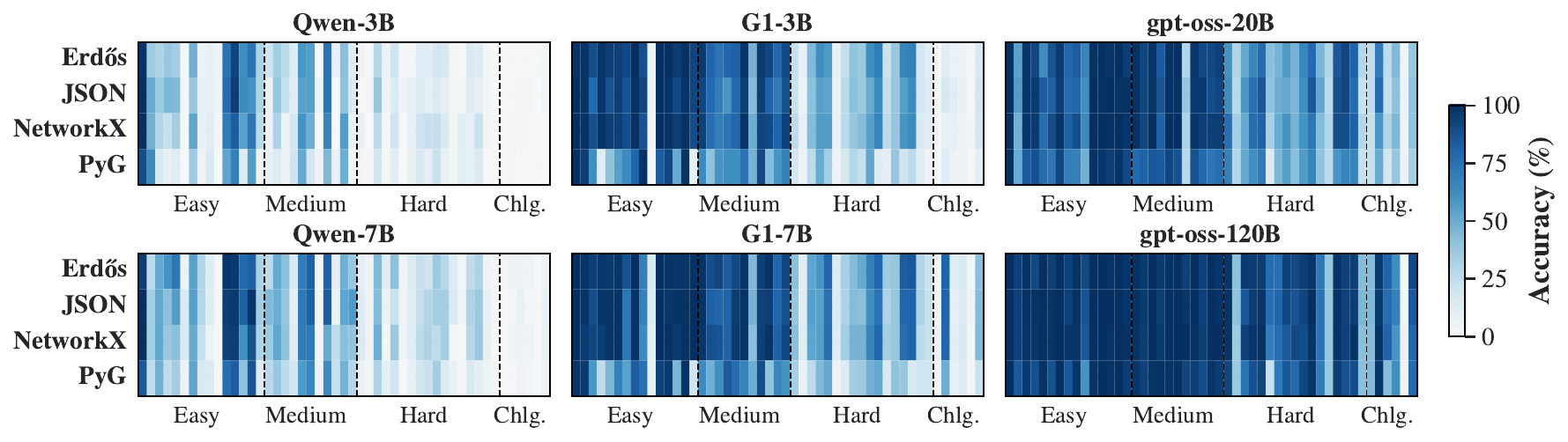}
    \captionsetup{margin=1cm}
    \caption{Accuracy of \Qwen, \GOne, and \GPT across surface encodings by task.}
\label{fig:encode_heatmap}
\end{figure}

\begin{figure}[htbp]
    \centering        \includegraphics[width=\textwidth]{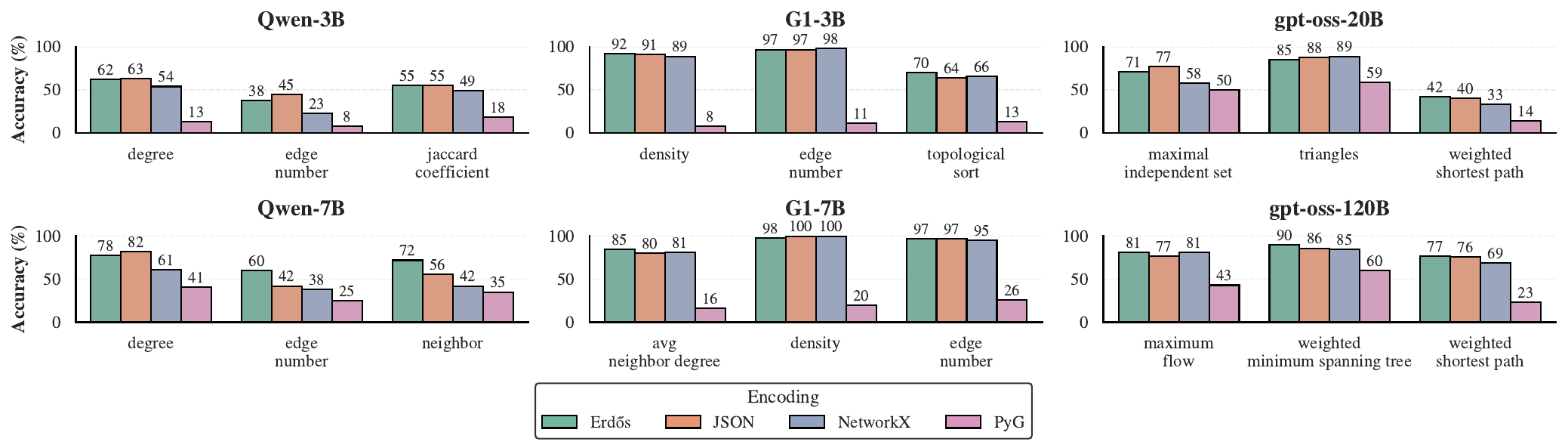}
        \caption{Top three tasks that have the highest difference between the best and worst performing surface encoding generated by the \GOne, \Qwen and \GPT models.}
\label{fig:top3_encode_examples}
\end{figure}

\Cref{fig:encode_heatmap} shows the accuracy of each encoding across the different tasks per model. 
\Cref{tab:encoding_results_all} reports the complete performance results for all encoding formats - \Erdos, \opt{JSON}, \opt{NetworkX}, and \opt{PyG} - evaluated with temperature set to 0 across all tasks for the \GOne (3B, 7B), \Qwen (3B, 7B) and \GPT (20B, 120B) models. 
Furthermore, \cref{fig:top3_encode_examples} shows the top three tasks for each model that exhibit the largest performance gap between the best and worst performing surface encodings.

\begin{sidewaystable}
    \small
    \centering
    \captionsetup{margin=1cm}
    \caption{Results generated by \Qwen, \GOne and \GPT models with the different surface encoding. \Erdos baseline encoding is shaded.}
    \resizebox{\linewidth}{!}{
        \renewcommand{\arraystretch}{1.2}
\begin{tabular}{ll >{\columncolor{gray!20}}rrrr >{\columncolor{gray!20}}rrrr >{\columncolor{gray!20}}rrrr >{\columncolor{gray!20}}rrrr >{\columncolor{gray!20}}rrrr >{\columncolor{gray!20}}rrrr}

& &  \multicolumn{4}{c}{\large\textbf{Qwen-3B}} & \multicolumn{4}{c}{\large\textbf{Qwen-7B}} & \multicolumn{4}{c}{\large\textbf{G1-3B}} & \multicolumn{4}{c}{\large\textbf{G1-7B}} & \multicolumn{4}{c}{\large\textbf{gpt-oss-20B}} & \multicolumn{4}{c}{\large\textbf{gpt-oss-120B}} \\[2ex]
\toprule
& Surface Encoding & \texttt{Erdős} & \texttt{JSON} & \texttt{NetworkX} & \texttt{PyG} & \texttt{Erdős} & \texttt{JSON} & \texttt{NetworkX} & \texttt{PyG} & \texttt{Erdős} & \texttt{JSON} & \texttt{NetworkX} & \texttt{PyG} & \texttt{Erdős} & \texttt{JSON} & \texttt{NetworkX} & \texttt{PyG} & \texttt{Erdős} & \texttt{JSON} & \texttt{NetworkX} & \texttt{PyG} & \texttt{Erdős} & \texttt{JSON} & \texttt{NetworkX} & \texttt{PyG} \\
\midrule
\multirow{15}{*}{\rotatebox{90}{Easy}} & node\_number & $98$ & $100$ & $100$ & $89$ & $95$ & $100$ & $100$ & $85$ & $98$ & $100$ & $100$ & $100$ & $100$ & $100$ & $100$ & $97$ & $100$ & $100$ & $100$ & $100$ & $100$ & $100$ & $100$ & $100$ \\
& dominating\_set & $36$ & $50$ & $48$ & $64$ & $27$ & $34$ & $35$ & $25$ & $98$ & $99$ & $99$ & $95$ & $98$ & $98$ & $96$ & $96$ & $54$ & $56$ & $48$ & $50$ & $93$ & $93$ & $93$ & $85$ \\
& common\_neighbor & $31$ & $37$ & $30$ & $16$ & $51$ & $51$ & $52$ & $48$ & $89$ & $78$ & $88$ & $61$ & $94$ & $89$ & $92$ & $58$ & $97$ & $97$ & $97$ & $84$ & $99$ & $99$ & $98$ & $96$ \\
& edge\_number & $38$ & $45$ & $23$ & $8$ & $60$ & $42$ & $38$ & $25$ & $97$ & $97$ & $98$ & $11$ & $97$ & $97$ & $95$ & $26$ & $91$ & $96$ & $96$ & $87$ & $90$ & $99$ & $97$ & $88$ \\
& neighbor & $35$ & $44$ & $34$ & $12$ & $72$ & $56$ & $42$ & $35$ & $94$ & $87$ & $94$ & $40$ & $96$ & $97$ & $97$ & $46$ & $62$ & $83$ & $76$ & $77$ & $100$ & $100$ & $100$ & $98$ \\
& bfs & $1$ & $2$ & $1$ & $0$ & $6$ & $15$ & $15$ & $6$ & $93$ & $95$ & $94$ & $55$ & $97$ & $98$ & $98$ & $68$ & $86$ & $82$ & $89$ & $72$ & $94$ & $98$ & $98$ & $87$ \\
& has\_cycle & $48$ & $39$ & $53$ & $36$ & $54$ & $55$ & $50$ & $53$ & $90$ & $86$ & $89$ & $64$ & $87$ & $72$ & $72$ & $54$ & $96$ & $95$ & $98$ & $91$ & $100$ & $100$ & $100$ & $100$ \\
& dfs & $6$ & $9$ & $8$ & $4$ & $30$ & $29$ & $23$ & $16$ & $100$ & $99$ & $99$ & $80$ & $100$ & $98$ & $98$ & $83$ & $79$ & $77$ & $82$ & $69$ & $93$ & $93$ & $97$ & $85$ \\
& minimum\_spanning\_tree & $8$ & $8$ & $11$ & $15$ & $12$ & $12$ & $6$ & $15$ & $88$ & $92$ & $86$ & $98$ & $68$ & $68$ & $64$ & $75$ & $83$ & $79$ & $88$ & $68$ & $99$ & $99$ & $97$ & $97$ \\
& weighted\_minimum\_spanning\_tree & $4$ & $4$ & $2$ & $0$ & $1$ & $1$ & $1$ & $1$ & $5$ & $5$ & $4$ & $1$ & $16$ & $13$ & $10$ & $7$ & $66$ & $62$ & $64$ & $48$ & $90$ & $86$ & $85$ & $60$ \\
& edge\_existence & $74$ & $77$ & $71$ & $53$ & $97$ & $95$ & $93$ & $76$ & $100$ & $100$ & $100$ & $83$ & $100$ & $100$ & $100$ & $93$ & $99$ & $100$ & $100$ & $99$ & $100$ & $100$ & $100$ & $99$ \\
& is\_regular & $92$ & $95$ & $83$ & $68$ & $96$ & $96$ & $95$ & $83$ & $100$ & $98$ & $98$ & $91$ & $98$ & $97$ & $99$ & $96$ & $100$ & $98$ & $100$ & $97$ & $100$ & $100$ & $100$ & $100$ \\
& degree & $62$ & $63$ & $54$ & $13$ & $78$ & $82$ & $61$ & $41$ & $95$ & $90$ & $94$ & $50$ & $97$ & $98$ & $96$ & $46$ & $97$ & $100$ & $99$ & $95$ & $98$ & $100$ & $98$ & $98$ \\
& is\_tournament & $73$ & $60$ & $72$ & $56$ & $82$ & $97$ & $87$ & $86$ & $99$ & $98$ & $99$ & $98$ & $97$ & $99$ & $99$ & $98$ & $100$ & $98$ & $100$ & $98$ & $100$ & $100$ & $100$ & $100$ \\
& density & $32$ & $31$ & $27$ & $4$ & $32$ & $35$ & $31$ & $12$ & $92$ & $91$ & $89$ & $8$ & $98$ & $100$ & $100$ & $20$ & $95$ & $99$ & $97$ & $90$ & $97$ & $97$ & $99$ & $95$ \\
\midrule
\multirow{11}{*}{\rotatebox{90}{Medium}} & adamic\_adar\_index & $21$ & $7$ & $6$ & $8$ & $27$ & $44$ & $36$ & $28$ & $96$ & $91$ & $93$ & $64$ & $98$ & $98$ & $97$ & $64$ & $99$ & $99$ & $99$ & $78$ & $99$ & $99$ & $99$ & $97$ \\
& clustering\_coefficient & $35$ & $39$ & $31$ & $14$ & $50$ & $49$ & $52$ & $38$ & $80$ & $78$ & $76$ & $41$ & $88$ & $82$ & $88$ & $48$ & $91$ & $92$ & $91$ & $81$ & $96$ & $96$ & $95$ & $93$ \\
& connected\_component\_number & $27$ & $23$ & $5$ & $6$ & $42$ & $39$ & $40$ & $26$ & $74$ & $69$ & $69$ & $59$ & $92$ & $79$ & $86$ & $61$ & $98$ & $98$ & $100$ & $85$ & $100$ & $100$ & $100$ & $98$ \\
& bipartite\_maximum\_matching & $14$ & $12$ & $18$ & $21$ & $15$ & $10$ & $11$ & $19$ & $79$ & $59$ & $73$ & $62$ & $83$ & $81$ & $81$ & $84$ & $84$ & $86$ & $82$ & $84$ & $96$ & $95$ & $96$ & $99$ \\
& local\_connectivity & $58$ & $54$ & $60$ & $51$ & $70$ & $71$ & $70$ & $57$ & $82$ & $78$ & $79$ & $62$ & $90$ & $95$ & $96$ & $77$ & $99$ & $99$ & $100$ & $91$ & $99$ & $100$ & $100$ & $98$ \\
& jaccard\_coefficient & $55$ & $55$ & $49$ & $18$ & $82$ & $82$ & $71$ & $65$ & $97$ & $95$ & $98$ & $74$ & $98$ & $98$ & $98$ & $67$ & $100$ & $98$ & $100$ & $87$ & $100$ & $100$ & $100$ & $100$ \\
& min\_edge\_covering & $0$ & $5$ & $0$ & $4$ & $3$ & $3$ & $2$ & $5$ & $47$ & $43$ & $49$ & $47$ & $44$ & $47$ & $42$ & $49$ & $30$ & $39$ & $32$ & $28$ & $93$ & $97$ & $100$ & $92$ \\
& is\_eularian & $74$ & $73$ & $67$ & $45$ & $82$ & $83$ & $50$ & $64$ & $96$ & $95$ & $92$ & $90$ & $93$ & $93$ & $93$ & $89$ & $97$ & $96$ & $97$ & $87$ & $100$ & $100$ & $99$ & $95$ \\
& degree\_centrality & $9$ & $0$ & $1$ & $7$ & $10$ & $6$ & $25$ & $23$ & $91$ & $81$ & $90$ & $43$ & $98$ & $94$ & $98$ & $40$ & $90$ & $96$ & $95$ & $78$ & $93$ & $93$ & $95$ & $89$ \\
& is\_bipartite & $42$ & $40$ & $56$ & $49$ & $48$ & $53$ & $42$ & $47$ & $79$ & $71$ & $81$ & $59$ & $79$ & $79$ & $82$ & $65$ & $90$ & $89$ & $94$ & $72$ & $100$ & $100$ & $100$ & $97$ \\
& resource\_allocation\_index & $12$ & $15$ & $7$ & $3$ & $38$ & $57$ & $40$ & $29$ & $90$ & $88$ & $93$ & $68$ & $92$ & $88$ & $92$ & $60$ & $94$ & $92$ & $95$ & $76$ & $97$ & $98$ & $100$ & $95$ \\
\midrule
\multirow{17}{*}{\rotatebox{90}{Hard}} & max\_weight\_matching & $4$ & $0$ & $7$ & $3$ & $10$ & $13$ & $15$ & $7$ & $23$ & $22$ & $26$ & $18$ & $42$ & $39$ & $42$ & $33$ & $60$ & $63$ & $75$ & $68$ & $97$ & $98$ & $98$ & $91$ \\
& closeness\_centrality & $1$ & $3$ & $0$ & $1$ & $4$ & $5$ & $5$ & $3$ & $8$ & $6$ & $8$ & $6$ & $11$ & $11$ & $12$ & $7$ & $30$ & $28$ & $35$ & $29$ & $41$ & $40$ & $36$ & $37$ \\
& traveling\_salesman\_problem & $29$ & $39$ & $26$ & $19$ & $44$ & $43$ & $49$ & $31$ & $44$ & $41$ & $43$ & $27$ & $53$ & $55$ & $53$ & $26$ & $51$ & $58$ & $51$ & $51$ & $96$ & $85$ & $93$ & $73$ \\
& strongly\_connected\_number & $6$ & $4$ & $1$ & $2$ & $11$ & $7$ & $2$ & $9$ & $62$ & $54$ & $56$ & $50$ & $59$ & $54$ & $55$ & $44$ & $65$ & $75$ & $77$ & $61$ & $95$ & $95$ & $97$ & $87$ \\
& shortest\_path & $21$ & $15$ & $20$ & $11$ & $41$ & $22$ & $37$ & $20$ & $57$ & $53$ & $56$ & $35$ & $79$ & $75$ & $76$ & $51$ & $77$ & $85$ & $81$ & $71$ & $95$ & $97$ & $99$ & $92$ \\
& weighted\_shortest\_path & $2$ & $4$ & $1$ & $0$ & $4$ & $4$ & $4$ & $1$ & $8$ & $7$ & $7$ & $5$ & $15$ & $15$ & $13$ & $6$ & $42$ & $40$ & $33$ & $14$ & $77$ & $76$ & $69$ & $23$ \\
& center & $1$ & $8$ & $8$ & $7$ & $13$ & $11$ & $11$ & $7$ & $24$ & $27$ & $27$ & $17$ & $33$ & $31$ & $34$ & $14$ & $47$ & $49$ & $50$ & $45$ & $76$ & $79$ & $76$ & $71$ \\
& diameter & $12$ & $14$ & $21$ & $15$ & $23$ & $23$ & $26$ & $19$ & $40$ & $42$ & $39$ & $29$ & $43$ & $43$ & $47$ & $35$ & $57$ & $51$ & $60$ & $55$ & $93$ & $91$ & $84$ & $85$ \\
& barycenter & $11$ & $9$ & $23$ & $10$ & $21$ & $29$ & $31$ & $22$ & $37$ & $37$ & $43$ & $24$ & $51$ & $42$ & $47$ & $37$ & $47$ & $46$ & $47$ & $42$ & $90$ & $86$ & $90$ & $78$ \\
& radius & $18$ & $15$ & $24$ & $22$ & $37$ & $34$ & $26$ & $35$ & $62$ & $50$ & $55$ & $53$ & $68$ & $61$ & $66$ & $48$ & $65$ & $61$ & $60$ & $56$ & $94$ & $92$ & $89$ & $91$ \\
& topological\_sort & $17$ & $10$ & $15$ & $7$ & $28$ & $34$ & $34$ & $15$ & $70$ & $64$ & $66$ & $13$ & $79$ & $73$ & $79$ & $51$ & $78$ & $85$ & $72$ & $66$ & $97$ & $100$ & $98$ & $97$ \\
& periphery & $2$ & $2$ & $3$ & $3$ & $13$ & $15$ & $3$ & $7$ & $22$ & $24$ & $26$ & $10$ & $36$ & $31$ & $34$ & $17$ & $46$ & $52$ & $43$ & $40$ & $72$ & $74$ & $74$ & $70$ \\
& betweenness\_centrality & $1$ & $3$ & $11$ & $10$ & $3$ & $4$ & $0$ & $2$ & $38$ & $39$ & $39$ & $39$ & $39$ & $39$ & $38$ & $40$ & $30$ & $26$ & $30$ & $26$ & $39$ & $39$ & $39$ & $38$ \\
& triangles & $16$ & $13$ & $8$ & $8$ & $25$ & $31$ & $28$ & $16$ & $67$ & $55$ & $59$ & $21$ & $83$ & $79$ & $76$ & $39$ & $85$ & $88$ & $89$ & $59$ & $98$ & $99$ & $98$ & $95$ \\
& avg\_neighbor\_degree & $14$ & $9$ & $22$ & $15$ & $32$ & $37$ & $38$ & $10$ & $68$ & $64$ & $63$ & $26$ & $85$ & $80$ & $81$ & $16$ & $91$ & $92$ & $89$ & $66$ & $93$ & $93$ & $93$ & $87$ \\
& harmonic\_centrality & $3$ & $3$ & $3$ & $1$ & $4$ & $8$ & $8$ & $5$ & $18$ & $18$ & $15$ & $11$ & $30$ & $32$ & $26$ & $20$ & $81$ & $78$ & $82$ & $67$ & $95$ & $95$ & $99$ & $89$ \\
& bridges & $0$ & $0$ & $0$ & $0$ & $3$ & $2$ & $0$ & $3$ & $11$ & $12$ & $12$ & $4$ & $22$ & $23$ & $23$ & $20$ & $29$ & $23$ & $29$ & $23$ & $43$ & $45$ & $42$ & $42$ \\
\midrule
\multirow{6}{*}{\rotatebox{90}{Challenging}} & global\_efficiency & $0$ & $1$ & $0$ & $0$ & $2$ & $1$ & $1$ & $0$ & $1$ & $1$ & $1$ & $0$ & $1$ & $2$ & $4$ & $2$ & $25$ & $25$ & $24$ & $24$ & $43$ & $45$ & $43$ & $44$ \\
& maximal\_independent\_set & $1$ & $1$ & $2$ & $2$ & $5$ & $2$ & $4$ & $1$ & $12$ & $17$ & $8$ & $12$ & $81$ & $80$ & $79$ & $49$ & $71$ & $77$ & $58$ & $50$ & $96$ & $96$ & $99$ & $97$ \\
& maximum\_flow & $1$ & $2$ & $2$ & $0$ & $5$ & $8$ & $5$ & $4$ & $7$ & $11$ & $9$ & $5$ & $12$ & $9$ & $11$ & $8$ & $26$ & $24$ & $30$ & $18$ & $81$ & $77$ & $81$ & $43$ \\
& wiener\_index & $1$ & $2$ & $1$ & $1$ & $6$ & $3$ & $5$ & $3$ & $6$ & $7$ & $6$ & $5$ & $15$ & $14$ & $15$ & $13$ & $44$ & $41$ & $45$ & $39$ & $62$ & $68$ & $59$ & $62$ \\
& hamiltonian\_path & $1$ & $0$ & $0$ & $1$ & $1$ & $2$ & $1$ & $1$ & $2$ & $2$ & $3$ & $1$ & $3$ & $5$ & $3$ & $3$ & $4$ & $5$ & $4$ & $5$ & $2$ & $3$ & $6$ & $4$ \\
& min\_vertex\_cover & $3$ & $4$ & $0$ & $0$ & $6$ & $8$ & $7$ & $6$ & $16$ & $18$ & $18$ & $10$ & $42$ & $43$ & $44$ & $22$ & $39$ & $40$ & $40$ & $33$ & $90$ & $85$ & $79$ & $78$ \\
\midrule
\bottomrule
\end{tabular}

    }
\label{tab:encoding_results_all}
\end{sidewaystable}

\FloatBarrier

\section{Spectral Tasks}\label{app:spectral}

\subsection{Spectral Tasks and Categorization}
We classify 12 spectral tasks as Easy (3), Medium (6), or Hard (3) based on the spectral information required (Table~\ref{tab:spectral_task_classification_easy_medium_hard}).
Easy tasks depend on simple aggregates or the dominant eigenvalue (e.g., graph energy as the sum of absolute eigenvalues). Medium tasks require interior eigenvalues that are harder to approximate, such as algebraic connectivity (the second-smallest Laplacian eigenvalue). Hard tasks rely on the entire spectrum with nonlinear transforms, for instance von Neumann entropy, which applies logarithmic weighting over all eigenvalues and is computationally demanding.

\subsection{Evaluation Metrics for Generalization on Spectral Tasks}
We evaluate spectral prediction tasks using three error metrics: normalized root mean squared error (nRMSE), symmetric mean absolute percentage error (sMAPE), and relative mean absolute error (RelMAE). 
The nRMSE is scale-free but sensitive to outliers because it relies on squared deviations. The sMAPE is bounded, which facilitates comparisons across tasks and models, but it inflates when both target and prediction are near zero.
The RelMAE rescales errors relative to a mean-predictor baseline and avoids the instability of sMAPE at near-zero values.
The mathematical definitions are outlined in the next section for reproducibility and a high level overview is in Table \ref{tab:eval_formulas}. The detailed results of the spectral tasks is shown in Table \ref{tab:spectral_results_on_all_models_all_metrics}. 
\begin{table}[h!]
\centering
\caption{Evaluation Metrics Used For Testing Generalisation}
\begin{tabular}{p{3cm} p{7.5cm} p{2cm}}
\toprule
\textbf{Metric} & \textbf{Formula} & \textbf{Reference} \\
\midrule \\
\textbf{nRMSE (range)} & 
$\mathrm{nRMSE}_{\text{range}} = 
 \dfrac{\mathrm{RMSE}}{\max_i y_i - \min_i y_i}$ 
 & \cite{botchkarev2018performance} \\[1.8em]

\textbf{nRMSE (std)} & 
$\mathrm{nRMSE}_{s} = 
 \dfrac{\mathrm{RMSE}}{s_y}$ 
 & \cite{botchkarev2018performance} \\[1.8em]

\textbf{sMAPE} & 
$\mathrm{sMAPE}_{0\text{--}100} = 
 \tfrac{100}{n}\sum_{i=1}^n \dfrac{|y_i - \hat y_i|}{(|y_i|+|\hat y_i|)/2 + \varepsilon}$ 
 & \cite{makridakis2000m3, makridakis1993accuracy} \\[1.8em]

\textbf{RelMAE} & 
$\mathrm{RelMAE} = \dfrac{\mathrm{MAE}}{\mathrm{MAE}_{\text{mean}}}$ 
 & \cite{hyndman2006another} \\[1.8em]
\bottomrule
\label{tab:eval_formulas}
\end{tabular}
\end{table}

\subsubsection{Normalized Root Mean Squared Error} 
In our study, for observations (ground truth answers) $\{y_i\}_{i=1}^n$ and predictions (generated outputs from the LLMs) $\{\hat{y}_i\}_{i=1}^n$, 
the root mean squared error is defined as
\[
\mathrm{RMSE} = \left( \frac{1}{n}\sum_{i=1}^n (y_i - \hat{y}_i)^2 \right)^{1/2}.
\]
We consider two different types of normalizations.
First, division by the empirical range of the target values,
\[
\mathrm{nRMSE}_{\mathrm{range}} = \frac{\mathrm{RMSE}}{\max_i y_i - \min_i y_i},
\]
and second, division by the sample standard deviation,
\[
s_y = \left( \frac{1}{n-1}\sum_{i=1}^n (y_i - \bar{y})^2 \right)^{1/2}, 
\qquad 
\mathrm{nRMSE}_{std} = \frac{\mathrm{RMSE}}{s_y}.
\]
This metric is between $[0,\infty)$, where $0$ would imply exact prediction. 
Since these metrics are derived from squared deviations, both  $\mathrm{nRMSE}_{\mathrm{range}}$ and $\mathrm{nRMSE}_{std}$ are sensitive to large errors. 

\textbf{Example: Interpretation of nRMSE by range} \\
For example, $\mathrm{nRMSE}_{\mathrm{range}}=0.05$ indicates an average error of 5\% of the observed spread, which is desirable when the spread reflects meaningful variability (e.g., eigenvalues in $[0,1000]$).
However, if the data are tightly concentrated, say within $[0,1]$, 
the same value corresponds to an error of 0.05 on a scale where natural fluctuations are of order 0.01, indicating poor performance.

\textbf{Example: Interpretation of nRMSE by standard deviation} \\
Under standard-deviation normalization, error is expressed relative to target variability. For instance, if eigenvalues are tightly clustered ($s_y=0.01$) and RMSE $=0.01$, then $\mathrm{nRMSE}_{std}=1$, indicating poor performance. If eigenvalues are dispersed ($s_y=10$) with the same RMSE, then $\mathrm{nRMSE}_{std}=0.001$, indicating negligible error.
Thus, $\mathrm{nRMSE}_{std}$ adapts to variability and enables comparison across tasks with different dispersions.

\subsubsection{Symmetric Mean Absolute Percentage Error (sMAPE)} 
The symmetric mean absolute percentage error on the $[0,200]\%$ scale is
\[
\mathrm{sMAPE}_{0\text{--}200}
= \frac{100}{n}\sum_{i=1}^n 
\frac{2\,|y_i-\hat y_i|}{|y_i| + |\hat y_i| + \varepsilon},
\qquad \varepsilon = 10^{-12}.
\]
A rescaled form on the $[0,100]\%$ scale is given by
\[
\mathrm{sMAPE}_{0\text{--}100} 
= \tfrac{1}{2}\,\mathrm{sMAPE}_{0\text{--}200}.
\]
Both forms are bounded and symmetric in $y_i$ and $\hat y_i$. The use of 
$|y_i|+|\hat y_i|$ in the denominator eliminates the singularity (undefined case such as division by 0) present in MAPE 
when $y_i=0$, although the metric remains sensitive when both $y_i$ and $\hat y_i$ are close to zero.

\textbf{Example: Interpretation of sMAPE} \\
To explain, we quote an example from the work by Makridakis \cite{makridakis1993accuracy}, consider $y=150$ and $\hat y=100$, so that the absolute error is $|150-100|=50$. 
The denominator is the average of the magnitudes, $(150+100)/2=125$, giving
\[
\mathrm{sMAPE}_{0\text{--}100} = \frac{50}{125} \times 100 = 40\%.
\]
If instead $y=100$ and $\hat y=150$, the same calculation returns $40\%$, 
an is symmetric across over-prediction and under-prediction.

\textbf{Example: Near Zero Behaviour of sMAPE} \\
To illustrate behaviour near zero, consider $y=0.1$ and $\hat y=0.2$, 
which gives an absolute error of $0.1$ and denominator $(0.1+0.2)/2=0.15$. 
This results in
\[
\mathrm{sMAPE}_{0\text{--}100} = \frac{0.1}{0.15}\times 100 \approx 66.7\%.
\]
Although the absolute error is small in absolute terms, the percentage error is large 
because both $y$ and $\hat y$ are close to zero. 

\subsubsection{Relative Mean Absolute Error (RelMAE)}  

RelMAE is a scale-free measure of predictive accuracy, defined as the MAE of a model relative to the MAE of a baseline predictor, taken here as the sample mean
\cite{hyndman2006another}.  

Let $\{y_i\}_{i=1}^n$ denote the observed values and $\{\hat{y}_i\}_{i=1}^n$ the corresponding predictions.  
The mean absolute error of the model is  
\[
\mathrm{MAE} = \frac{1}{n}\sum_{i=1}^n \lvert y_i - \hat{y}_i \rvert,
\]  
and the mean absolute error of the mean predictor is  
\[
\mathrm{MAE}_{\text{mean}} = \frac{1}{n}\sum_{i=1}^n \lvert y_i - \bar{y} \rvert,
\quad \bar{y} = \frac{1}{n}\sum_{i=1}^n y_i.
\]  
The RelMAE is defined as  
\[
\mathrm{RelMAE} = \frac{\mathrm{MAE}}{\mathrm{MAE}_{\text{mean}}}.
\]  

By construction, $\text{RelMAE} < 1$ indicates better performance than the mean predictor, $\mathrm{RelMAE}=1$  indicates equal performance, and $\mathrm{RelMAE}>1$ indicates worse performance.
A limitation of RelMAE is that its interpretation depends on the choice of baseline. 

\textbf{Example: Interpretation of RelMAE} \\
For example, if a model yields $\mathrm{MAE}=2$ while the mean predictor yields $\mathrm{MAE}_{\text{mean}}=5$, then $\mathrm{RelMAE}=0.4$, showing that the model achieves better-than-baseline accuracy. Conversely, if $\mathrm{MAE}=8$ while $\mathrm{MAE}_{\text{mean}}=5$, then $\mathrm{RelMAE}=1.6$, showing that the model performs worse than the mean predictor.

\textbf{Example: Near Zero Behavior of RelMAE}  \\
To illustrate the behavior of RelMAE near zero, consider again $y=0.1$ and $\hat y=0.2$, 
which gives an absolute error of $0.1$.  
Suppose that, over the dataset, the mean predictor (which always outputs 
$\bar y$) has a mean absolute error of
\[
\mathrm{MAE}_{\text{mean}} = \frac{1}{n}\sum_{i=1}^n |y_i - \bar{y}| \approx 2.
\]
Then the relative mean absolute error for this example is
\[
\mathrm{RelMAE} = \frac{0.1}{2} = 0.05.
\]
Unlike sMAPE, the RelMAE remains small here, reflecting that the error $0.1$ is negligible 
relative to the overall variability of the data.

\subsection{Experiments on Benchmarking Spectral Graph Reasoning}\label{app:spectral_results}
We construct a novel suite of twelve spectral tasks and evaluate the capability of fine-tuned (\GOne) and non-fine-tuned (\Qwen) LLM graph reasoners on 100 graphs from the \Erdos test dataset. Generalization is assessed across multiple axes: task difficulty (Easy, Medium, Hard), spectral task type, and reasoning style of generated answers. We illustrated the generalization potential of \GOne-7B and \Qwen-7B model in Section \ref{sec:spectral}. Here, we report the results on the spectral tasks in \ref{appendix_quantitative} and the reasoning style of models in \ref{appendix_qualitative}.

\subsubsection{Spectral Task-wise and Model-wise Analysis}
\label{appendix_quantitative}

\begin{table*}[t]
\centering
\caption{Best-performing model per task and standard deviation of errors across models. Larger standard deviation values indicate greater disagreement in model performance.
}
\label{tab:best_and_agreement}
\resizebox{\textwidth}{!}{%
\begin{tabular}{lcccccccc}
\toprule
\multirow{2}{*}{\textbf{Task}} & 
\multicolumn{4}{c}{\textbf{Best Model}} & 
\multicolumn{4}{c}{\textbf{Std. Deviation of Errors}} \\
\cmidrule(lr){2-5} \cmidrule(lr){6-9}
 & nRMSE$_{\text{range}}$ & nRMSE$_{\text{std}}$ & sMAPE & RelMAE 
 & nRMSE$_{\text{range}}$ & nRMSE$_{\text{std}}$ & sMAPE & RelMAE \\
\midrule
algebraic connectivity   & Qwen-7B & Qwen-7B & Qwen-7B & Qwen-7B & 0.333 & 1.040 & 7.427 & 1.068 \\ 
eigenvector cent top     & G1-7B   & G1-7B   & G1-7B   & G1-7B   & 0.055 & 0.225 & 13.811 & 0.265 \\
estrada index  & G1-7B   & G1-7B   & G1-7B   & G1-7B   & \textbf{938.220} & \textbf{3003.889} & \textbf{3.093} & \textbf{590.392} \\
graph energy  & Qwen-7B & Qwen-7B & Qwen-7B & Qwen-7B & \textbf{2.065} & \textbf{8.885} & \textbf{15.645} & \textbf{4.303} \\
heat trace t1   & G1-7B   & G1-7B   & G1-7B   & G1-7B   & \textbf{11.895} & \textbf{40.478} & \textbf{5.511} & \textbf{5.098} \\
laplacian energy& \textbf{G1-7B}   & \textbf{Qwen-7B }& \textbf{G1-7B}   & \textbf{G1-3B}   & 0.027 & 0.091 & 4.077 & 0.074 \\
n components             & G1-7B   & G1-7B   & G1-7B   & G1-7B   & 0.377 & 2.002 & 10.680 & 1.285 \\
natural connectivity & \textbf{Qwen-3B }& \textbf{Qwen-3B} & \textbf{Qwen-7B} & \textbf{Qwen-7B} & 1.192 & 6.245 & 14.646 & 2.153 \\
spectral gap    & \textbf{G1-7B}   & \textbf{Qwen-7B }& \textbf{G1-3B}   & \textbf{Qwen-7B} & 0.382 & 1.223 & 5.808 & 0.180 \\
spectral radius & \textbf{Qwen-7B} & \textbf{Qwen-7B} & \textbf{G1-7B} & \textbf{G1-7B}   & 0.084 & 0.473 & 2.753 & 0.285 \\ 
sum lambda squared       & Qwen-7B & Qwen-7B & Qwen-7B & Qwen-7B & 0.576 & 4.145 & 5.135 & 1.823 \\
von neumann entropy      & Qwen-7B & Qwen-7B & Qwen-7B & Qwen-7B & \textbf{2.348} & \textbf{8.837} & \textbf{15.510} & \textbf{2.407} \\
\bottomrule
\end{tabular}
}
\label{tab:best_perf_model}
\end{table*}

\begin{figure}[h]
    \centering
    \includegraphics[width=0.6\linewidth]{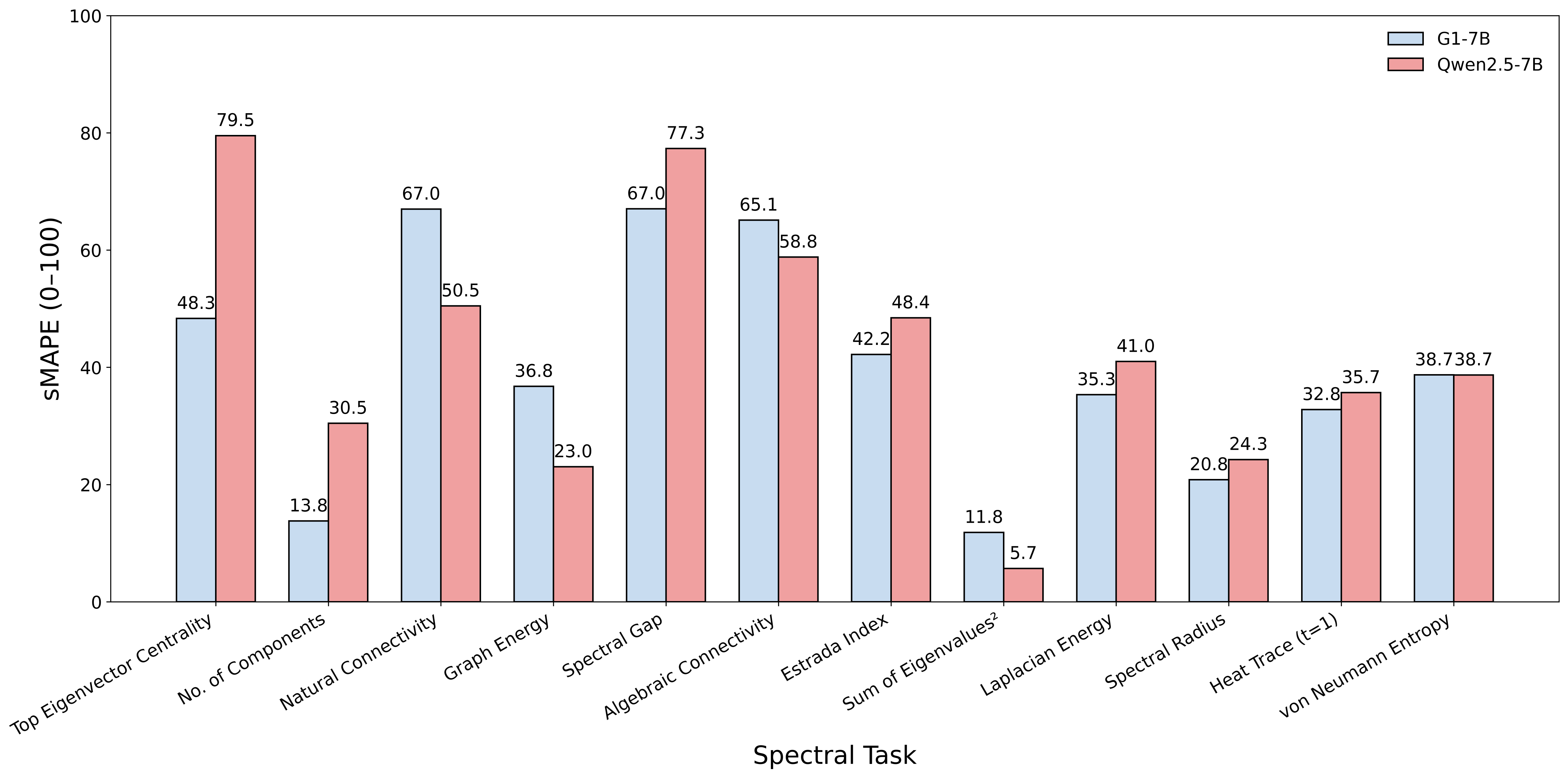}
    \caption{Performance of \texttt{Qwen-7B} and \texttt{G1-7B} mdoels across spectral tasks. Lower is better.}
    \label{fig:qwen_g1_task_wise_smape}
\end{figure}

\textbf{Consistency of Best-Performing Models:} \textit{Is the same model consistently the best across all metrics for a given spectral task?}
\Cref{tab:spectral_results_on_all_models_all_metrics} reports numerical results for all tasks, models, and metrics on the spectral tasks. 
Analysis of the best-performing model per task shows that eight tasks exhibit full consistency, with the same model ranking best across all four metrics. 
The split is even among the largest models: \texttt{Qwen-7B} leads on four tasks and \texttt{G1-7B} on four tasks. 
\Cref{fig:qwen_g1_task_wise_smape} illustrates a head-to-head comparison between \Qwen-7B and \GOne-7B.
In contrast, four tasks show variation, with different models ranking best depending on the metric and are further highlighted in Table \ref{tab:best_perf_model}. 

\begin{figure}[ht!]
    \centering
    \includegraphics[width=0.95\linewidth]{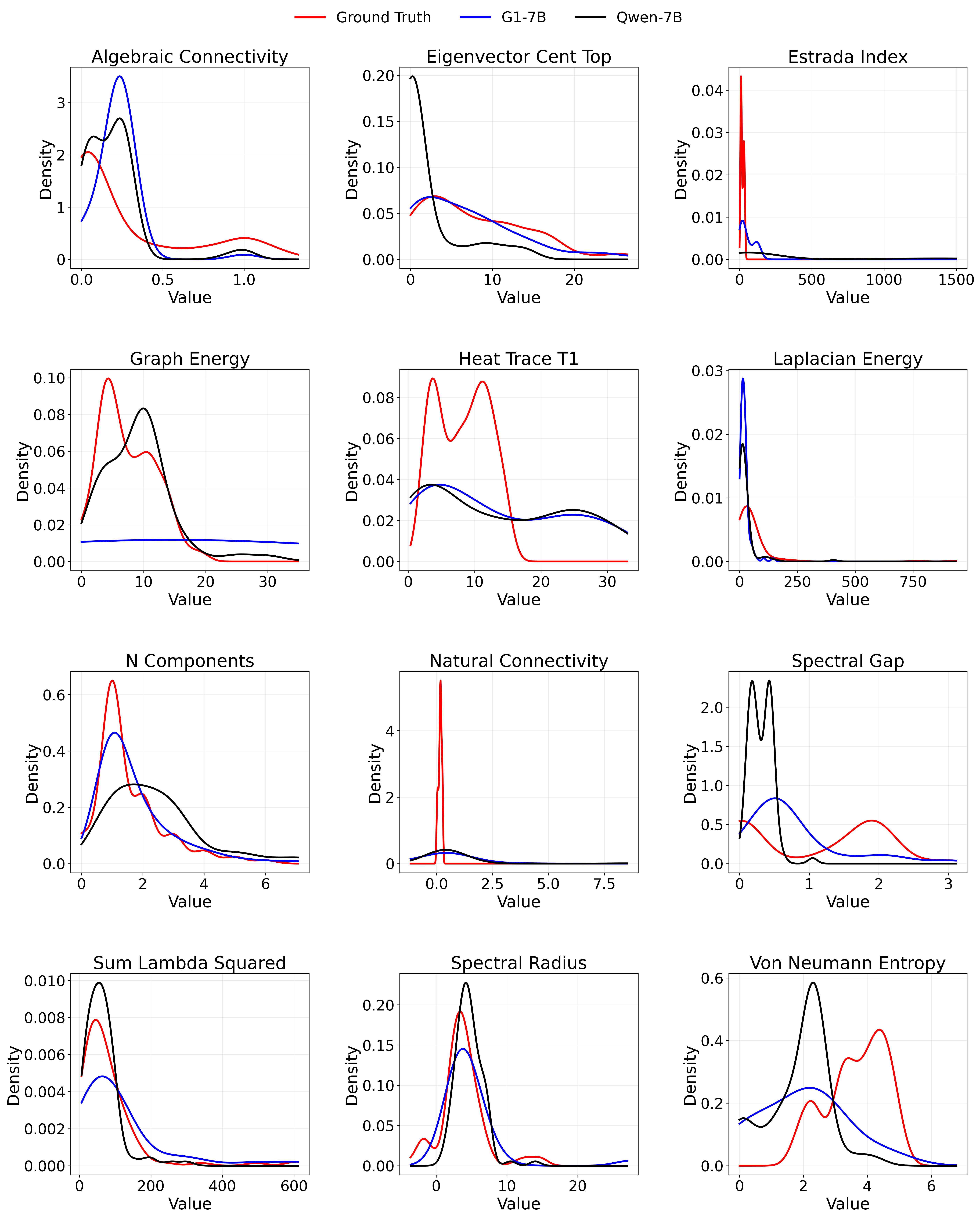}
    \caption{Distributions of ground truth (red), G1-7B (blue), and Qwen-7B (black).}
    \label{fig:distributions_ground_predicted}
\end{figure}

\textbf{Model Variability Across Tasks:}
\textit{Which spectral tasks show the greatest variability in model performance, and why?}
The Estrada Index exhibits the highest standard deviation across models, with large variance in both nRMSE and RelMAE as seen in Table \ref{tab:best_perf_model}. 
This variance reflects its exponential dependence on the full spectrum. 
In addition, we observe that LLMs often resort to assumptive heuristics for this task~\ref{appendix_qualitative}, substituting illustrative values rather than executing full computations. 
Graph Energy and Heat Trace also show high variance. 
As shown in Figure \ref{fig:distributions_ground_predicted}, the distributions of model predictions barely overlap with the ground truth, which explains the elevated error values.
In contrast, easier tasks such as \texttt{Algebraic Connectivity} and \texttt{Eigenvector Centrality} display low variance. 
For \texttt{Eigenvector Centrality}, qualitative inspection of fifteen samples revealed that nine answers were derived heuristically 
and six analytically, as categorized in \Cref{appendix_qualitative}. 
We hypothesize that tasks eliciting a higher proportion of analytically derived answers are more likely to exhibit stronger and more consistent model performance. 

\textbf{Cross-Metric Correlation:} 
\textit{Do different error metrics capture distinct failure modes in spectral reasoning?}
We compute correlations between the four error metrics averaged across tasks. 
The results show that nRMSE$_{\text{range}}$, nRMSE$_{\text{std}}$, and RelMAE are nearly perfectly correlated ($\rho > 0.99$), 
indicating that they are essentially redundant and capture the same failure mode. 
In contrast, sMAPE is almost uncorrelated with the other three metrics ($\rho \approx 0.01$), 
suggesting that it measures a different dimension of model error.
This justifies our choice to report \textbf{sMAPE and RelMAE} in the main text, 
as they provide complementary perspectives on model performance.

\textbf{Best-Performing Models Across Metrics:}  
\textit{Does model size or metric choice determine which model dominates spectral reasoning tasks?}  
We count the number of wins within each error metric across tasks.
\texttt{G1-7B} and \texttt{Qwen-7B} achieve the highest number of task-wise best performances, whereas the 3B models seldom achieve top performance.
\texttt{G1-7B} wins 6 tasks compared to 5 for \texttt{Qwen-7B} on sMAPE, while the counts reverse on RelMAE (5 and 6 respectively), which suggests that relative model ranking depends on the chosen metric.

\textbf{Global Error Analysis:} 
\textit{Can a global error measure reconcile different evaluation metrics to identify the most reliable spectral reasoning model?}
To address the issue caused by selecting one metric, we construct a global error measure by combining the two independent metrics, sMAPE and RelMAE.  
Since these metrics are on different scales, we first apply min--max normalization within each metric 
and then compute a global normalized error by averaging across tasks and metrics.  
This analysis identifies \textbf{G1-7B (0.214)} as the overall best-performing model, closely followed by \textbf{Qwen-7B (0.232)}. Similarly, \textbf{G1-3B (0.283)} outperforms \textbf{Qwen-3B (0.318)}.

\textbf{Mapping Distributions:} 
\textit{Does the model reproduce the range, variability, and shape of all values for this task?} 
To dive deeper beyond global error scores, we examine the distributions of predicted versus ground-truth values for each spectral task.
We construct kernel density estimation (KDE) curves by aggregating values across all instances of a task, 
restricting the plotting range to the 0.5--99.5 percentile to mitigate outliers. 
KDE curves are overlaid for Ground Truth (\textcolor{red}{red}), \texttt{G1-7B} (\textcolor{blue}{blue}), and \texttt{Qwen-7B} (\textcolor{black}{black}) and are shown in Figure \ref{fig:distributions_ground_predicted}.  

The most interesting tasks are those where the\textbf{ ground-truth values form complex distributions}. However, scalar error metrics such as nRMSE, RelMAE, and sMAPE reduce model performance to a single scalar value. 
These measures can mask important qualitative differences, for example, two models may obtain similar error scores but fail in different ways. Specifically, scalar metrics do not reveal whether a model fails to capture rare but extreme values (e.g., heavy-tailed behaviour in \texttt{Estrada Index}), or collapses multiple distinct regimes into a single average (e.g., smoothing multimodal distributions in \texttt{Heat Trace T1}, \texttt{Spectral Gap}, and \texttt{Von Neumann Entropy}).

\begin{sidewaystable}
    \small
    \centering
    \captionsetup{margin=1cm}
    \caption{Spectral task results across four models. Lower is better for all metrics.}
    \resizebox{\linewidth}{!}{
        \begin{tabular}{lcccccccccccccccc}
\toprule
Task & \multicolumn{4}{c}{nRMSE$_{\mathrm{std}}$} & \multicolumn{4}{c}{nRMSE$_{\mathrm{range}}$} & \multicolumn{4}{c}{sMAPE$_{0\text{--}100}$} & \multicolumn{4}{c}{RelMAE} \\
\cmidrule(lr){2-5}\cmidrule(lr){6-9}\cmidrule(lr){10-13}\cmidrule(lr){14-17}
 & G1-3B & G1-7B & Qwen-3B & Qwen-7B & G1-3B & G1-7B & Qwen-3B & Qwen-7B & G1-3B & G1-7B & Qwen-3B & Qwen-7B & G1-3B & G1-7B & Qwen-3B & Qwen-7B \\
\midrule
algebraic connectivity & 3.14 & 1.11 & 2.33 & 0.96 & 0.93 & 0.32 & 0.79 & 0.27 & 76.71 & 65.10 & 67.82 & 58.81 & 3.13 & 0.99 & 1.55 & 0.75 \\
eigenvector cent top & 1.38 & 1.14 & 1.69 & 1.43 & 0.33 & 0.28 & 0.41 & 0.35 & 61.65 & 48.35 & 73.68 & 79.53 & 1.23 & 0.99 & 1.63 & 1.24 \\
estrada index & 1035.37 & 6.40 & 6296.12 & 50.10 & 307.00 & 1.89 & 1963.05 & 14.79 & 44.87 & 42.19 & 48.64 & 48.44 & 123.09 & 4.29 & 1227.62 & 26.64 \\
graph energy & 21.88 & 18.20 & 13.13 & 1.48 & 5.08 & 4.22 & 3.35 & 0.34 & 59.10 & 36.76 & 49.43 & 23.04 & 10.96 & 4.34 & 7.89 & 1.04 \\
heat trace t1 & 2.87 & 2.56 & 83.63 & 2.61 & 0.84 & 0.75 & 24.57 & 0.76 & 44.45 & 32.79 & 42.48 & 35.70 & 2.83 & 2.26 & 12.68 & 2.41 \\
laplacian energy & 1.01 & 1.06 & 1.19 & 0.99 & 0.13 & 0.11 & 0.18 & 0.13 & 36.17 & 35.34 & 43.95 & 40.99 & 0.82 & 0.91 & 0.99 & 0.96 \\
n components & 4.34 & 1.24 & 5.44 & 1.83 & 0.80 & 0.24 & 1.04 & 0.35 & 19.21 & 13.79 & 37.39 & 30.46 & 1.86 & 0.58 & 3.66 & 1.57 \\
natural connectivity & 36.53 & 33.68 & 22.10 & 30.18 & 9.50 & 8.65 & 6.72 & 7.79 & 86.05 & 66.99 & 64.25 & 50.47 & 16.29 & 14.96 & 12.60 & 11.58 \\
spectral gap & 3.74 & 1.27 & 2.82 & 1.26 & 1.16 & 0.38 & 0.86 & 0.39 & 64.09 & 67.04 & 71.95 & 77.33 & 1.46 & 1.18 & 1.40 & 1.08 \\
sum lambda squared & 1.24 & 0.97 & 9.21 & 0.61 & 0.21 & 0.16 & 1.31 & 0.10 & 12.67 & 11.84 & 18.23 & 5.68 & 0.92 & 0.60 & 4.24 & 0.36 \\
spectral radius & 1.88 & 1.09 & 1.46 & 0.78 & 0.32 & 0.18 & 0.26 & 0.13 & 26.55 & 20.84 & 26.76 & 24.27 & 1.44 & 0.86 & 1.21 & 0.87 \\
von neumann entropy & 20.33 & 2.67 & 3.14 & 2.20 & 5.43 & 0.72 & 0.90 & 0.60 & 71.32 & 38.72 & 53.66 & 38.70 & 7.48 & 2.53 & 3.26 & 2.39 \\
\bottomrule
\end{tabular}

    }
    \label{tab:spectral_results_on_all_models_all_metrics}
\end{sidewaystable}

\newpage
\subsubsection{Reasoning Styles of Graph LLMs}
\label{appendix_qualitative}

\begin{tcolorbox}[title=\textbf{G1-7B Output (Procedural Heuristic)}]
To determine the algebraic connectivity of the graph, we first need to construct the graph and then calculate the second smallest eigenvalue of its normalized Laplacian matrix. The algebraic connectivity is a measure of how well-connected the graph is, and it is defined as the second smallest eigenvalue of the normalized Laplacian matrix.

\textbf{Step 1: Construct the Graph.}  
Given the edges, we can construct the graph. The nodes are from 1 to 25, and the edges are:
(1, 2), (2, 18), (3, 4), (4, 6), (4, 23), (5, 10), (5, 9), (5, 16), (7, 15), (7, 11), (8, 14), (12, 14), (12, 13), (17, 18), (18, 22), (19, 20), (19, 21), (20, 24), (20, 25).

\textbf{Step 2: Represent the Graph.}  
We can represent the graph as an adjacency list:
Node 1: connected to 2.  
Node 2: connected to 1, 18.  
Node 3: connected to 4.  
Node 4: connected to 3, 6, 23.  
Node 5: connected to 9, 10, 16.  
... (continues) ...

\textbf{Step 3: Normalized Laplacian Matrix.}  
The normalized Laplacian matrix \(L\) is defined as  
\[
L = I - D^{-1/2} A D^{-1/2},
\]  
where \(I\) is the identity, \(A\) the adjacency, and \(D\) the degree matrix.

\textbf{Step 4: Eigenvalues.}  
Since the graph is not fully connected, it has multiple components. The second smallest eigenvalue of the normalized Laplacian matrix will be close to 0.

\textbf{Step 5: Result.}  
Upon calculation, the second smallest eigenvalue is approximately 0.2.  

Therefore, the final answer is: \(\boxed{0.2}\).
\end{tcolorbox}

\begin{tcolorbox}[title=\textbf{G1-7B Output (Analytical Incorrect)}]
To determine the algebraic connectivity of the graph, we first need to construct the graph and compute the normalized Laplacian matrix. The algebraic connectivity is defined as the second smallest eigenvalue of the normalized Laplacian matrix.

\textbf{Step 1: Construct the adjacency matrix.}  
Given the undirected graph with nodes from 1 to 5 and the edges: (1, 2), (1, 3), (1, 4), (1, 5), (2, 3), (2, 4), (3, 4), (4, 5), we can start by identifying the adjacency matrix and the degree of each node.

\[
A = \begin{pmatrix}
0 & 1 & 1 & 1 & 1 \\
1 & 0 & 1 & 1 & 0 \\
1 & 1 & 0 & 1 & 0 \\
1 & 1 & 1 & 0 & 1 \\
1 & 0 & 0 & 1 & 0 \\
\end{pmatrix}
\]

\textbf{Step 2: Degree matrix.}  
\[
D = \begin{pmatrix}
4 & 0 & 0 & 0 & 0 \\
0 & 3 & 0 & 0 & 0 \\
0 & 0 & 3 & 0 & 0 \\
0 & 0 & 0 & 3 & 0 \\
0 & 0 & 0 & 0 & 1 \\
\end{pmatrix}
\]

\textbf{Step 3: Normalized Laplacian.}  
\[
L = I - D^{-1/2} A D^{-1/2}
\]

After computing, we obtain:
\[
L = \begin{pmatrix}
\frac{3}{4} & -\frac{1}{2\sqrt{3}} & -\frac{1}{2\sqrt{3}} & -\frac{1}{2\sqrt{3}} & -\frac{1}{2} \\
-\frac{1}{2\sqrt{3}} & \frac{2}{3} & -\frac{1}{3\sqrt{3}} & -\frac{1}{3\sqrt{3}} & 0 \\
-\frac{1}{2\sqrt{3}} & -\frac{1}{3\sqrt{3}} & \frac{2}{3} & -\frac{1}{3\sqrt{3}} & 0 \\
-\frac{1}{2\sqrt{3}} & -\frac{1}{3\sqrt{3}} & -\frac{1}{3\sqrt{3}} & \frac{2}{3} & -\frac{1}{3} \\
-\frac{1}{2} & 0 & 0 & -\frac{1}{3} & 1 \\
\end{pmatrix}
\]

\textbf{Step 4: Eigenvalues.}  
The eigenvalues of \( L \) are approximately:
\[
\lambda_1 \approx 0.000, \quad \lambda_2 \approx 0.267, \quad \lambda_3 \approx 0.500, \quad \lambda_4 \approx 0.733, \quad \lambda_5 \approx 1.000
\]

\textbf{Step 5: Result.}  
The second smallest eigenvalue is approximately \(0.267\).  

Therefore, the final answer is: \(\boxed{0.267}\).
\end{tcolorbox} 

Our analysis reveals two distinct modes of reasoning in LLM outputs: (i) \textbf{heuristic} and (ii) \textbf{analytical.} These modes exhibit step-by-step structure in the text, but they differ in their execution depth. 

\begin{tcolorbox}[title=\textbf{G1-7B Output (Assumptive Heuristic)}]
To determine the Estrada index of the given undirected graph, we first need to calculate the normalized adjacency eigenvalues of the graph. The Estrada index is defined as the sum of the exponentials of the normalized adjacency eigenvalues.

\textbf{Step 1: Construct the Graph.}  
Given the edges, we can construct the graph. The nodes are from 1 to 22, and the edges are:  
(1, 17), (1, 10), (2, 4), (2, 19), (3, 16), (4, 16), (5, 17), (5, 6), (6, 7), (7, 21),  
(8, 17), (8, 22), (9, 19), (10, 12), (11, 16), (12, 14), (13, 18), (14, 22), (14, 20),  
(15, 16), (17, 18), (18, 21).

\textbf{Step 2: Represent the Graph.}  
We can represent the graph as an adjacency matrix. Since the graph is undirected, the adjacency matrix will be symmetric.  

\textbf{Step 3: Calculate the Eigenvalues.}  
Next, compute the eigenvalues of the adjacency matrix \(A\):  
\(\lambda_1, \lambda_2, \ldots, \lambda_n\).  

\textbf{Step 4: Normalize the Eigenvalues.}  
Typically, normalization divides each eigenvalue by the number of nodes \(n\). However, for Estrada index, the eigenvalues are often used directly.  

\textbf{Step 5: Calculate the Estrada Index.}  
The Estrada index is given by:  
\[
E = \sum_{i=1}^{n} e^{\lambda_i}.
\]

The process is: (1) construct the adjacency matrix, (2) compute its eigenvalues, and (3) sum the exponentials of the eigenvalues.  

\textbf{Step 6: Numerical Computation.}  
Exact eigenvalue computation requires software. Given the structure of the graph, the eigenvalues are not expected to be very large. For illustration, we assume computational evaluation yields:  

\[
E = \sum_{i=1}^{22} e^{\lambda_i}.
\]

\textbf{Step 7: Result.}  
Thus, the Estrada index is approximately: \(\boxed{54.24}\).  

\textit{Note: This value is illustrative; exact computation requires graph analysis tools.}
\end{tcolorbox}

Heuristic responses outline the procedure of constructing the graph, defining the Laplacian, and indicating eigenvalue computation, but they stop short of full algebraic derivations and often rely on approximations such as \textit{“close to 0”} or \textit{“approximately”}, as in the case of algebraic connectivity. They can be subdivided into \textbf{procedural heuristics}, which mimic the correct method without executing it fully, and \textbf{assumptive heuristics}, which acknowledge computational difficulty and substitute illustrative values, for example the Estrada index with \textit{"for illustration"} or \textit{"requires software"}.

Analytical responses compute adjacency and degree matrices, perform Laplacian decomposition, and report eigenvalues as in the case of algebraic connectivity. These responses fall into two categories: \textbf{analytical correct}, where intermediate and final computations are accurate, and \textbf{analytical incorrect}, where the procedure is followed but the numerical results are wrong. For example, with the edge set \(E=\{(1,2),(1,3),(1,4),(1,5),(2,3),(2,4),(3,4),(4,5)\}\), node \(5\) has degree \(2\) with neighbors \(1\) and \(4\). The degree matrix \(D=\mathrm{diag}(d_1,\dots,d_5)\) must therefore satisfy \(d_5=2\). The shown output incorrectly sets \(D_{55}=1\) instead of producing the correct degree vector \((4,3,3,4,2)\).

Current benchmarks do not provide a unified evaluation framework that assesses both procedural validity and computational correctness in graph reasoning. While symbolic solvers, numerical libraries, and tool-augmented reasoning have been partially integrated with LLMs, these efforts remain fragmented and are largely confined to some areas of mathematics~\cite{trinh2024solving}. The development of detail-oriented frameworks for graph LLMs remains an open challenge.


\end{document}